\documentclass[3p]{elsarticle}

\usepackage[utf8]{inputenc} 
\usepackage[T1]{fontenc}    
\usepackage{hyperref}       
\usepackage{url}            
\usepackage{booktabs}       
\usepackage{amsfonts}       
\usepackage{xcolor}         
\usepackage{graphicx}       
\usepackage{algorithmic}    
\usepackage{float}          
\usepackage{algorithm}      
\usepackage{url}            
\usepackage{amsmath}        
\usepackage{caption}        
\usepackage{subcaption}     
\usepackage{comment}        
\usepackage{multirow}       

\DeclareMathOperator*{\argmax}{argmax}

\newcounter{propositionCounter}
\newenvironment{proposition}{
\refstepcounter{propositionCounter}
\paragraph{Proposition \thepropositionCounter}
}{ }

\newcounter{corollaryCounter}
\newenvironment{corollary}{
\refstepcounter{corollaryCounter}
\paragraph{Corollary \thecorollaryCounter}
}{ }

\makeatletter
\def\ps@pprintTitle{%
 \let\@oddhead\@empty
 \let\@evenhead\@empty
 \def\@oddfoot{}%
 \let\@evenfoot\@oddfoot}
\makeatother

\begin{document}

\begin{frontmatter}

\title{Distributional Reinforcement Learning with Unconstrained Monotonic \\ Neural Networks}
\author[1]{Thibaut Th\'{e}ate}\corref{cor1}
\ead{thibaut.theate@uliege.be}
\author[1]{Antoine Wehenkel}
\ead{antoine.wehenkel@uliege.be}
\author[1]{Adrien Bolland}
\ead{adrien.bolland@uliege.be}
\author[1]{Gilles Louppe}
\ead{g.louppe@uliege.be}
\author[1,2]{Damien Ernst}
\ead{dernst@uliege.be}
\address[1]{Department of Electrical Engineering and Computer Science, University of Liège, Liège, Belgium}
\address[2]{Information Processing and Communications Laboratory, Institut Polytechnique de Paris, Paris, France}
\cortext[cor1]{Corresponding author.}

\begin{abstract}
    The distributional reinforcement learning (RL) approach advocates for representing the complete probability distribution of the random return instead of only modelling its expectation. A distributional RL algorithm may be characterised by two main components, namely the representation of the distribution together with its parameterisation and the probability metric defining the loss. The present research work considers the \textit{unconstrained monotonic neural network} (UMNN) architecture, a universal approximator of continuous monotonic functions which is particularly well suited for modelling different representations of a distribution. This property enables the efficient decoupling of the effect of the function approximator class from that of the probability metric. The research paper firstly introduces a methodology for learning different representations of the random return distribution (PDF, CDF and QF). Secondly, a novel distributional RL algorithm named \textit{unconstrained monotonic deep Q-network} (UMDQN) is presented. To the authors' knowledge, it is the first distributional RL method supporting the learning of \textit{three}, \textit{valid} and \textit{continuous} representations of the random return distribution. Lastly, in light of this new algorithm, an empirical comparison is performed between three probability quasi-metrics, namely the Kullback-Leibler divergence, Cramer distance, and Wasserstein distance. The results highlight the main strengths and weaknesses associated with each probability metric together with an important limitation of the Wasserstein distance.
    
\end{abstract}


\end{frontmatter}

\section{Introduction}
\label{SectionIntroduction}

\textit{Reinforcement learning} (RL) is a family of techniques belonging to the area of \textit{machine learning} (ML), which is concerned with the learning process of an agent sequentially interacting within an environment and aiming to maximise the notion of cumulative reward. \textit{Deep reinforcement learning} (DRL) extends this approach by using \textit{deep learning} (DL) techniques to generalise the information acquired from the interaction of the agent with its environment. Depending on whether a model of the environment is available and exploited or not, the RL algorithms can be either classified \textit{model-based} or \textit{model-free}. The present research focuses exclusively on the second category, which can be subdivided into two classes: \textit{policy optimisation} and \textit{Q-learning}. The RL algorithms based on the \textit{Q-learning} approach generally model the expectation of the random return to be maximised \cite{Watkins1992}. Alternatively, the \textit{distributional RL} approach proposes learning the entire probability distribution of the random return. This methodology presents key advantages including learning richer representations of the returns generated by the environment, which leads to more efficient and stable learning, as well as making risk-sensitive control and exploration policies possible \cite{Bellemare2017C51, Lyle2019}.\\

A distributional RL algorithm may be characterised by two main components. The first one relates to both the representation and the parameterisation of the random return distribution. A unidimensional distribution possesses several different representations, such as its probability density function (PDF), its cumulative distribution function (CDF) and its quantile function (QF). Typically, \textit{deep neural networks} (DNNs) are considered for approximating these various functions. The second component concerns the probability quasi-metric adopted for comparing two distributions. Multiple quasi-metrics do exist for that purpose, the main ones experimented in distributional RL being the Kullback-Leibler (KL) divergence, the Cramer distance (which is also named energy distance), and the Wasserstein distance. In the rest of this research paper, they will simply be referred to as \textit{probability metrics}. In the context of distributional RL, the role of the probability metric is to quantitatively compare two distributions of the random return in order to apply a \textit{temporal difference} (TD) learning method, in a similar way to the mean squared error between Q-values in classical RL. The choice of the probability metric is particularly important since each metric offers different theoretical convergence guarantees for distributional RL.\\

The core idea of this research work is to consider the \textit{unconstrained monotonic neural network} (UMNN) architecture \cite{Wehenkel2019} in the scope of distributional RL. Originally designed for autoregressive flows, this particular architecture is in fact a universal approximator of continuous monotonic functions. Several works have already demonstrated the ability of this neural network to accurately model continuous monotonic functions in practice \cite{Rahimi2020, Wang2020}. Since both the CDF and QF are monotonic, the UMNN architecture is expected to offer superior capability compared to classical neural networks when it comes to representing distributions. Moreover, the PDF can also be efficiently represented by this architecture, when standing at the heart of a \textit{normalizing flow} \cite{Rezende2015} by taking advantage of the \textit{change of variables theorem} \cite{Wehenkel2019}. Because the single UMNN architecture can effectively model different representations of the random return distribution, it enables the efficient decoupling of the effect of the function approximator class from that of the probability metric, making a fair comparison between probability metrics possible.\\

This leads to the main contributions of the present research work, which are threefold. Firstly, the paper introduces a methodology for learning three representations of the random return probability distribution, namely the PDF, CDF and QF. Secondly, the article presents a novel distributional RL algorithm, denominated \textit{unconstrained monotonic deep Q-network} (UMDQN), combining a UMNN with this new methodology for learning different valid representations of the continuous distribution of the random return. Thirdly, taking advantage of this innovative algorithm, the research work proposes an empirical comparison of three probability metrics commonly used in distributional RL, namely the KL divergence, the Cramer distance and the Wasserstein distance. This analysis highlights the main strengths and weaknesses associated with each probability metric, but also reveals an important limitation of the Wasserstein distance. Actually, the observed limitation highlights a critical approximation made by several state-of-the-art distributional RL algorithms, leading to the learning of inaccurate distributions for the random return. To the authors' knowledge, the proposed algorithmic solution is the first distributional RL approach supporting the learning of \textit{several} (PDF, CDF and QF), \textit{valid} (by ensuring monotonicity) and \textit{continuous} (as opposed to discrete) representations of the random return distribution. To end this introductory section, it should be emphasised that the core objective of this research work is not to present a novel distributional RL algorithm competing with the state-of-the-art algorithms on a given testbench, typically the Atari-57 benchmark \cite{Bellemare2013}, but rather to empirically derive new insights about distributional RL from this algorithmic solution.\\

The present research paper is structured as follows. First of all, a concise review of the scientific literature about distributional RL in general and the state-of-the-art algorithms is presented in Section \ref{SectionLiteratureReview}. Afterwards, Section \ref{SectionDistributionalRL} formally introduces the distributional RL approach together with the mathematical notations adopted in this research work. Then, Section \ref{SectionMethods} presents in detail the novel distributional RL algorithm proposed, together with a methodology for learning different representations of the random return probability distribution. Following on, Section \ref{SectionResults} presents the performance assessment methodology adopted and discusses the results achieved by this new distributional RL algorithm. Finally, Section \ref{SectionConclusions} draws some conclusions and briefly discusses interesting leads for future work.

\section{Literature review}
\label{SectionLiteratureReview}

\textit{Q-learning} is a \textit{model-free} RL approach based on the learning of the quantity $Q$ representing the quality of executing a certain action in a particular state \cite{Watkins1992}. Originally based on tabular or linear approximations, the DQN algorithm \cite{Mnih2015} extends this approach by using a DNN for approximating the quantity $Q$ in a non-linear setting. New to the field is the distributional RL approach advocating for learning the entire probability distribution of the random return instead of only modelling its expectation \cite{Bellemare2017C51}. Fundamental research on distributional RL is still in its early stages, but key benefits have already been discovered \cite{Lyle2019, Rowland2019}.\\

\begin{figure}[H]
    \centering
    \includegraphics[width=1\linewidth, trim={2.8cm 4.8cm 3.5cm 5.5cm}, clip]{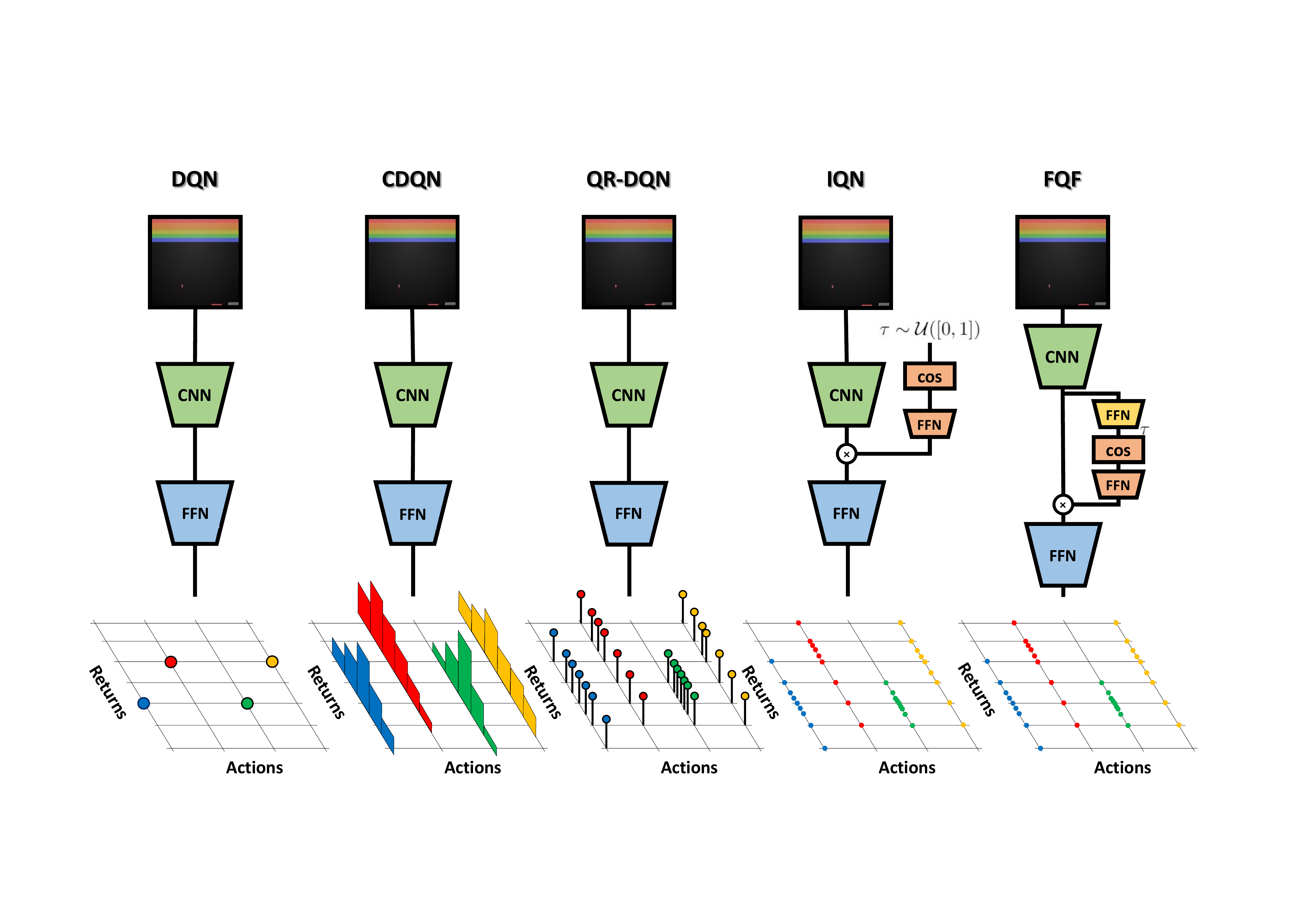}
    \caption{Main distributional RL algorithms from scientific literature, for the Atari-57 benchmark \cite{Bellemare2013}.}
    \label{IllustrationAlgosDistributionalRL}
\end{figure}

Several distributional RL algorithms can be found in scientific literature, based on diverse representations of the random return distribution but also different probability metrics. The \textit{categorical DQN} (CDQN) algorithm \cite{Bellemare2017C51}, also known as \textit{C51}, approximates the PDF of the random return through categorical distributions and uses the KL divergence for quantitatively comparing these distributions. The link between this original distributional RL algorithm and the Cramer distance probability metric was later highlighted \cite{Rowland2018}. Alternatively, the \textit{quantile regression DQN} (QR-DQN) algorithm \cite{Dabney2018QRDQN} learns the distribution of the random return by manipulating the QF with fixed uniform quantile values and the Wasserstein distance. Compared to the CDQN algorithm, this approach has the key advantage of avoiding the specification of a fixed support for the random return values. Nevertheless, both algorithms suffer from the same drawback of estimating the distribution of the random return on fixed locations (either value or probability), with as a consequence that the distributions learnt are discrete. The \textit{implicit quantile network} (IQN) algorithm \cite{Dabney2018IQN} solves this problem by learning the quantile values from quantile fractions sampled from a uniform distribution $\mathcal{U}([0, 1])$. This is achieved with a specific DNN representing the QF by mapping quantile fractions to quantile values and trained by minimising the Wasserstein distance. Finally, the \textit{fully parameterised quantile function} (FQF) algorithm \cite{Yang2019FQF} extends the previous methodology by parameterising both quantile fraction and value axes. To do so, two DNNs are used: one for generating appropriate quantile fractions and one for mapping these quantile fractions to quantile values. They are jointly trained by minimising the Wasserstein distance once again. Figure \ref{IllustrationAlgosDistributionalRL} illustrates these distributional RL algorithms in the context of the Atari-57 benchmark \cite{Bellemare2013}.\\

\begin{table}[H]
      \caption{Key characteristics (representation of the random return probability distribution and probability metric) of the main state-of-the-art distributional RL algorithms.}
      \label{DistributionalRLAlgorithms}
      \centering
      \begin{tabular}{lll}
        \toprule
        \textbf{Algorithm} & \textbf{Probability distribution representation} & \textbf{Probability metric} \\
        \midrule
        DQN & Expectation (non distributional RL) & L1 metric \\
        CDQN & Categorical PDF (fixed discrete support) & KL divergence \\
        QR-DQN & Discrete QF (fixed quantile fractions) & Wasserstein distance \\
        IQN & Continuous QF (quantiles drawn from $\mathcal{U}([0, 1])$) & Wasserstein distance \\
        FQF & Continuous QF (quantiles sampled by a DNN) & Wasserstein distance \\
        \bottomrule
      \end{tabular}
    \end{table}

Besides the previous distributional RL algorithms that are well-established in the research community, one can mention several recent research works bringing new interesting insights about distributional RL. The \textit{moment matching DQN} (MMDQN) algorithm \cite{NguyenTang2021} learns via a DNN a finite set of statistics for the distribution of the random return by implicitly matching all orders of moments between the random return distribution and its target. The key benefit of this approach is to avoid the predefined statistic principle used in prior distributional RL works, which leads to a simpler objective amenable to backpropagation. This is achieved by learning unrestricted statistics, i.e. deterministic samples, of the random return distribution by leveraging the maximum mean discrepancy technique from hypothesis testing. Sharing a similar philosophy to the present research paper, the \textit{non-crossing QR-DQN} algorithm \cite{Zhou2020} is an improvement of the well-established QR-DQN algorithm implementing non-crossing quantile regression to ensure the monotonicity constraint for the QF. This enhancement is built on the observation that the non-decreasing property of learnt quantile curves is not guaranteed, which leads to abnormal distribution estimates and reduced model interpretability. However, this technique is not directly transferable to the IQN or FQF algorithms. To end this literature review, an important study reveals that the reward system of the human brain would operate similarly to distributional RL \cite{Dabney2020}. Indeed, the findings suggest that the human brain represents possible future rewards as a complete probability distribution and not as a single mean of stochastic outcomes. This is naturally very encouraging news supporting the soundness of the distributional RL approach.\\

\section{Distributional Reinforcement Learning}
\label{SectionDistributionalRL}

This research paper adopts the standard RL setting where the agent interacts with its environment modelled as a \textit{Markov decision process} (MDP). An MDP is a 6-tuple $(\mathcal{S},\ \mathcal{A},\ p_R,\ p_T,\ p_0,\ \gamma)$ where $\mathcal{S}$ and $\mathcal{A}$ respectively are the state and action spaces, $p_R(r|s,a)$ is the probability distribution from which the reward $r \in \mathbb{R}$ is drawn given a state-action pair $(s, a)$, $p_T(s'|s,a)$ is the transition probability distribution, $p_0(s_0)$ is the probability distribution over the initial states $s_0 \in \mathcal{S}$, and $\gamma \in [0, 1[$ is the discount factor. The RL agent makes decisions according to its policy $\pi: \mathcal{S} \rightarrow \mathcal{A}$, which is considered deterministic in the rest of this research paper, mapping the states $s \in \mathcal{S}$ to the actions $a \in \mathcal{A}$.\\

The Q-learning approach focuses on modelling the \textit{state-action value function} $Q^{\pi}$ of a policy $\pi$. This quantity represents the expected discounted sum of rewards to be obtained by executing an action $a$ in a state $s$ and then following a policy $\pi$, and satisfies the \textit{Bellman equation} \cite{Bellman1957}:
\begin{equation}
    Q^{\pi}(s, a) = \underset{s_t, r_t}{\mathbb{E}} \left[\sum_{t=0}^{\infty} \gamma^{t} r_t\right] \text{,} \ \ \ \ \ (s_0, a_0) := (s, a),\ a_{t} = \pi(s_{t})\ \text{,}
\end{equation}
\begin{equation}
    Q^{\pi}(s, a) = \underset{s', r}{\mathbb{E}} \left[r + \gamma Q^{\pi}(s', \pi(s'))\right] \text{.}
\end{equation}

In a similar way, the \textit{optimal policy} $\pi^*$ based on the \textit{optimal state-action value function} $Q^*$ can be defined as the following:

\begin{equation}
    Q^{*}(s, a) = \underset{s', r}{\mathbb{E}} \left[r + \gamma \max_{a' \in \mathcal{A}}Q^{*}(s', a')\right] \text{,}
\end{equation}
\begin{equation}
    \pi^*(s) \in \argmax_{a \in \mathcal{A}} Q^*(s, a)\ \text{.}
\end{equation}

Distributional RL aims at modelling the entire probability distribution over returns instead of only its expectation. To this end, let the reward $R(s, a)$ be a random variable distributed under $p_R(\cdot|s, a)$, the \textit{state-action value distribution} $Z^{\pi} \in \mathcal{Z}$ of a policy $\pi$ is a random variable defined as follows:
\begin{equation}
    Z^{\pi}(s, a) \stackrel{D}{=} \sum_{t=0}^{\infty} \gamma^{t} R(s_t, a_t)\ \text{,} \ \ \ \ \ (s_0, a_0) := (s, a),\ a_{t} = \pi(s_{t}),\ s_{t+1} \sim p_T(\cdot|s_t, a_t)\ \text{,}
\end{equation}
where $A \stackrel{D}{=} B$ denotes the equality in distribution between the random variables $A$ and $B$. Therefore, the state-action value function $Q^{\pi}$ is the expectation of the \textit{random return} $Z^{\pi}$. In the same way, there is a \textit{distributional Bellman equation} recursively describing $Z^{\pi}$:
\begin{equation}
\label{EquationDistributionalBellman}
    Z^{\pi}(s, a) \stackrel{D}{=} R(s, a) + \gamma P^{\pi} Z^{\pi}(s, a)\ \text{,}
\end{equation}
\begin{equation}
    P^{\pi} Z^{\pi}(s, a) :\stackrel{D}{=} Z^{\pi}(s', a')\ \text{,} \ \ \ \ \ \ \ \ \ \ s' \sim p_T(\cdot|s, a),\ a' = \pi(s')\  \text{,}
\end{equation}
where $P^{\pi} : \mathcal{Z} \rightarrow \mathcal{Z}$ is the transition operator. Finally, one can define the \textit{distributional Bellman operator} $\mathcal{T}^{\pi} : \mathcal{Z} \rightarrow \mathcal{Z}$ and the \textit{distributional Bellman optimality operator} $\mathcal{T}^* : \mathcal{Z} \rightarrow \mathcal{Z}$ as follows:
\begin{equation}
    \mathcal{T}^{\pi} Z^{\pi}(s, a) \stackrel{D}{=} R(s, a) + \gamma P^{\pi} Z^{\pi}(s, a)\ \text{,}
\end{equation}
\begin{equation}
    \mathcal{T}^* Z^*(s, a) \stackrel{D}{=} R(s, a) + \gamma Z^*\left(s', \pi^*(s')\right) \text{,} \ \ \ \ \ s' \sim p_T(\cdot|s, a)\ \text{.}
\end{equation}

Theoretically, the distributional Bellman operator $\mathcal{T}^{\pi}$ may potentially be a contraction mapping or not depending on the probability metric. This property implies that there exists a unique fixed point $Z^{\pi}$ to converge towards when repeatedly applying the operator $\mathcal{T}^{\pi}$. For the distributional Bellman optimality operator $\mathcal{T}^*$, another condition is required for this contraction mapping property to hold: the optimal policy $\pi^*$ has to be unique \cite{Bellemare2017C51}. Multiple probability metrics do exist for quantitatively comparing the probability distributions of two continuous random variables. In this research work, the emphasis is set on the three main probability metrics used in distributional RL, namely the KL divergence, Cramer distance and Wasserstein distance. Table \ref{ProbabilityMetrics} formally introduces these probability metrics, together with their impact on the contraction mapping property of the distributional Bellman operator $\mathcal{T}^{\pi}$.\\

\begin{table}[H]
  \caption{Formal definition of the probability metrics studied, where $A$ and $B$ are two random variables, and where $p_D$, $F_D$ and $F_D^{-1}$ denote the PDF, CDF and QF of the random variable $D$, respectively.}
  \label{ProbabilityMetrics}
  \centering
  \begin{tabular}{llc}
    \toprule
    \textbf{Probability metric} &  & \textbf{$\mathcal{T}^{\pi}$ contraction?} \\
    \midrule
    KL divergence & $\mathcal{L}_{KL}(A, B) = D_{KL}(p_A, p_B) = \int_{\mathbb{R}} p_A(x) \log \left(\frac{p_A(x)}{p_B(x)}\right) dx$ & No \cite{Morimura2010} \\
    Cramer distance & $\mathcal{L}_{C}(A, B) = D_C(F_A, F_B) = \left(\int_{\mathbb{R}}\left(F_A(x) - F_B(x)\right)^2 dx \right)^{1/2}$ \vphantom{$\int_{\mathbb{R}} p_A(x) \log \left(\frac{p_A(x)}{p_B(x)}\right) dx$} & Yes \cite{Rowland2018}\\
    Wasserstein distance & $\mathcal{L}_{W}(A, B) = D_W(F_A^{-1}, F_B^{-1}) = \int_{0}^{1} F_A^{-1}(x) - F_B^{-1}(x) dx$ \vphantom{$\int_{\mathbb{R}} p_A(x) \log \left(\frac{p_A(x)}{p_B(x)}\right) dx$} & Yes \cite{Bellemare2017C51}\\
    \bottomrule
  \end{tabular}
\end{table}

\section{Unconstrained monotonic deep Q-network}
\label{SectionMethods}

\subsection{Learning different representations of a probability distribution}
\label{SectionAbstraction}

\begin{figure}[H]
    \centering
    \includegraphics[width=0.9\linewidth, trim={1cm 8.5cm 1.2cm 5cm}, clip]{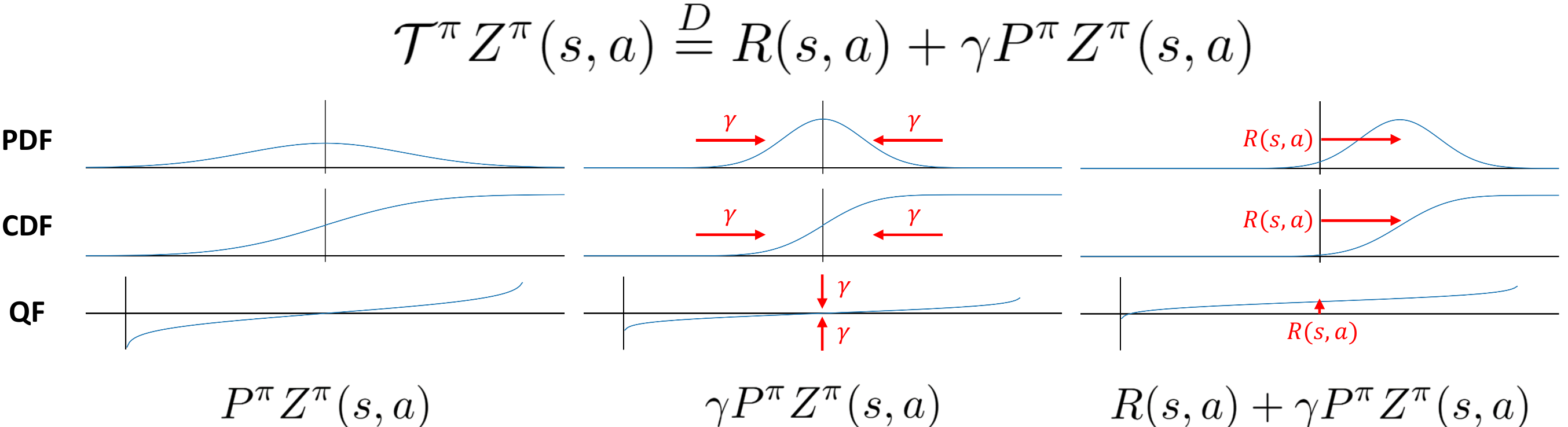}
    \caption{Illustration of the effect of the distributional Bellman operator on different representations of the random return probability distribution in the simplified situation of both a deterministic reward and a deterministic transition function.}
    \label{FigureIntuitiveDistributionalBellmanOperator}
\end{figure}

This section presents a methodology for learning different representations of the probability distribution of the random return: the PDF, CDF and QF. The learning process is based on the comparison of the left- and right-hand sides of the distributional Bellman equation \eqref{EquationDistributionalBellman}. For a given probability metric $\mathcal{L}$, the random return $Z^{\pi}$ is a fixed point of the Bellman operator $\mathcal{T}^{\pi}$ if it minimises the following loss:
\begin{equation}
\label{EquationLoss}
    \mathcal{L} \left(\mathcal{T}^{\pi} Z^{\pi}(s, a),\ Z^{\pi}(s, a) \right) \text{.}
\end{equation}
for all state-action pairs $(s, a) \in \mathcal{S} \times \mathcal{A}$. The distributional RL problem at hand will be addressed by defining a hypothesis space for the quantity $Z^{\pi}$ and minimising the loss function \eqref{EquationLoss} over this space using \textit{stochastic gradient descent} (SGD). In the following, the effect of the distributional Bellman operator $\mathcal{T}^{\pi}$ on the different representations of the random return distribution is rigorously studied. Intuitively, the discount factor $\gamma$ \textit{squeezes} the random return distribution while the reward $R$ \textit{shifts} this probability distribution, as illustrated in Figure \ref{FigureIntuitiveDistributionalBellmanOperator} in the simplified situation of deterministic reward and transition function.

\paragraph{\textbf{PDF representation}} Let $p_{Z^{\pi}}(z|s, a)$ be the PDF of the random variable $Z^{\pi}$ given the state-action pair $(s, a)$ at the return $z$. Assuming the KL divergence $\mathcal{L}_{KL}$ as the probability metric considered, the loss to be minimised defined in Equation \eqref{EquationLoss} can be re-expressed as follows:
\begin{align}
    \label{EquationLossPDF}
    \mathcal{L}_{KL}(\mathcal{T}^{\pi} Z^{\pi}(s, a),\ Z^{\pi}(s, a))
    &= D_{KL}\left(p_{\mathcal{T}^{\pi} Z^{\pi}}(z|s, a),\ p_{Z^{\pi}}(z|s, a)\right) \\
    &= D_{KL}\left(\underset{s', r}{\mathbb{E}} \left[\frac{1}{\gamma}\  p_{Z^{\pi}}\left(\frac{z - r}{\gamma} \bigg| s', \pi(s') \right) \right], p_{Z^{\pi}}(z|s, a) \right) \text{.} \label{EquationLossPDFBis}
\end{align}

\paragraph{\textbf{CDF representation}} Let $F_{Z^{\pi}}(z|s, a)$ be the CDF of the random variable $Z^{\pi}$ conditioned by the state-action pair $(s, a)$ at the return $z$. Assuming the Cramer distance $\mathcal{L}_{C}$ as the probability metric considered, the loss formally defined in Equation \eqref{EquationLoss} can be re-expressed as follows:
\begin{align}
    \label{EquationLossCDF}
    \mathcal{L}_{C}(\mathcal{T}^{\pi} Z^{\pi}(s, a),\ Z^{\pi}(s, a))
    &= D_{C}\left(F_{\mathcal{T}^{\pi} Z^{\pi}}(z|s, a),\ F_{Z^{\pi}}(z|s, a)\right) \\
    &= D_{C}\left(\underset{s', r}{\mathbb{E}} \left[ F_{Z^{\pi}}\left(\frac{z - r}{\gamma} \bigg| s', \pi(s') \right) \right], F_{Z^{\pi}}(z|s, a) \right) \text{.} \label{EquationLossCDFBis}
\end{align}

\paragraph{\textbf{QF representation}} Let $F^{-1}_{Z^{\pi}}(\tau|s, a)$ be the QF of the random variable $Z^{\pi}$ given the state-action pair $(s, a)$ at the quantile fraction $\tau \in [0, 1]$. Assuming the Wasserstein distance $\mathcal{L}_{W}$ as the probability metric considered, the loss to be minimised defined in Equation \eqref{EquationLoss} can be re-expressed as follows:
\begin{align}
    \label{EquationLossQF}
    \mathcal{L}_{W}(\mathcal{T}^{\pi} Z^{\pi}(s, a),\ Z^{\pi}(s, a))
    &= D_{W}\left(F^{-1}_{\mathcal{T}^{\pi} Z^{\pi}}(\tau|s, a),\ F^{-1}_{Z^{\pi}}(\tau|s, a)\right) \\
    &\simeq D_{W}\left(\underset{s', r}{\mathbb{E}} \left[r + \gamma F^{-1}_{Z^{\pi}}(\tau|s', \pi(s'))\right], F^{-1}_{Z^{\pi}}(\tau|s, a)\right) \text{.} \label{EquationLossQFBis}
\end{align}

As far as mathematical proofs are concerned, Equations \eqref{EquationLossPDFBis} and \eqref{EquationLossCDFBis} are respectively supported by Proposition \ref{prop:rec_pdf_def_z} and Corollary \ref{prop:rec_cdf_def_z} in \ref{AppendixDemo}. On the contrary, Equation \eqref{EquationLossQFBis} could not be rigorously proven as originally intended. In order to get a better understanding of the challenge faced, some basic experiments have been conducted. The results suggest that Equation \eqref{EquationLossQFBis} results from an approximation of $F^{-1}_{\mathcal{T}^{\pi} Z^{\pi}}(\tau|s, a)$, leading to a random variable with the correct expectation but potentially different higher-order moments. In the scope of distributional RL, such an approximation may have two completely different implications depending on the objective pursued. If the intention is to accurately learn the probability distribution of the random return for implementing risk-aware policies, this approximation is obviously problematic. On the contrary, if the goal is to learn policies maximising the expectation of the random return, this approximation may have no negative effect since the distribution learnt has the correct first-order moment. In fact, this approach is adopted by the state-of-the-art QR-DQN, IQN and FQF algorithms which are able to learn valuable policies in practice, based on the expectation of the random return alone \cite{Dabney2018QRDQN, Dabney2018IQN, Yang2019FQF}.\\

\subsection{Unconstrained monotonic neural network}
\label{SectionUMNN}

The PDF, CDF and QF of continuous random variables share the important property of being effectively modelled with strictly monotonic functions. This is the main reason for this research paper to consider \textit{unconstrained monotonic neural networks} (UMNNs), which are universal approximators of continuous monotonic functions, for parameterising the random return probability distribution. Formally, a UMNN defines a parametric continuous monotonic function $G(\cdot; \theta): \mathbb{R} \rightarrow \mathbb{R}$ as follows:
\begin{align}
\label{EquationUMNN}
    G(x; \theta)
    := \int^x_0 g(t; \theta) dt + \beta\ \text{,}
\end{align}
where $g(\cdot; \theta) : \mathbb{R} \rightarrow \mathbb{R}^+$ is a free-form neural network whose output positiveness is enforced via an appropriate activation function (e.g. ReLU or exponential), where $ \theta$ denotes its parameters, and where $\beta \in \mathbb{R}$ is a trainable scalar parameter. This parameterisation can efficiently generalise to random variables conditioned by other quantities, e.g., the state $s$ and the action $a$. A natural solution is to add these conditioning variables $c$ as an additional vector input to the neural network $g$ and to parameterise $\beta$ as another neural network. In this particular case, Equation \eqref{EquationUMNN} can be re-expressed as follows:
\begin{align}
G(x|c;\theta) := \int^x_0 g(t, c; \theta_g) dt + \beta(c; \theta_\beta)\ \text{,}
\end{align}
where the parameters of the monotonic transformation are $\theta = \theta_g \cup \theta_\beta$. Evaluating the function $G$ requires solving an integral, which is performed numerically via Clenshaw-Curtis quadrature.\\

In the scope of distributional RL, the QF of the random return $Z$ taking as inputs quantile fractions $\tau \in \left[0, 1\right]$ can be parameterised by a UMNN as $F^{-1}_Z(\tau|s, a; \theta) := G(\tau|s, a; \theta)$. Modelling the CDF of the random return $Z$ requires the output to be bounded in $\left[0, 1\right]$, which is achieved by passing the output of the UMNN through a sigmoid function $\sigma$: $ F_Z(z|s, a; \theta) := \sigma(G(z|s, a; \theta))$. Modelling the random return PDF $p_Z(z|s, a; \theta)$ can be done via \textit{normalizing flows} \cite{Rezende2015}. More precisely, it is achieved by using a fixed latent distribution $p_Y$ and exploiting the property that there exists a unique continuous monotonic function $f$ satisfying the following equation (\textit{change of variables theorem}) \cite{Wehenkel2019}:
\begin{align}
    p_Z(z|s, a; \theta) = p_Y(f(z|s, a; \theta)) \bigg|\frac{\partial f}{\partial z}\bigg|\ \text{.}
\end{align}

The representation of $p_Z$ is achieved by modelling the function $f$ with a UMNN and fixing $p_Y$ to an isotropic normal distribution. With such a representation, drawing samples from $p_Z$ is performed by drawing samples from $p_Y$ and applying the function $f^{-1}$. This requires inverting the UMNN, which can be done numerically by using any inversion method such as a binary search, since the inverse of a monotonic function is also monotonic. \ref{AppendixUMNN} provides additional information about the use of UMNNs in this research work.\\

\subsection{Unconstrained monotonic deep Q-network algorithm}
\label{SectionUMDQN}

\begin{figure}[H]
    \centering
    \includegraphics[width=0.8\linewidth, trim={3.6cm 5.4cm 3.4cm 5.5cm}, clip]{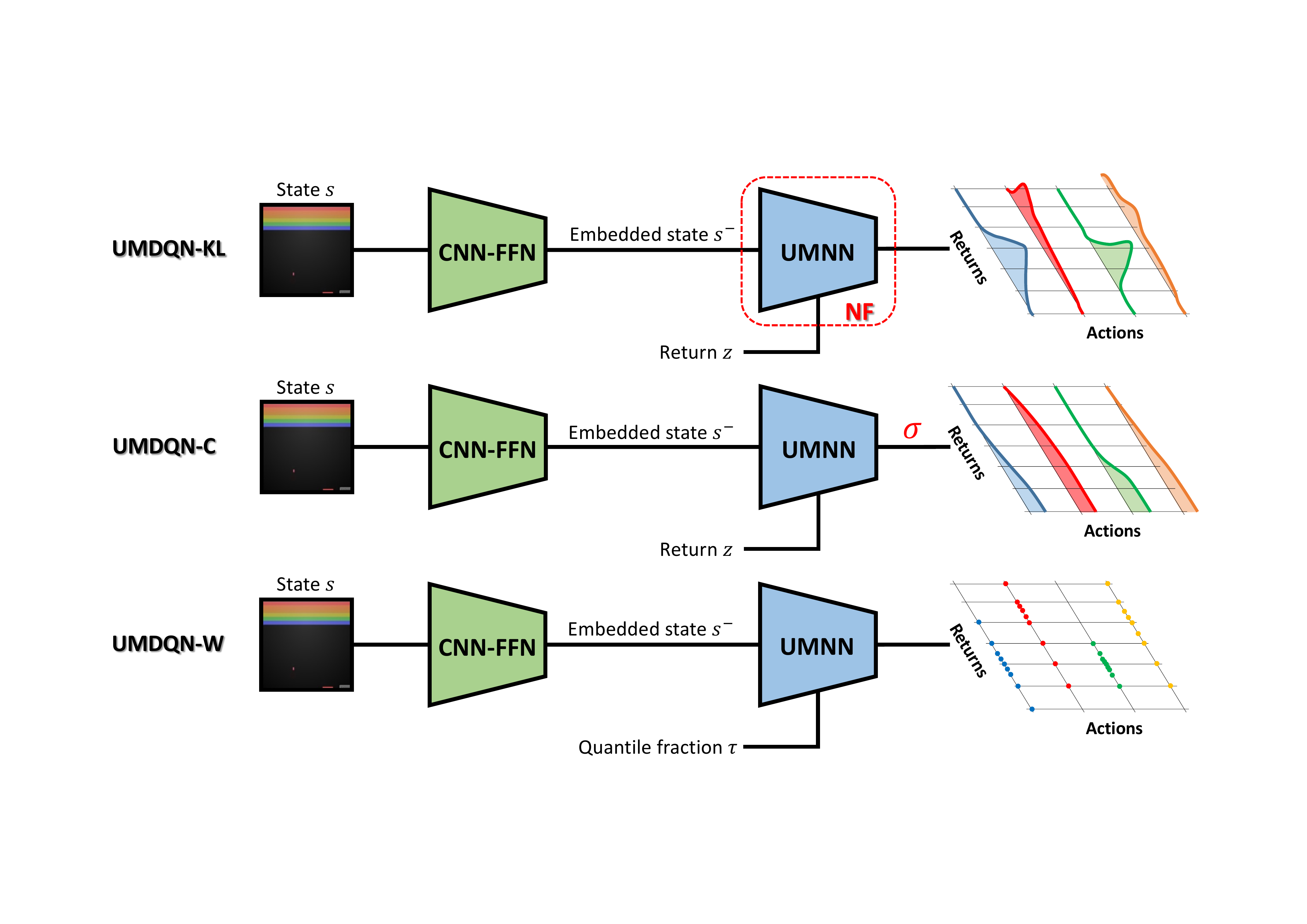}
    \caption{Illustration of the three versions of the UMDQN algorithm in the context of Atari games.}
    \label{IllustrationUMDQN}
\end{figure}

This section presents the \textit{unconstrained monotonic deep Q-network} (UMDQN) algorithm, a novel generic distributional RL algorithm based on the methodology introduced in Section \ref{SectionAbstraction} and working with the UMNN architecture for validly representing the continuous probability distribution of the random return. More precisely, this research work details three versions of the generic UMDQN distributional RL algorithm: the UMDQN-KL, UMDQN-C and UMDQN-W algorithms, which respectively approximate the continuous PDF, CDF and QF of the random return $Z^{\pi}$ by minimising the KL divergence, Cramer distance and Wasserstein distance. Therefore, in contrast to previous works on distributional RL, the proposed approach presents the key advantage of offering a choice regarding the representation of the probability distribution together with the probability metric to work with. An illustration of the three novel distributional RL algorithms in the context of Atari games is provided in Figure \ref{IllustrationUMDQN}.\\

The UMDQN algorithm is an \textit{off-policy} and \textit{value iteration} DRL algorithm which is based on the same procedure as the DQN algorithm for generating trajectories and learning from that information. Numerous experiences $e = \left(s, a, r, s'\right)$ are generated by sequentially interacting with the environment and are stored into an \textit{experience replay memory} of fixed size with a first-in-first-out (FIFO) replacement policy. Additionally, a \textit{target network}, whose parameters are denoted $\theta^-$, is used for fixing the Bellman probability distribution to be learnt and is updated at regular intervals. As far as exploration is concerned, it is ensured through the use of the $\epsilon$-greedy technique. At regular intervals during the interactions between the agent and its environment, batches of experiences are sampled from the replay memory to compute \textit{Monte Carlo} (MC) estimates of an approximation of the loss defined in Equation \eqref{EquationLoss} and perform stochastic gradient descent.\\

In fact, three important approximations are made regarding the loss defined in Equation \eqref{EquationLoss}. The first one results from the evaluation of the loss in expectation over the distribution of state-action pairs \textit{sampled} from the environment. The second approximation  originates from the fact that the expectation $\mathbb{E}_{s', r}$ in Equations \eqref{EquationLossPDFBis}, \eqref{EquationLossCDFBis}, \eqref{EquationLossQFBis} is computed \textit{outside} the probability metric $\mathcal{L}$. The last approximation comes from the estimation of the two expectations $\mathbb{E}_{s, a}$ and $\mathbb{E}_{s', r}$ using \textit{Monte Carlo} with the experiences sampled from the replay memory. The second approximation may potentially introduce a bias, as it has already been demonstrated for the Wasserstein distance \cite{Bellemare2017C51}. However, there is a solution for this probability metric in particular: the \textit{(conditional) quantile regression} method \cite{Koenker2005}. This approach is claimed to allow for the unbiased stochastic approximation of the QF, and is adopted in the QR-DQN, IQN and FQF algorithms.\\

The learning process of the UMDQN algorithm is described in Algorithm \ref{UMDQNAlgorithm}. Within this description, $G_Z(\cdot|s, a; \theta)$ denotes the random return probability distribution modelled by a UMNN with parameters $\theta$ for the state-action pair $(s, a)$, the operator $T^{\pi}$ is defined in Equation \eqref{EquationOperatorT} and reproduces the effect of the distributional Bellman operator on $G_Z(\cdot|s, a; \theta)$ in line with Equations \eqref{EquationLossPDFBis}, \eqref{EquationLossCDFBis} and \eqref{EquationLossQFBis}, the function $L$ computes the error according to the probability metric selected, and $\mathcal{X}$ is a discretisation of the domain of the function representing probability distribution of the random return (PDF, CDF or QF). In this research work, the policy $\pi$ considered simply selects the action maximising the expectation of the random return $Z^{\pi}$ learnt so far. The detailed pseudocodes of the three versions of the UMDQN algorithm, together with some implementation details, are provided in \ref{AppendixPseudocode}.\\

\begin{equation}
\label{EquationOperatorT}
    T^{\pi} G_Z(x| s, a; \theta) = \left\{
                                    \begin{array}{ll}
                                        \frac{1}{\gamma} G_Z\left(\frac{x-r}{\gamma} \big| s', \pi(s'); \theta\right) & \mbox{if the UMNN models a PDF,}\\
                                        G_Z\left(\frac{x-r}{\gamma} \big| s', \pi(s'); \theta\right) & \mbox{if the UMNN models a CDF,} \vphantom{\frac{1}{\gamma} G_Z\left(\frac{x-r}{\gamma} \big| s', \pi(s'); \theta\right)}\\
                                        r + \gamma G_Z(x| s', \pi(s'); \theta) & \mbox{if the UMNN models a QF.} \vphantom{\frac{1}{\gamma} G_Z\left(\frac{x-r}{\gamma} \big| s', \pi(s'); \theta\right)}
                                    \end{array}
                                   \right.
\end{equation}

\begin{algorithm*}
\small
\caption{Learning process of the UMDQN algorithm}
\begin{algorithmic} 
\STATE Sample a batch of $N_e$ experiences $e = (s, a, r, s')$ from the replay memory.
\STATE Determine for each experience the next optimal action $a' = \pi(s') = \argmax_{a \in \mathcal{A}} \mathbb{E}\left[G_Z(s', a; \theta^-)\right]$.
\STATE Compute the loss $\hat{\mathcal{L}} = \frac{1}{N_e} \sum_{s, a, r, s'} \left[ \sum_{x \in \mathcal{X}}  \left[L\left(T^{\pi} G_Z(x| s, a; \theta^-),\ G_Z(x| s, a; \theta) \right) \right] \right]$.
\STATE Optimise the UMNN parameters $\theta$ according to the resulting gradients $\nabla \hat{\mathcal{L}}$.
\end{algorithmic}
\label{UMDQNAlgorithm}
\end{algorithm*}

\section{Results}
\label{SectionResults}

\subsection{Benchmark environments}
\label{SectionBenchmarkEnvironments}

The performance assessment methodology adopted by this research work to evaluate the performance of the UMDQN distributional RL algorithm includes four different types of benchmark environments:
\begin{itemize}
    \item [$\bullet$] a stochastic grid world environment,
    \item [$\bullet$] a set of classic control environments,
    \item [$\bullet$] a set of Atari games,
    \item [$\bullet$] a set of MinAtar games.
\end{itemize}

The first benchmark environment is a \textit{stochastic grid world} designed in the scope of this particular research on distributional RL. It consists of a $7 \times 7$ grid world within which an agent has to reach a fixed target location while avoiding a fixed trap. In order to provide sound and interesting analyses in relation to the distributional RL approach, both transition and reward functions are set stochastic ($p_T$ and $p_R$). In addition to evaluating the policy performance, this specific environment will be particularly useful for visualising and interpreting the random return probability distributions learnt by the distributional RL algorithm.\\

The next type of benchmark environment is a set of \textit{four classic control} problems from OpenAI Gym \cite{Brockman2016}: CartPole, Acrobot, MountainCar and LunarLander. Although the distributional RL community generally prefers Atari games to these simpler environments, they remain particularly valuable and popular benchmarks for evaluating RL algorithms. Moreover, these environments are promoted by the article \cite{Obando2021} which proposes an alternative set of benchmarks which are less computationally intensive. That particular work being interesting and well received by the RL research community, this research paper adopts its suggestions.\\

The third type of benchmark environment is a set of \textit{three Atari games} from the Atari-57 benchmark \cite{Bellemare2013}: Pong, Boxing and Freeway. Distributional RL algorithms are generally evaluated on the complete Atari-57 benchmark, which offers a relevant performance assessment methodology but also presents some drawbacks for distributional RL. Indeed, the environments are mostly deterministic and require a tremendous amount of computational power. Since the original publication of the Atari-57 benchmark, diverse evaluation methodologies have progressively appeared. In this research work, the best practices proposed by the article \cite{Machado2018} are adopted. Moreover, the mostly deterministic transitions within Atari games are made stochastic by using the sticky action generalisation technique. This last addition makes the Atari environments from the present work slightly more complex compared to the ones from previous publications in distributional RL.\\

The last type of benchmark environment is a set of \textit{five MinAtar games} \cite{Young2019}: Asterix, Breakout, Freeway, Seaquest and SpaceInvaders. These environments are miniaturised and slightly simplified versions of several Atari games representative of the complete Atari-57 suite. The core objective behind these MinAtar environments is to make RL experimentation around Atari games more accessible and efficient. Moreover, this alternative benchmark is also promoted by the same article \cite{Obando2021} as a replacement for the Atari-57 benchmark in order to achieve more inclusive DRL research.\\

More information about these benchmark environments is provided in \ref{AppendixEnvironments}. To end this section about the performance assessment methodology, an argument for bypassing the complete Atari-57 benchmark generally adopted in research works about distributional RL is presented. As previously hinted, the computational cost associated with this particular benchmark is significant. In this case, two entire weeks’ worth of computations are required for training one RL agent on a single Atari game using the UMDQN algorithm with hardware acceleration enabled (NVIDIA RTX 2080 Ti). Therefore, running this novel distributional RL algorithm for five different random seeds on the complete Atari-57 benchmark would approximately require $57*5*14 \simeq 4000$ days when having access to a single GPU, without even considering the hyperparameters tuning phase. It naturally becomes totally impracticable without parallelisation with numerous GPUs. Although the UMDQN algorithm presents the drawback of being slightly more computationally expensive compared to the state-of-the-art distributional RL algorithms, the previous conclusion remains in line with findings from the scientific literature. For instance, the simpler DQN algorithm takes roughly $1425$ days to fully train for each Atari game using specialised hardware (NVIDIA Tesla P100) \cite{Obando2021}. Because this problem creates a real barrier to entry for modest laboratories having access to a limited amount of computational power, it is repeatedly discussed by the RL research community. For this reason, the present research work adopts a different yet insightful set of benchmark environments for evaluating distributional RL algorithms, based on the article \cite{Obando2021} for more inclusive DRL research.\\

\subsection{Results discussion}
\label{SectionResultsDiscussion}

\begin{figure}[H]
    \centering
    \includegraphics[width=1\linewidth, trim={4.1cm 6.2cm 8.5cm 6cm}, clip]{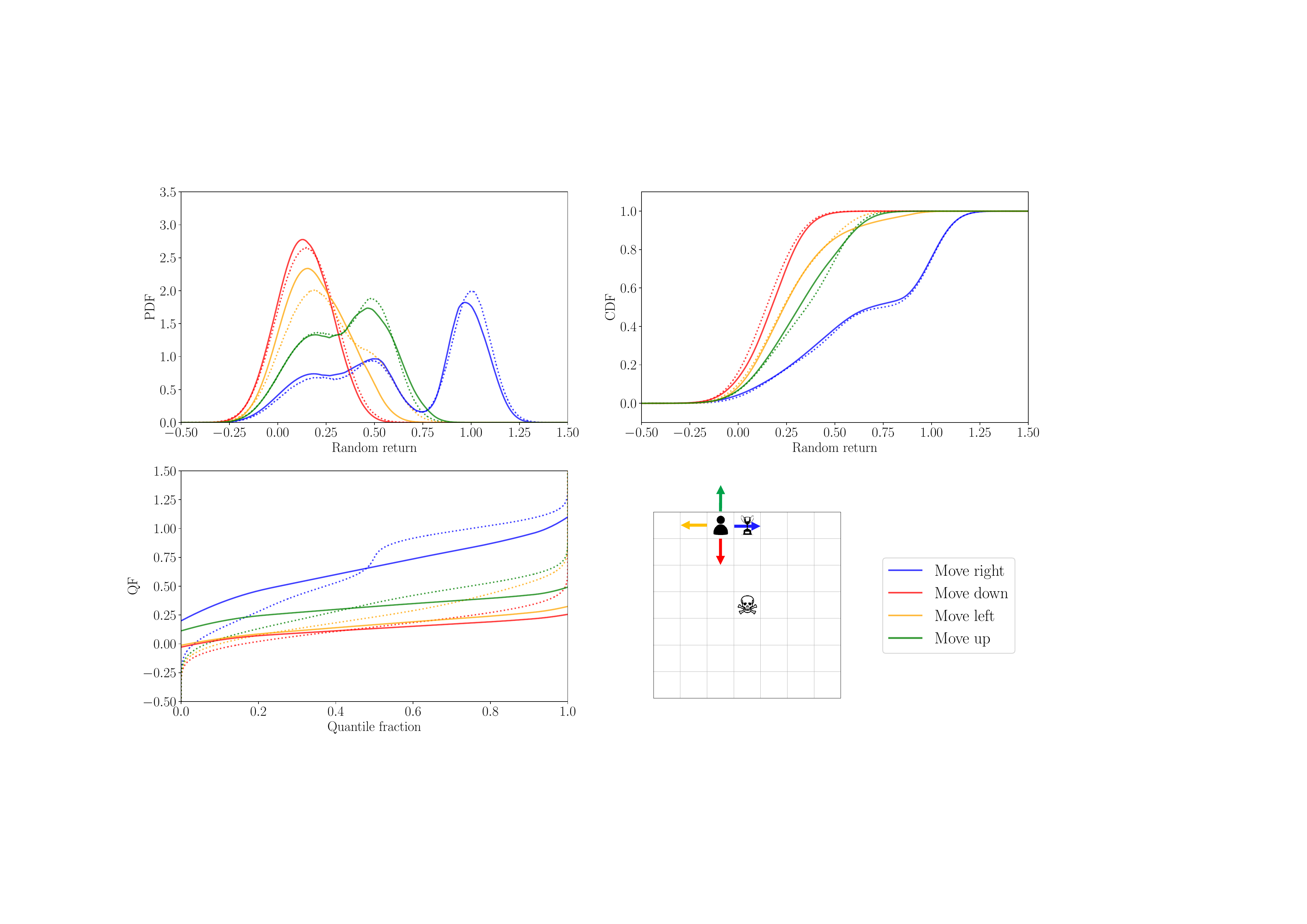}
    \caption{Comparison of the random return distributions (PDF, CDF and QF) learnt by the UMDQN algorithm (plain lines) with the true random return probability distributions (PDF, CDF and QF) estimated via Monte Carlo and associated with an optimal policy (dotted lines), for a particular state of the stochastic grid world environment.}
    \label{ResultsDistributions}
\end{figure}

\begin{figure}[H]
    \centering
    \includegraphics[width=0.95\linewidth, trim={5.5cm 5.0cm 6.8cm 5.0cm}, clip]{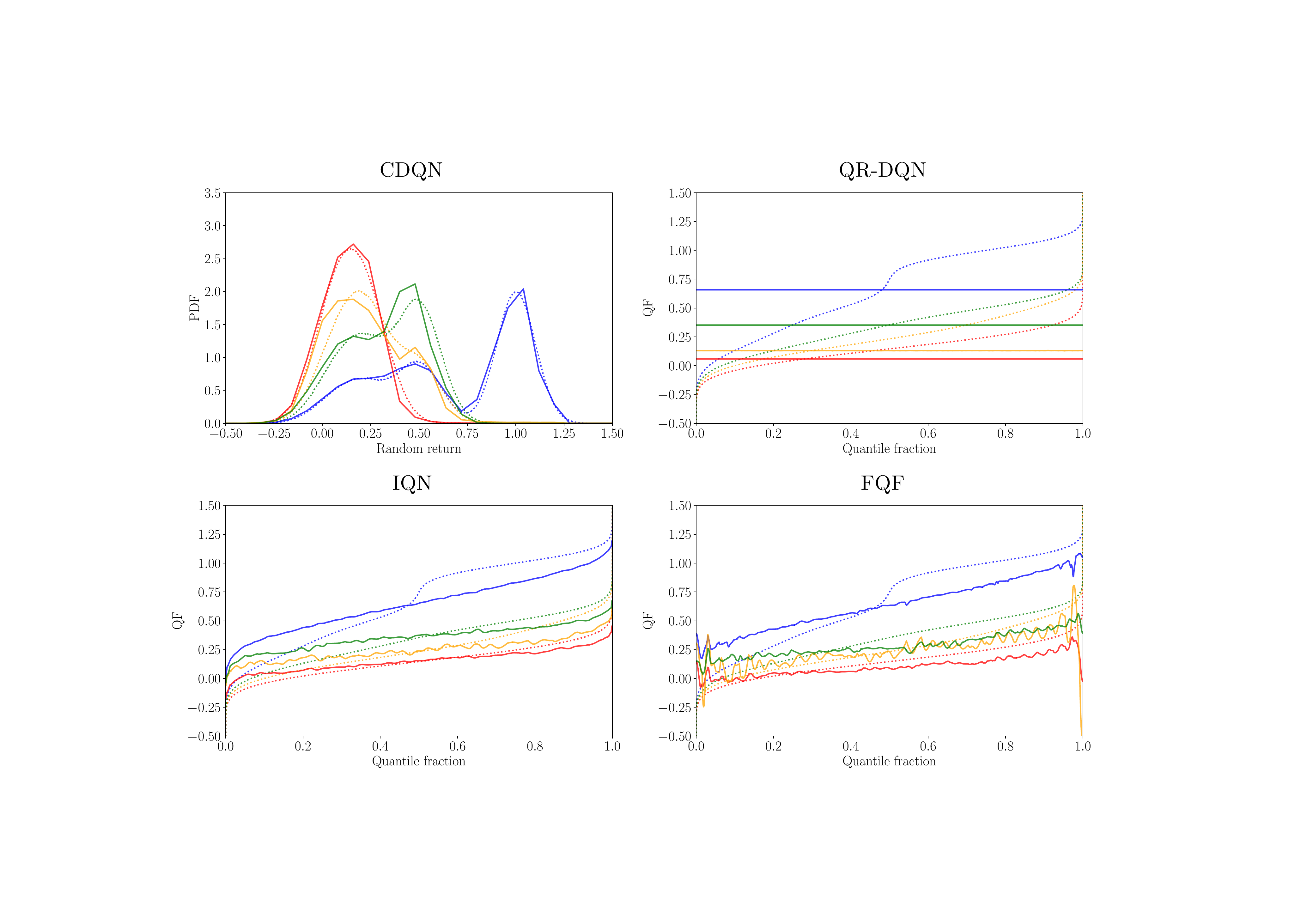}
    \caption{Comparison of the random return distributions learnt by the CDQN, QR-DQN, IQN and FQF state-of-the-art algorithms (plain lines) with the true random return probability distributions estimated via Monte Carlo and associated with an optimal policy (dotted lines), for a particular state of the stochastic grid world environment.}
    \label{ResultsDistributionsBis}
\end{figure}

\paragraph{\textbf{Random return distribution visualisation}} Besides the evaluation of the resulting policy performance, it is important to assess the correctness of the probability distributions learnt by a distributional RL algorithm. As previously hinted, this particular analysis is performed on the stochastic grid world environment. Since the underlying control problem is relatively easy to solve from a human perspective, an optimal policy can be manually derived for that specific environment. That property significantly eases the assessment of the soundness of the random return probability distributions learnt. Once the optimal policy is available, the true probability distributions of the random return can be effectively estimated via Monte Carlo. The same operation should also be performed with the policy learnt by the distributional RL algorithm, since incorrect probability distributions could also be caused by suboptimal policies. Based on that methodology, Figure \ref{ResultsDistributions} graphically compares the probability distributions learnt by the three versions of the UMDQN algorithm (PDF, CDF and QF) with the true random return distributions associated with an optimal policy for a particular state of the environment. Although the PDF and CDF of the random return learnt by the UMDQN-KL and UMDQN-C algorithms are not entirely correct, they remain qualitatively very similar to the true random return distribution, with the multimodality preserved (see blue line). This observation not only validates the soundness of the probability distributions learnt, but also indicates that these two distributional RL algorithms can effectively learn an optimal policy for this benchmark environment. On the contrary, the error made by the UMDQN-W algorithm learning the QF of the random return is much more concerning. In this case, the distributions multimodality is no longer preserved (see blue line). Additional analyses reveal that this difference does not originate from a suboptimal policy learnt by the distributional RL algorithm. In fact, this important observation is consistent with Equation \eqref{EquationLossQFBis} together with the explanation from Section \ref{SectionAbstraction} regarding the learning of the QF based on the distributional Bellman operator: the expectation of the random return is preserved, but the probability distribution higher-order moments are not. As previously explained, this analysis is not specific to the UMDQN-W algorithm and applies to several state-of-the-art distributional RL algorithms. To illustrate that claim, Figure \ref{ResultsDistributionsBis} plots the probability distributions of the random return learnt by the CDQN, QR-DQN, IQN and FQF algorithms, which all achieve an optimal policy for the stochastic grid world environment. On the one hand, the CDQN algorithm learning the categorical PDF of the random return based on the KL divergence achieves satisfying results, in line with the previous observation for the UMDQN-KL algorithm. On the other hand, the QR-DQN, IQN and FQF algorithms clearly show their limitations for accurately modelling the QF of the random return. Therefore, this particular learning methodology adopted by several state-of-the-art distributional RL algorithms should only be considered when the objective is to learn policies maximising the expectation of the random return, but should instead be discarded when the intention is to exploit the complete probability distribution, for learning risk-aware policies for instance.\\

\begin{figure}
    \centering
    \begin{subfigure}[b]{0.328\textwidth}
        \centering
        \includegraphics[width=1\linewidth, trim={0.25cm 0cm 2.1cm 1cm}, clip]{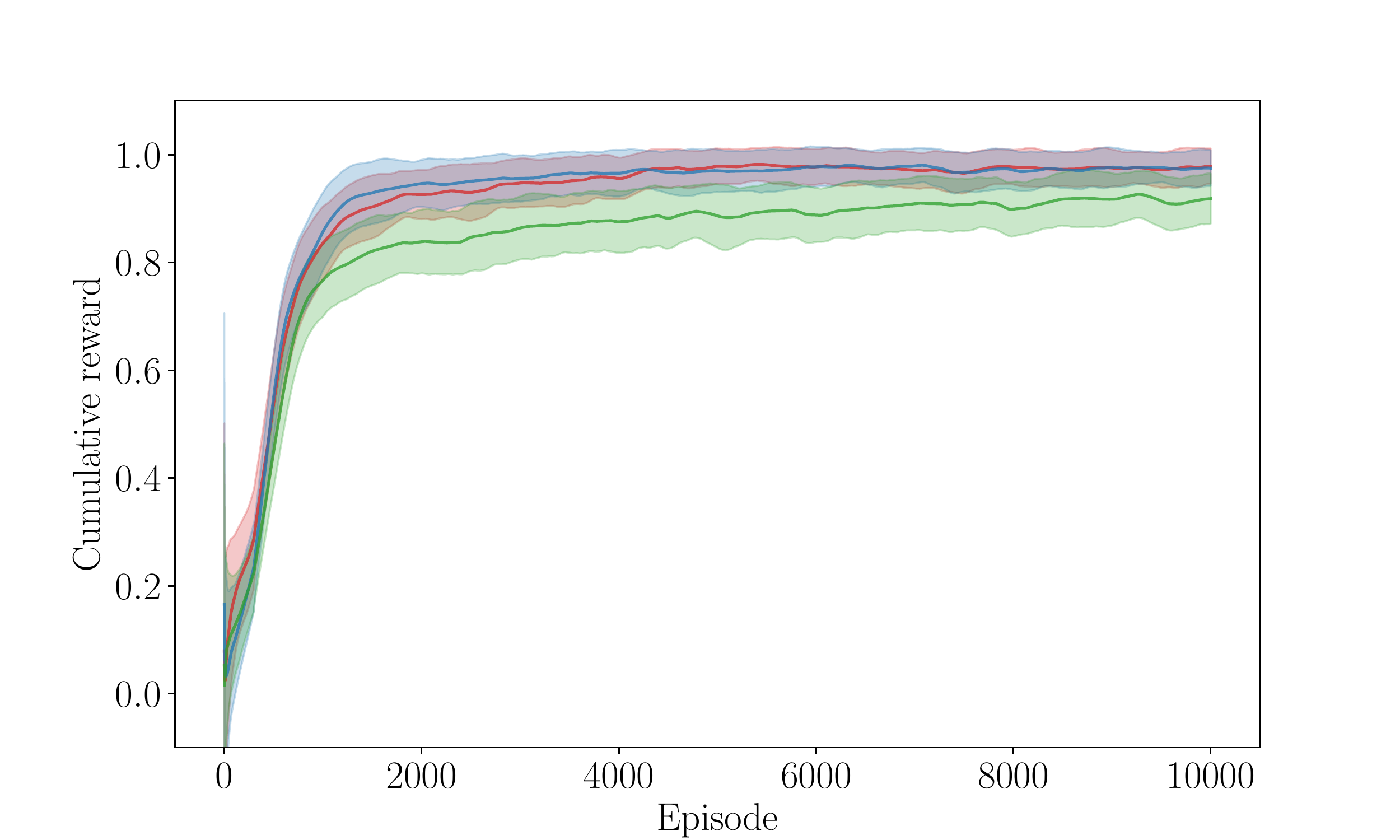}
        \caption{Stochastic grid world}
        \label{ResultsStochasticGridWorld}
    \end{subfigure}
    \hfill
    \begin{subfigure}[b]{0.328\textwidth}
        \centering
        \includegraphics[width=1\linewidth, trim={0.25cm 0cm 2.1cm 1cm}, clip]{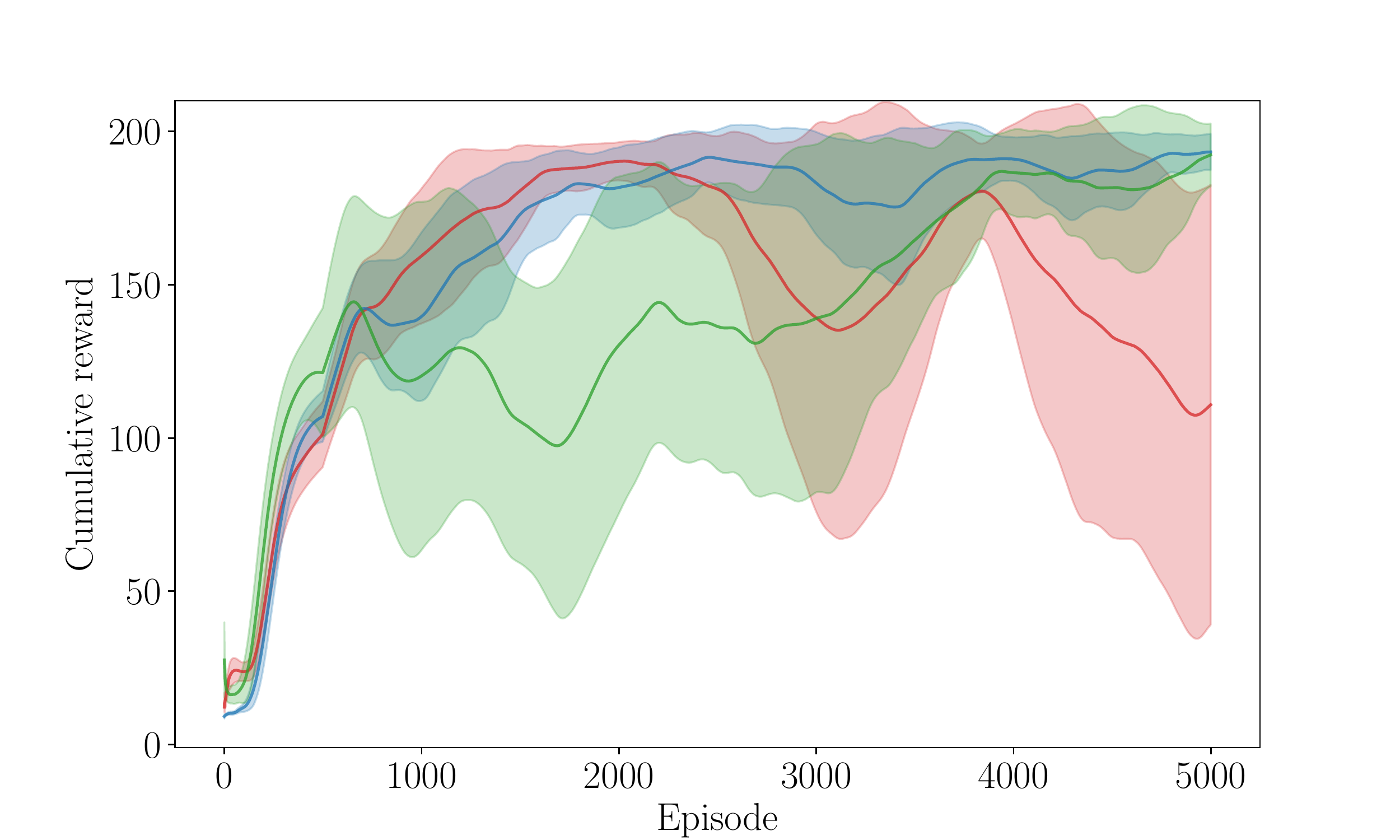}
        \caption{CartPole}
        \label{ResultsCartPole}
    \end{subfigure}
    \hfill
    \begin{subfigure}[b]{0.328\textwidth}
        \centering
        \includegraphics[width=1\linewidth, trim={0.25cm 0cm 2.1cm 1cm}, clip]{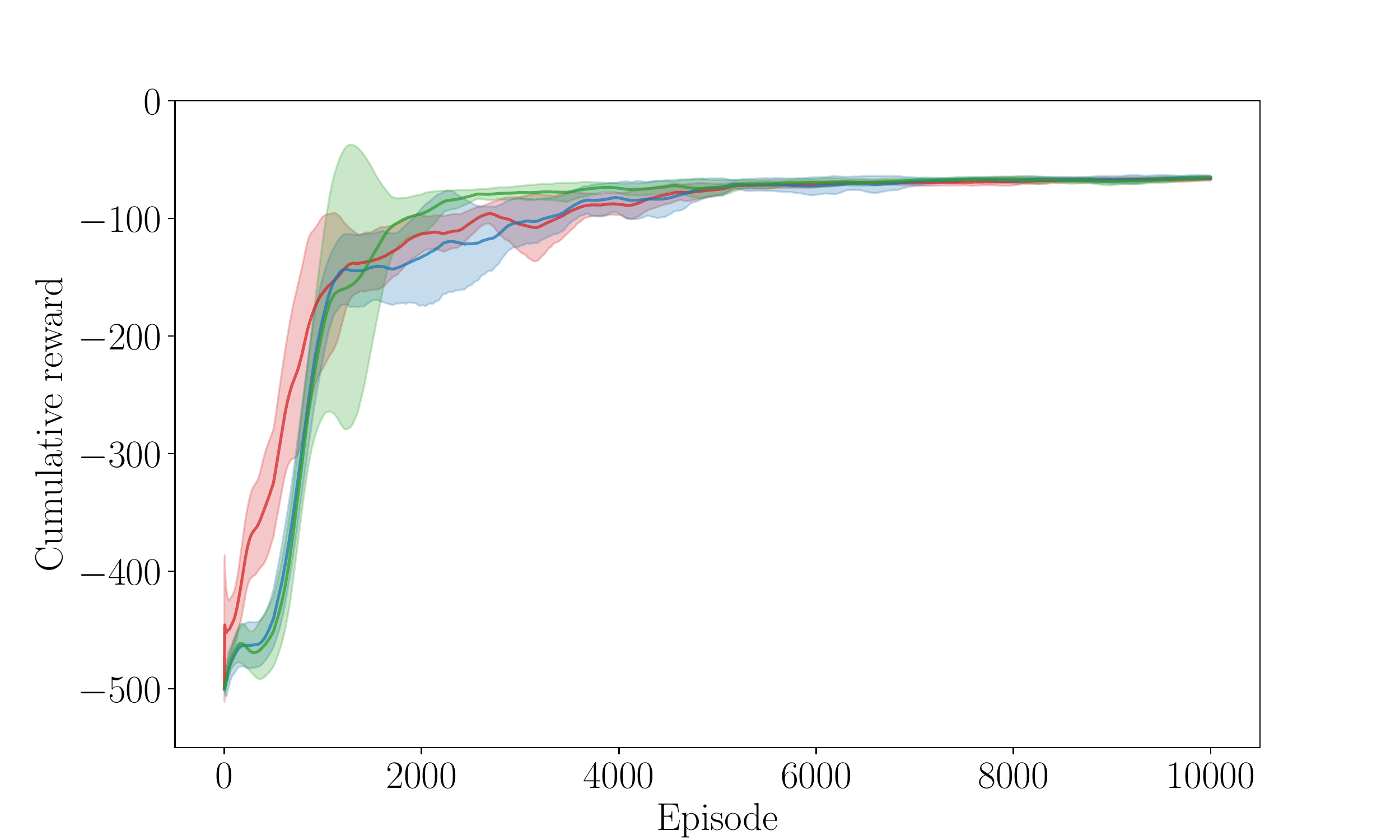}
        \caption{Acrobot}
        \label{ResultsAcrobot}
    \end{subfigure}
    \hfill
    \begin{subfigure}[b]{0.328\textwidth}
        \centering
        \includegraphics[width=1\linewidth, trim={0.25cm 0cm 2.1cm 1cm}, clip]{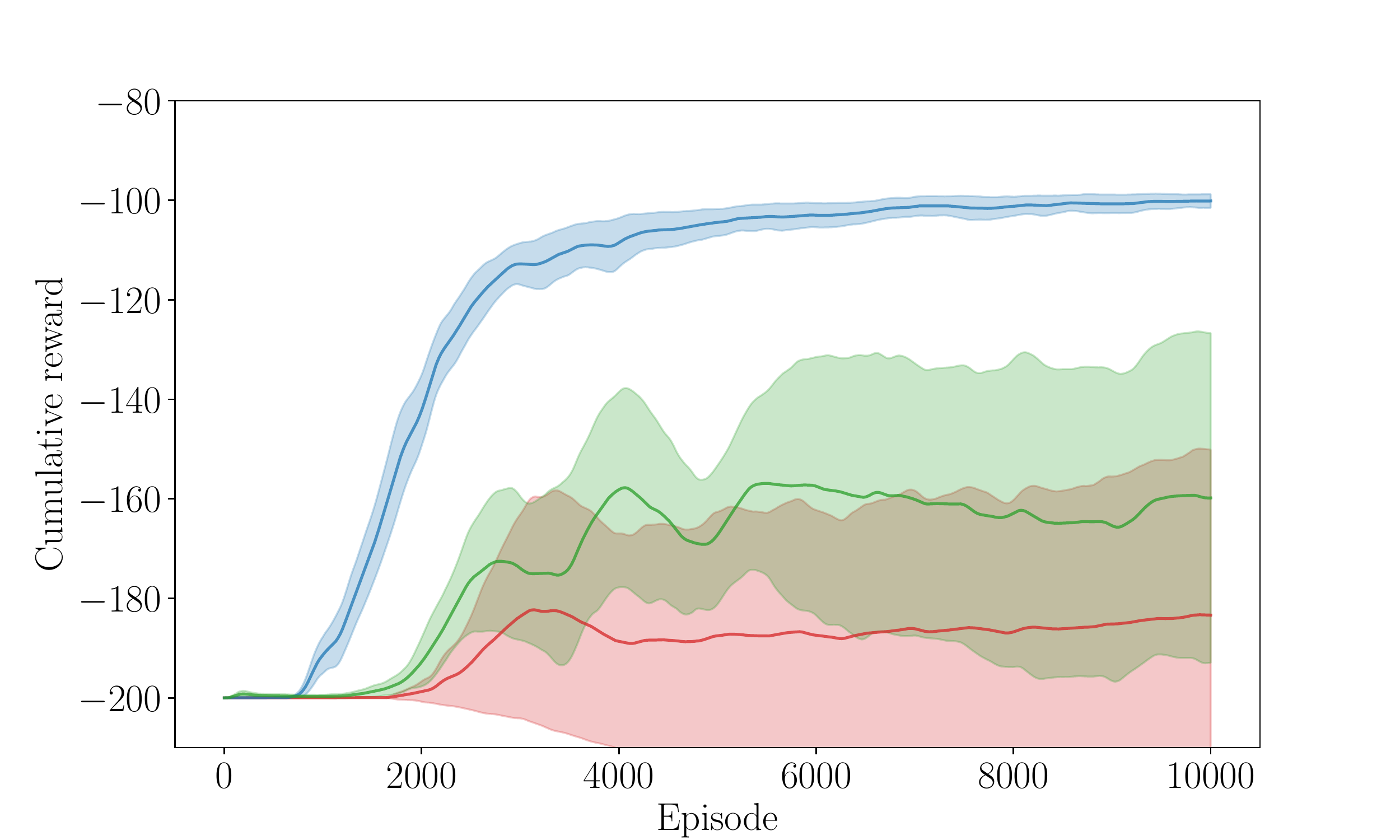}
        \caption{MountainCar}
        \label{ResultsMountainCar}
    \end{subfigure}
    \hfill
    \begin{subfigure}[b]{0.328\textwidth}
        \centering
        \includegraphics[width=1\linewidth, trim={0.25cm 0cm 2.1cm 1cm}, clip]{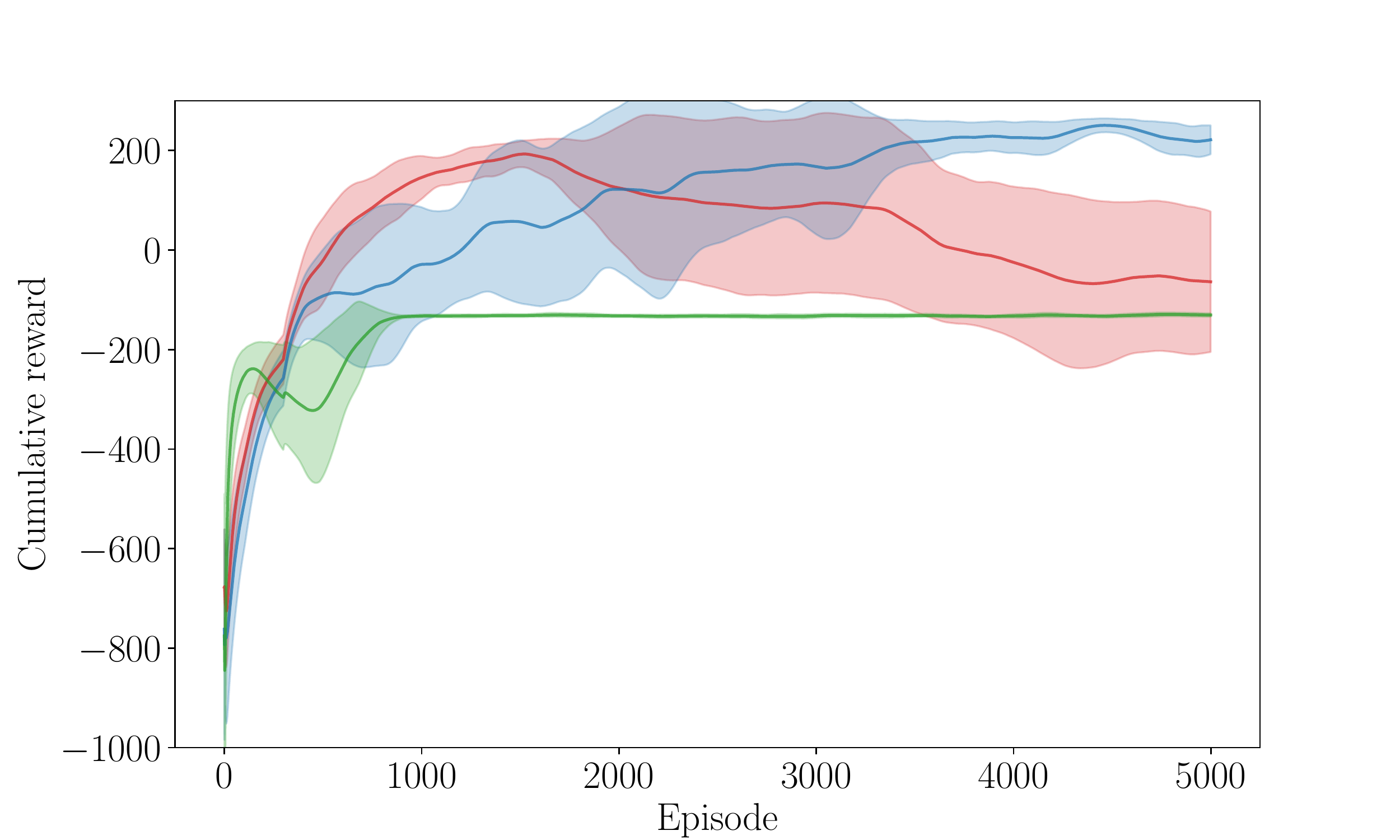}
        \caption{LunarLander}
        \label{ResultsLunarLander}
    \end{subfigure}
    \hfill
    \begin{subfigure}[b]{0.328\textwidth}
        \centering
        \includegraphics[width=1\linewidth, trim={0.25cm 0cm 2.1cm 1cm}, clip]{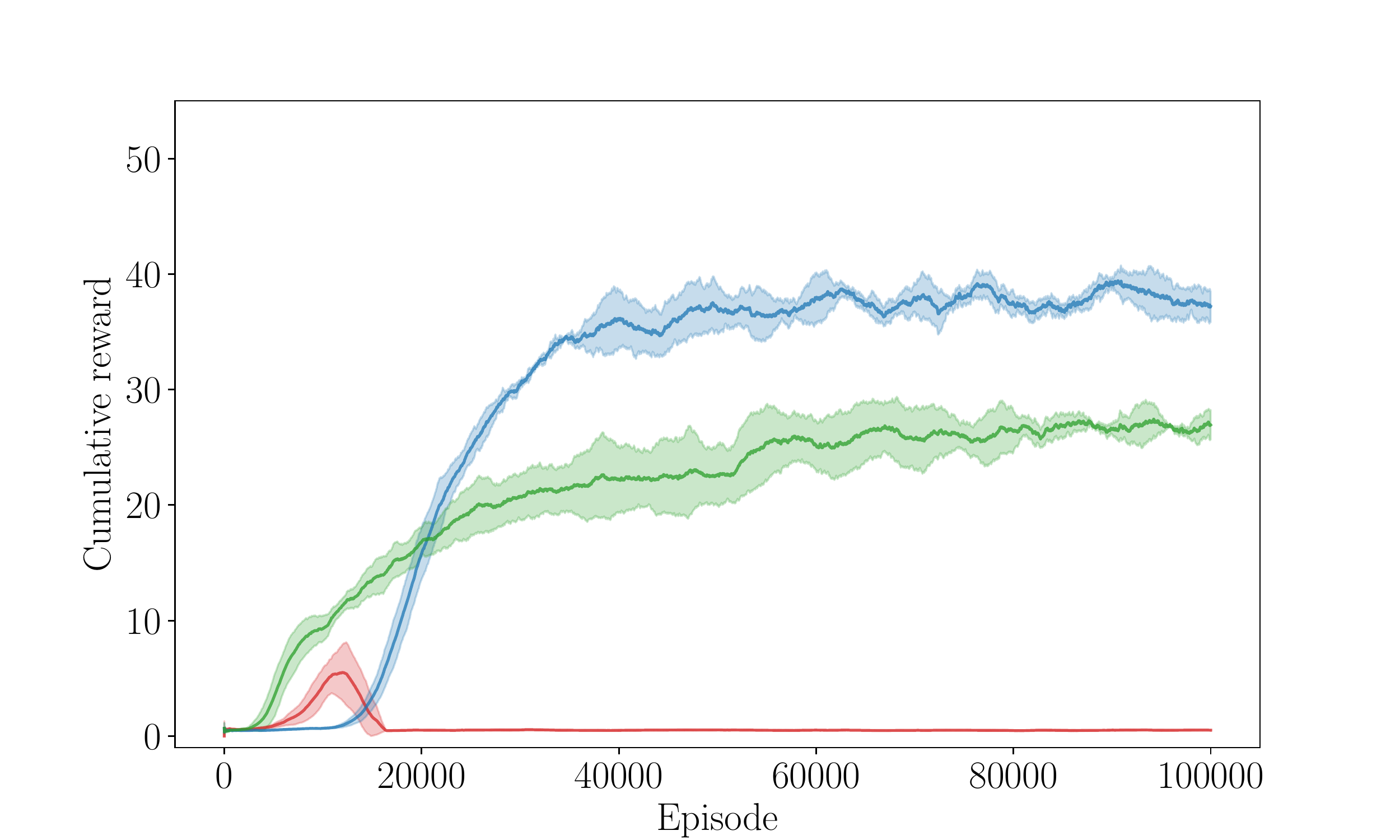}
        \caption{MinAtar Asterix}
        \label{ResultsMinAtarAsterix}
    \end{subfigure}
    \hfill
    \begin{subfigure}[b]{0.328\textwidth}
        \centering
        \includegraphics[width=1\linewidth, trim={0.25cm 0cm 2.1cm 1cm}, clip]{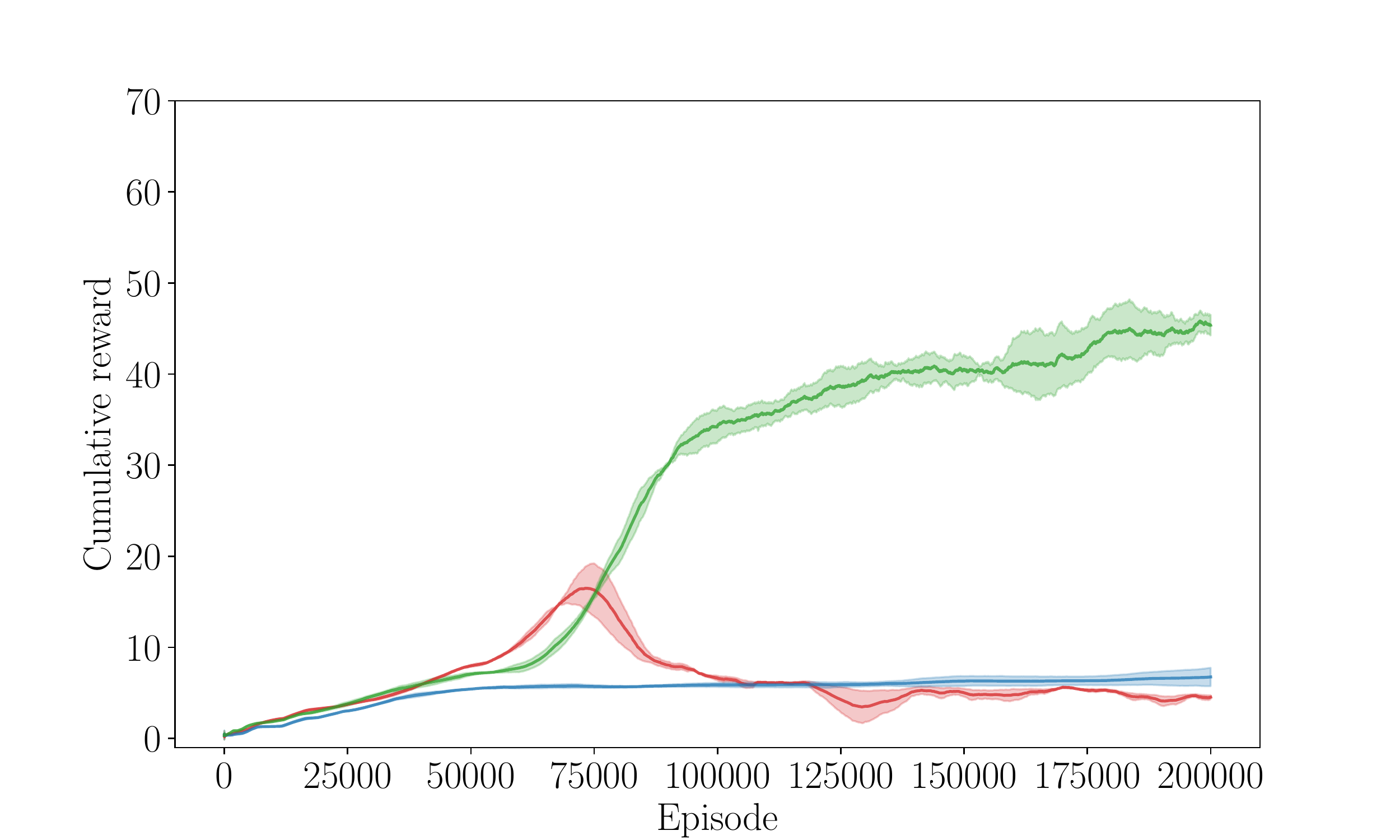}
        \caption{MinAtar Breakout}
        \label{ResultsMinAtarBreakout}
    \end{subfigure}
    \hfill
    \begin{subfigure}[b]{0.328\textwidth}
        \centering
        \includegraphics[width=1\linewidth, trim={0.25cm 0cm 2.1cm 1cm}, clip]{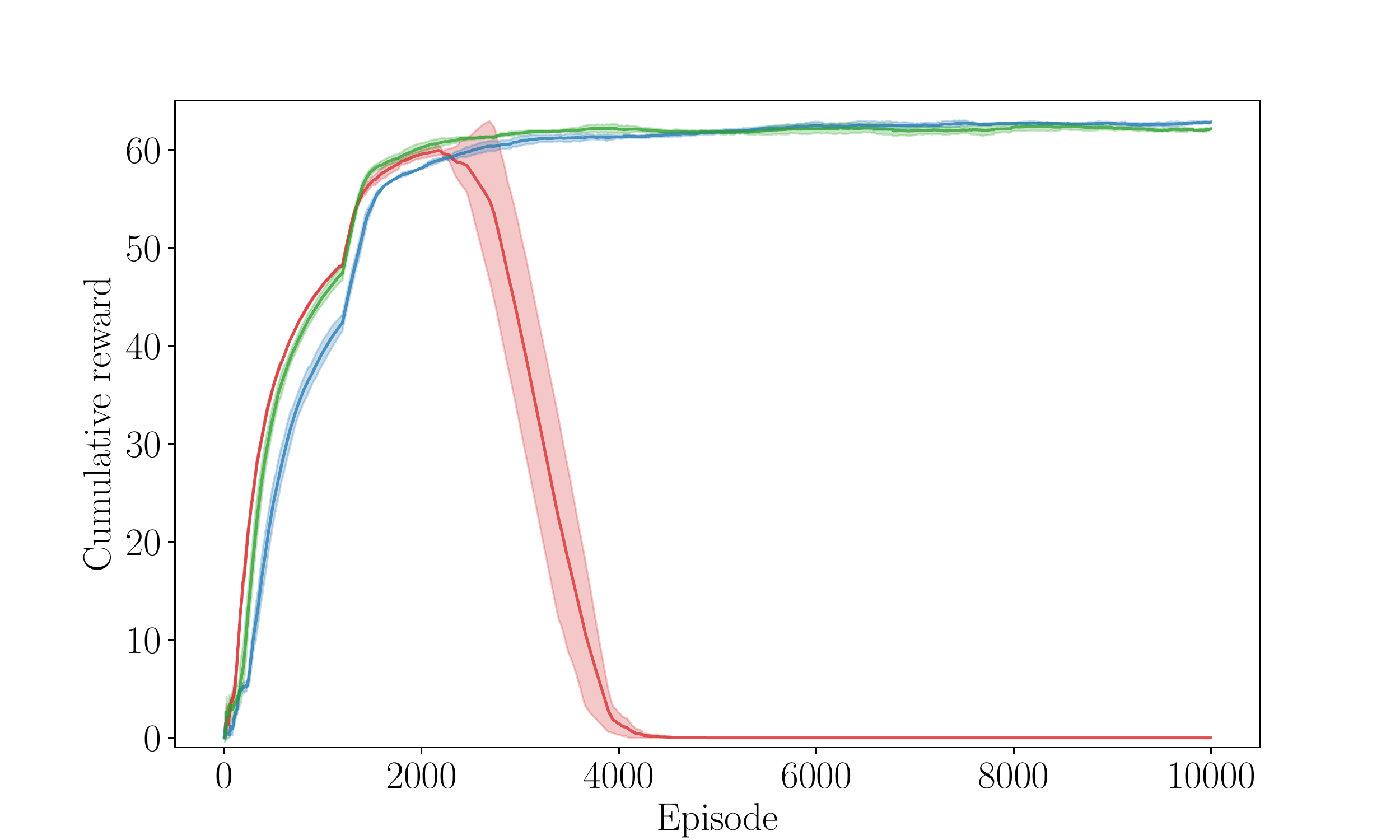}
        \caption{MinAtar Freeway}
        \label{ResultsMinAtarFreeway}
    \end{subfigure}
    \hfill
    \begin{subfigure}[b]{0.328\textwidth}
        \centering
        \includegraphics[width=1\linewidth, trim={0.25cm 0cm 2.1cm 1cm}, clip]{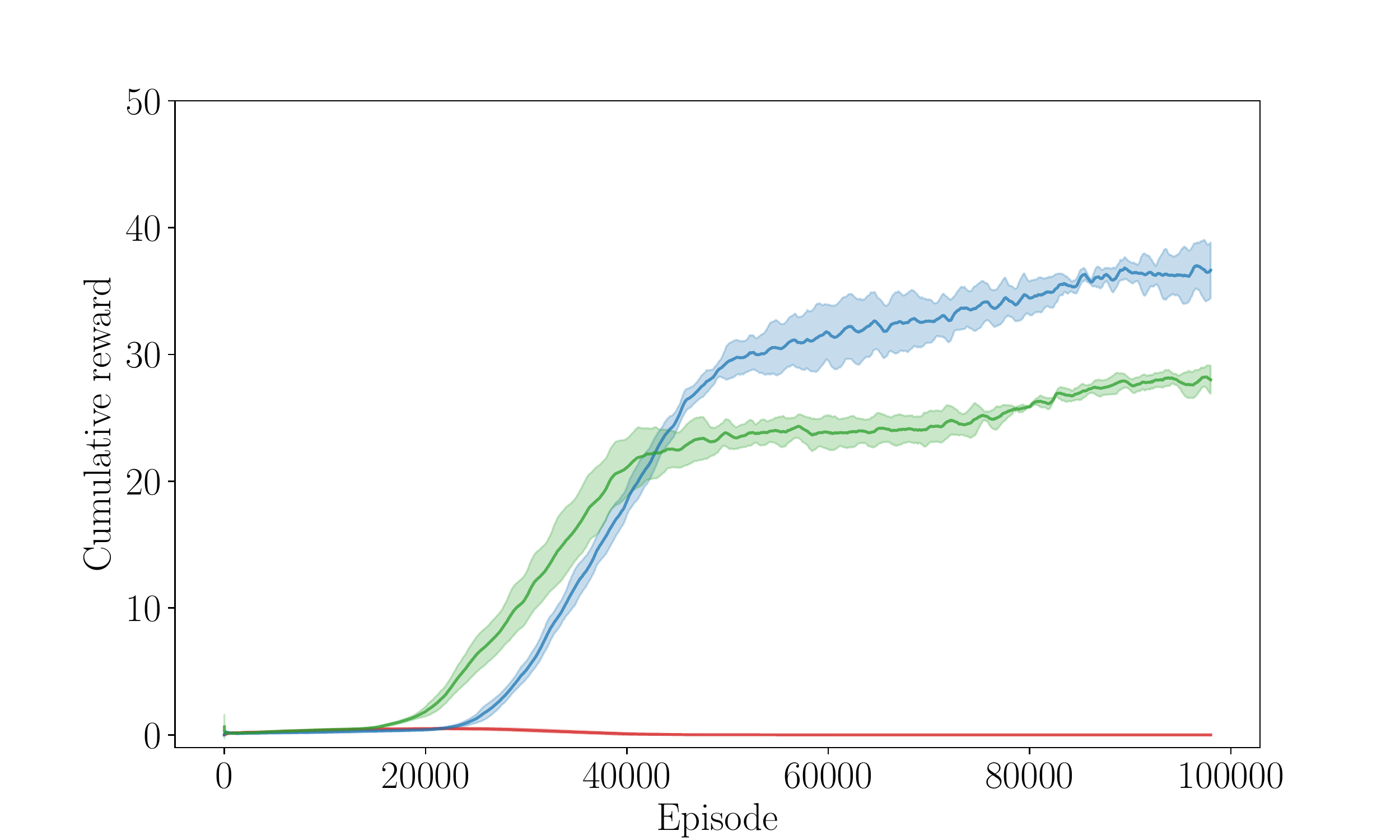}
        \caption{MinAtar Seaquest}
        \label{ResultsMinAtarSeaquest}
    \end{subfigure}
    \hfill
    \begin{subfigure}[b]{0.328\textwidth}
        \centering
        \includegraphics[width=1\linewidth, trim={0.25cm 0cm 2.1cm 1cm}, clip]{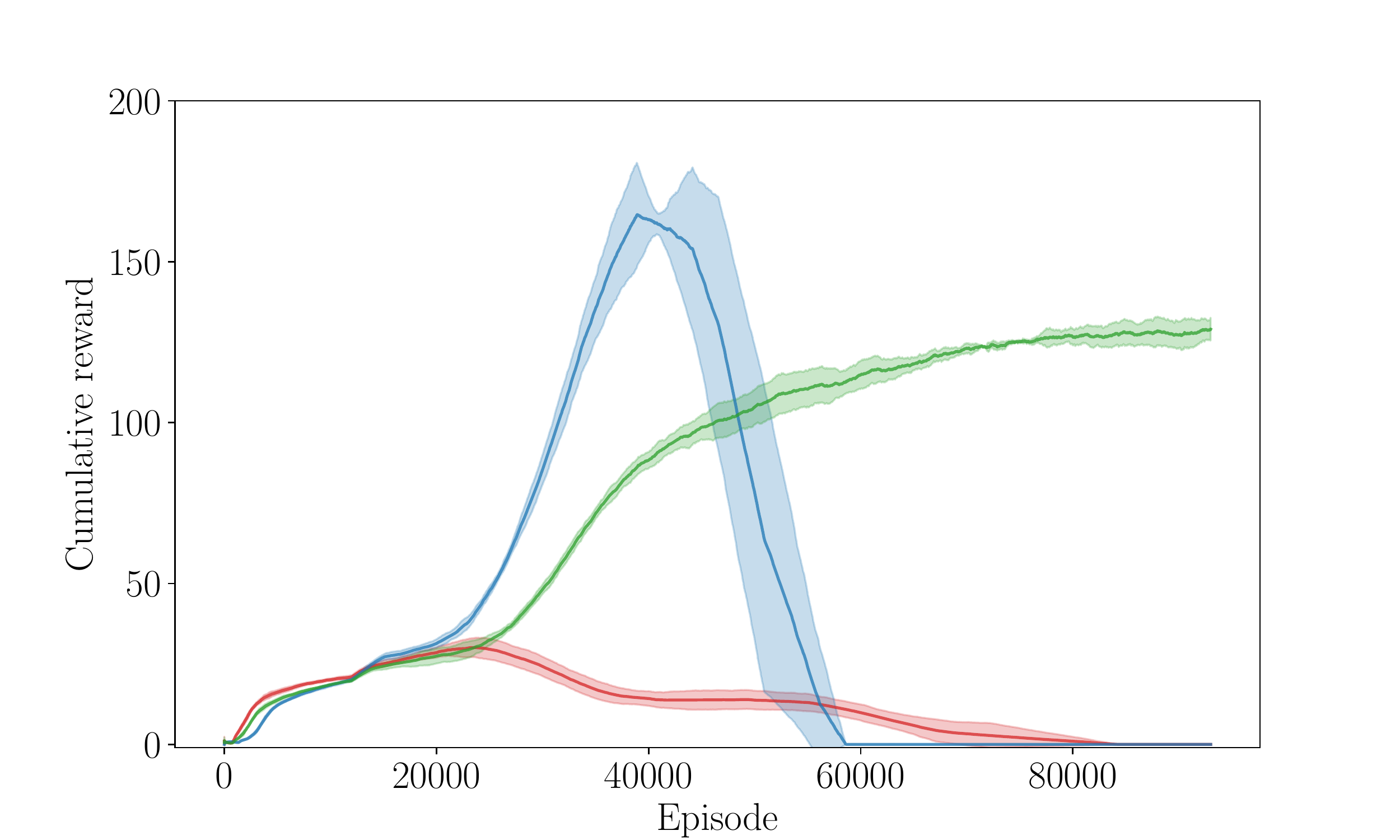}
        \caption{MinAtar SpaceInvaders}
        \label{ResultsMinAtarSpaceInvaders}
    \end{subfigure}
    \hfill
    \begin{subfigure}[b]{0.328\textwidth}
        \centering
        \includegraphics[width=1\linewidth, trim={0.25cm 0cm 2.1cm 1cm}, clip]{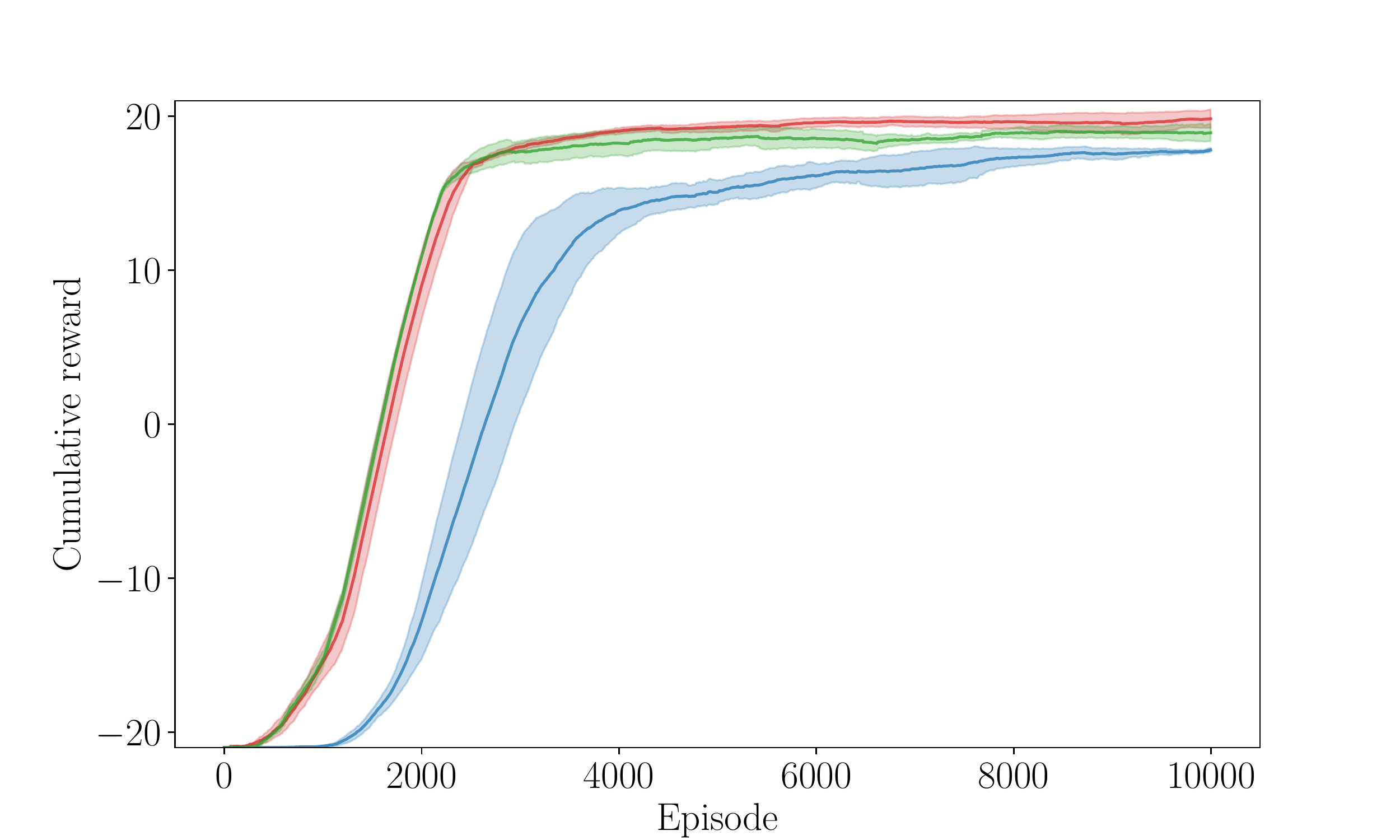}
        \caption{Atari Pong}
        \label{ResultsAtariPong}
    \end{subfigure}
    \hfill
    \begin{subfigure}[b]{0.328\textwidth}
        \centering
        \includegraphics[width=1\linewidth, trim={0.25cm 0cm 2.1cm 1cm}, clip]{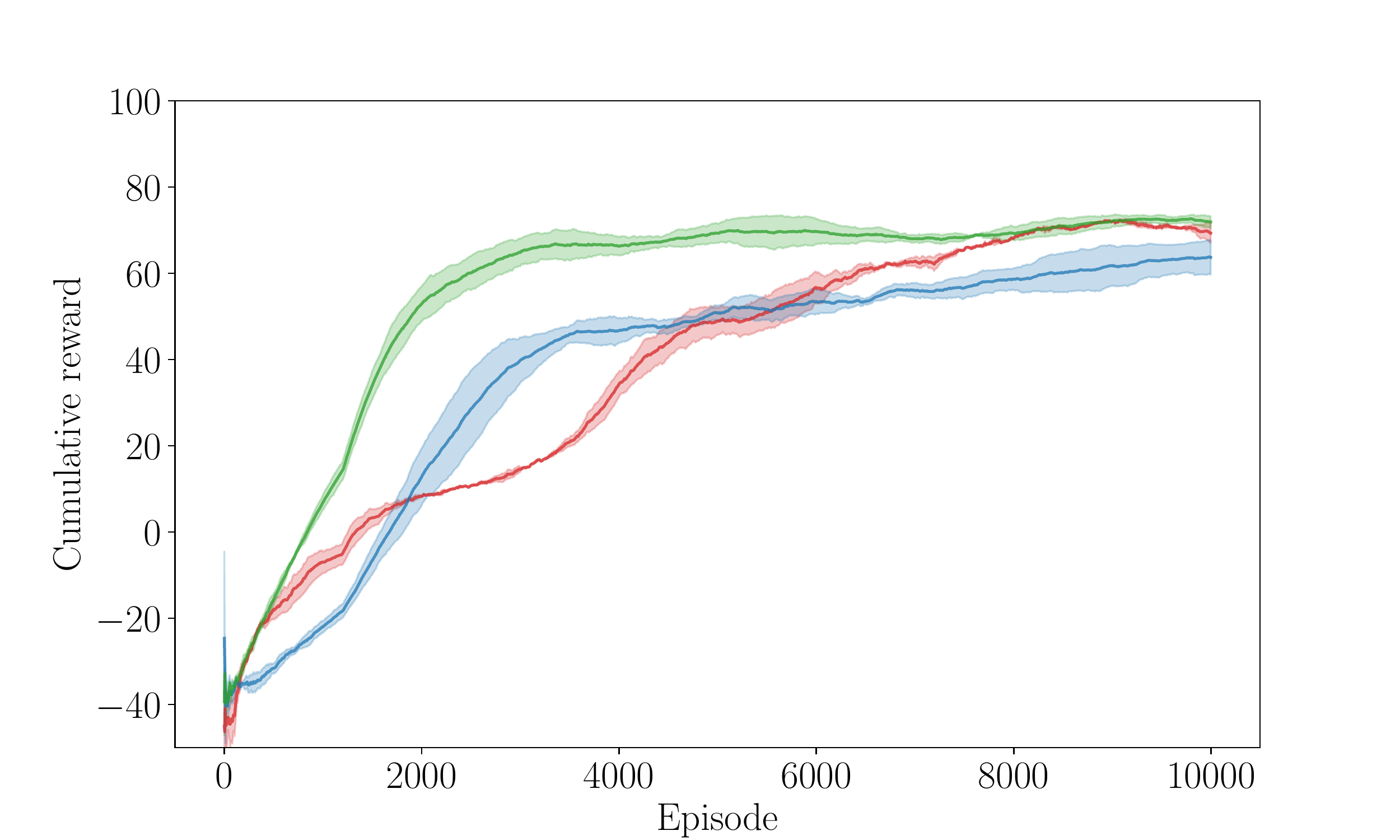}
        \caption{Atari Boxing}
        \label{ResultsAtariBoxing}
    \end{subfigure}
    \hfill
    \begin{subfigure}[b]{0.328\textwidth}
        \centering
        \includegraphics[width=1\linewidth, trim={0.25cm 0cm 2.1cm 1cm}, clip]{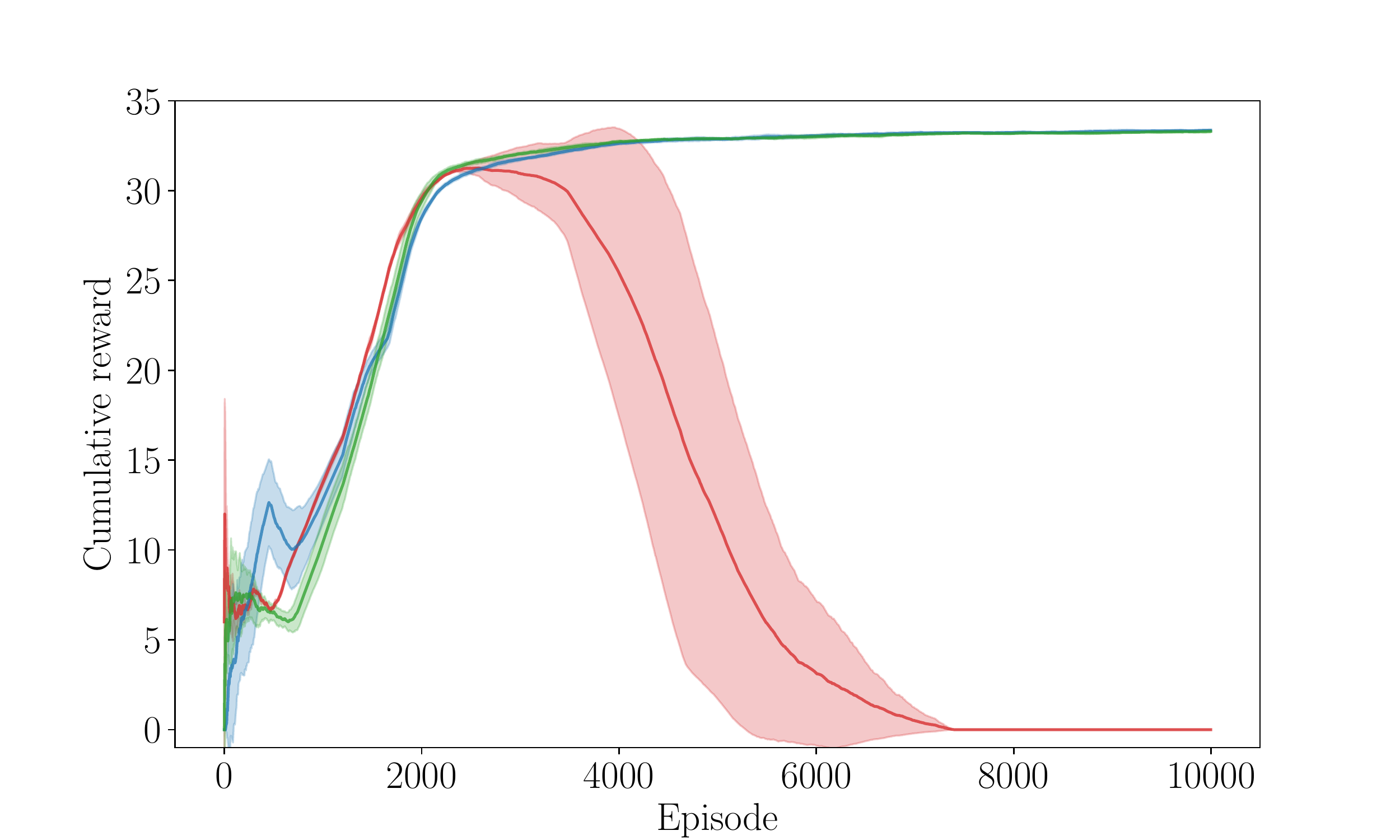}
        \caption{Atari Freeway}
        \label{ResultsAtariFreeway}
    \end{subfigure}
    \hfill
    \begin{subfigure}[b]{0.7\textwidth}
        \centering
        \includegraphics[width=1\linewidth, trim={1.5cm 13.6cm 1.5cm 12cm}, clip]{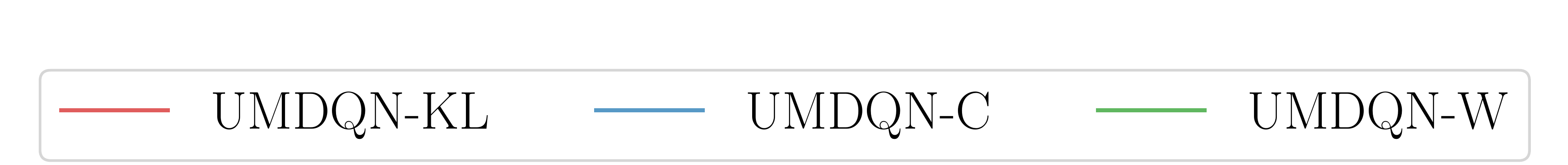}
        \label{Legend}
    \end{subfigure}
    \caption{Performance of the UMDQN algorithm on the benchmark environments proposed.}
    \label{ResultsUMDQN}
\end{figure}

\paragraph{\textbf{Policy performance}} As far as the quality of the decision-making policies learnt by the distributional RL algorithms is concerned, Figure \ref{ResultsUMDQN} presents the results achieved by the three versions of the UMDQN algorithm on the benchmark environments introduced in Section \ref{SectionBenchmarkEnvironments}. The policy performance plotted is the cumulative reward achieved by the RL agent over one episode. For the sake of reliability, the results are averaged over five different random seeds and the variance is highlighted. Moreover, for improved readability, a moving average operation is performed to further smooth the curves. Taking into account their respective strengths and weaknesses detailed below, it is quite difficult to identify a clear winner overall in terms of policy performance among the three versions of the UMDQN algorithm, even though the UMDQN-KL algorithm lags behind the other two. Since the same function approximator class is used, this conclusion also stands for the distribution representations and the probability metrics underneath the distributional RL algorithms. An argument for potentially explaining this observation is the fact that a neural network may more efficiently model the PDF, CDF or QF of the random return distributions depending on the characteristics of these particular probability distributions (multimodality, values of the moments). Another hypothesis is to point out the approximation of the loss defined in Equation \eqref{EquationLoss}, whose effect is not yet clearly understood for the different distribution representations and probability metrics, but also potentially depending on the control problem. This is an important open research question for distributional RL. Therefore, based on these observations, the distribution representation of the random return together with the probability metric should ideally be hyperparameters to be tuned depending on the environment and the control problem at hand. This claim contrasts with the current trend observed in distributional RL research, with the focus being mainly set on the QF and Wasserstein distance, as illustrated by the recent QR-DQN, IQN and FQF algorithms. For this reason, the present research work calls for a reconsideration of all distribution representations and probability metrics for future research in distributional RL.\\

\paragraph{\textbf{UMDQN-KL algorithm}} Even though the distributional Bellman operator $\mathcal{T}^{\pi}$ is not a contraction mapping in the KL divergence, Figures \ref{ResultsDistributions} and \ref{ResultsUMDQN} empirically show that this probability metric can still lead to the learning of both valuable decision-making policies and relevant random return probability distributions. This observation suggests that the contraction property is not a necessary condition for converging towards the correct random return probability distribution. Nevertheless, the learning process of the UMDQN-KL algorithm has been observed to be fairly less stable compared to other distributional RL algorithms. For several benchmark environments, the learning process may even suddenly stop with the performance entirely collapsing and not recovering, as illustrated in Figure \ref{ResultsAtariFreeway}. Additional experiments suggest that the occurrence of this problematic behaviour for a given environment is strongly tied to the domain $\mathcal{X}$ specified as hyperparameter (lower and upper bounds). A too restrained domain inevitably leads to truncated and hence wrong probability distributions for the random return $Z^{\pi}$. On the contrary, if the domain is too wide, it may lead to numerical instabilities in the regions of the domain with almost no mass due to the definition of the KL divergence ($\lim_{x \rightarrow 0^{+}} \log(x) = -\infty$). Appropriately setting the hyperparameters associated with this domain $\mathcal{X}$ may be a particularly challenging task since it is strongly dependent on the control problem and because the probability distribution of the random return $Z^{\pi}$ may significantly vary with different state-action pairs $(s, a)$ as well as during the learning process. Another interesting observation about the UMDQN-KL algorithm is related to the asymmetry of the KL divergence ($\mathcal{L}_{KL}(A, B) \neq \mathcal{L}_{KL}(B, A)$). Empirically, the learning of valuable policies is observed with the loss $\mathcal{L}_{KL}(\mathcal{T}^{\pi} Z^{\pi}, Z^{\pi})$ but not with $\mathcal{L}_{KL}(Z^{\pi}, \mathcal{T}^{\pi} Z^{\pi})$.\\

\paragraph{\textbf{UMDQN-C algorithm}} Although this distributional RL algorithm also requires the specification of hyperparameters associated with the domain $\mathcal{X}$, it is empirically observed to be far more stable and performing compared to the UMDQN-KL algorithm. This behaviour may potentially be explained by the distributional Bellman operator $\mathcal{T}^{\pi}$ being a contraction in the Cramer distance, but also by the fact that the loss to learn from is symmetric and does not numerically explode around regions of the domain with no probability density. Still, a relevant domain $\mathcal{X}$ has to be specified to expect reliable and satisfying results from the UMDQN-C algorithm, meaning that the range of the returns has to be rigorously approximated beforehand. This requirement is the main weakness of this particular distributional RL algorithm. In Figure \ref{ResultsUMDQN}, relevant domains $\mathcal{X}$ are adopted to ensure a fair and interesting comparison. The UMDQN-C algorithm may be the top-performing approach at first glance, but its performance inevitably decreases with less accurate domains.\\

\paragraph{\textbf{UMDQN-W algorithm}} Regarding the performance of the learnt policy, this distributional RL algorithm may probably be the most versatile of the UMDQN algorithms, for two reasons. Firstly, the distributional Bellman operator $\mathcal{T}^{\pi}$ is a contraction mapping in the Wasserstein distance. Secondly, learning the QF of the random return $Z^{\pi}$ does not require the challenging specification of the returns domain $\mathcal{X}$, since the QF takes inputs bounded in the range $[0, 1]$. However, when it comes to the accuracy of the probability distributions learnt, the UMDQN-W algorithm is no longer an acceptable solution, as previously explained in this section. Consequently, this distributional RL algorithm should only be considered for learning decision-making policies maximising the expectation of the random return, but not exploiting the full probability distributions.\\

For the sake of reproducibility, the complete code used for generating the results presented in this section is made publicly available at the following link: \url{https://github.com/ThibautTheate/Unconstrained-Monotonic-Deep-Q-Network-algorithm}. Moreover, the hyperparameters are provided in \ref{AppendixParameters}. To end this section, \ref{AppendixResults} briefly compares the policy performance achieved by the UMDQN algorithm with that of the state-of-the-art distributional RL algorithms on the benchmark environments, even though such a comparison is not an objective of this research work. In short, the figure suggests that the results achieved are comparable, which reinforces the soundness of the proposed approach.\\

\section{Conclusions}
\label{SectionConclusions}

This research work introduces the \textit{unconstrained monotonic deep Q-network} (UMDQN) distributional RL algorithm, by combining a novel methodology for learning the probability distribution of the random return independently of its representation and the UMNN architecture for modelling these distributions. The experiments performed take advantage of some interesting properties of this novel distributional RL algorithm to yield three important observations. Firstly, the choice of the probability distribution representation coupled with the probability metric has to ideally be dependent on the control problem, since no clear winner could be identified for the set of benchmark environments studied. This result contrasts with the current trend in distributional RL research, which mainly focuses on the QF and Wasserstein distance. Secondly, the methodology adopted by several state-of-the-art algorithms for learning the QF of the random return involves an important approximation, which results in the learning of inaccurate probability distributions. This approach remains totally sound when attempting to learn decision-making policies maximising the expectation of the random return. On the contrary, it should be discarded when aiming to take advantage of other characteristics of the random return distribution, for instance with risk-aware policies. Thirdly, the contraction mapping property for the distributional Bellman operator is not a necessary condition to learn the correct probability distribution of the random return, but may still be beneficial. This highlights the existing gap between theory and practice in distributional RL, and encourages future research on the distributional Bellman operator as well as on the convergence of distributional RL algorithms in general.\\

To conclude, several avenues are proposed for future work. Firstly, the gap between theory and practice in distributional RL highlighted in this research paper could be narrowed by deriving theoretical guarantees and properties for the novel UMDQN algorithm introduced. Secondly, building on the visualisation and qualitative analysis of the probability distributions presented in this research work, a new performance assessment methodology has to be designed to quantitatively evaluate the accuracy of the random return distributions learnt by a distributional RL algorithm, independently of the resulting policy performance. Indeed, research on distributional RL is generally primarily focused on the latter, neglecting the evaluation of the accuracy of the probability distributions learnt. Thirdly, the approximation in Equation \ref{EquationLossQFBis} for the learning of the QF based on the distributional Bellman equation deserves more research, in order to acquire a better understanding of the problem and potentially find an alternative solution. Fourthly, the performance achieved by the UMDQN algorithm is expected to be significantly improved by implementing the diverse enhancements from the Rainbow algorithm \cite{Hessel2018}: multi-step learning \cite{Sutton1988}, double Q-learning \cite{Hasselt2016}, prioritised experience replay \cite{Schaul2016}, duelling architecture \cite{Wang2016} and noisy networks \cite{Fortunato2018}. Lastly, an interesting evolution of the UMDQN algorithm could be to concurrently manage different distribution representations and probability metrics, and intelligently combine this information to further improve the performance of the newly introduced distributional RL algorithm.\\

\section*{Acknowledgments}

Thibaut Th\'{e}ate, Antoine Wehenkel and Adrien Bolland are Research Fellows of the F.R.S.-FNRS, of which they acknowledge their financial support.

\bibliography{Manuscript.bib}

\clearpage

\appendix

\section{Mathematical proofs}
\label{AppendixDemo}

This section mathematically supports Equations \eqref{EquationLossPDFBis}, \eqref{EquationLossCDFBis} and \eqref{EquationLossQFBis} introduced in Section \ref{SectionAbstraction}. To do so, the relationship between the random variables $Z^\pi(s, a)$ and $\mathcal{T}^\pi Z^\pi(s, a)$ is rigorously determined for different probability distribution representations (PDF, CDF and QF). Proposition \ref{prop:rec_pdf_def_z} and Corollary \ref{prop:rec_cdf_def_z} respectively provide and prove this link for the PDF and CDF of the random return. However, the case of the QF is more complex and involves an approximation, which is discussed at the end of this section.\\

\proposition \label{prop:rec_pdf_def_z} Let $Z^\pi \in \mathcal{Z}$ be the random return associated with the policy $\pi: \mathcal{S} \rightarrow \mathcal{A}$, which is a random variable mapping the state-action pair $(s, a)\in \mathcal{S} \times \mathcal{A}$ to the realisation of the return $z \in \mathbb{R}$. Additionally, let $p_R(r|s,a)$ be  the probability distribution from which the reward $r \in \mathbb{R}$ is drawn, and $p_T(s'|s,a)$ be the transition probability distribution. Finally, let $\mathcal{T}^\pi: \mathcal{Z} \rightarrow \mathcal{Z}$ be the distributional Bellman operator, and let ${Z^\pi}' \in \mathcal{Z}$ be a random variable such that $Z^\pi = \mathcal{T}^\pi {Z^\pi}'$. Then, the probability density functions $p_{Z^\pi}$ and $p_{{Z^\pi}'}$ associated with the random variables $Z^\pi$ and ${Z^\pi}'$ respect the following equality:
\begin{equation}
\label{EquationProposition1}
    p_{Z^\pi}(z|s, a) = \underset{\underset{s' \sim p_T(\cdot| s, a)}{r\sim p_R(\cdot|s, a)}}{\mathbb{E}} \left[\frac{1}{\gamma}\ p_{{Z^\pi}'} \left( \frac{z - r}{\gamma} \bigg| s', a' \right)\bigg |_{a'=\pi(s')} \right] \quad \forall z \in \mathbb{R},\ s \in \mathcal{S},\ a \in \mathcal{A}\ \text{.}
\end{equation}

\paragraph{Proof} Let $z$ be the return sampled from the random variable $Z^\pi(s, a)$ for the state-action pair $(s, a)$. By marginalising over the reward $r$ collected and over the next state-action pair $(s', a')$ with $a'=\pi(s')$, the PDF of the random return can be expressed as follows:
\begin{equation}
\label{EquationProof1}
    p_{Z^\pi}(z|s, a) = \int p_{Z^\pi}(z|s, a, r, s', a')\ p(r, s', a'|s, a)\ dr\ ds'\ \text{.}
\end{equation}

Considering both the conditional independence and the Markov property of the decision-making process, the expression $p(r, s', a'|s, a)$ can be re-written as follows:
\begin{equation}
\label{EquationProof2}
    p(r, s', a'|s, a) = p_R(r|s, a) p_T(s'|s, a)\ \text{.}
\end{equation}

According to the distributional Bellman equation, the return $z$ can be expressed as a function of both the reward $r$ and the next return $z'$:
\begin{equation}
    z = r + \gamma z'\ \text{.}
\end{equation}

Based on this expression and making use of the change of variables theorem, the PDF $p_{Z^\pi}(z|s, a, r, s', a')$ can be re-expressed as follows:
\begin{align}
    p_{Z^\pi}(z|s, a, r, s', a') 
    &= \left | \gamma \right |^{-1} p_{{Z^\pi}'}\left(z' | s, a, r, s', a'\right) \big|_{z' = \frac{z - r}{\gamma}}\\
    &= \frac{1}{\gamma}\ p_{{Z^\pi}'}\left(\frac{z - r}{\gamma} \bigg| s', a'\right)\ \text{.} \label{EquationProof3}
\end{align}

Finally, by substitution of Equations \eqref{EquationProof2} and \eqref{EquationProof3} into \eqref{EquationProof1}, the following relation is obtained:
\begin{align}
    p_{Z^\pi}(z|s, a)
    &= \int \frac{1}{\gamma}\ p_{{Z^\pi}'}\left(\frac{z - r}{\gamma} \bigg| s', a'\right) p_R(r|s, a) p_T(s'|s, a)\ dr\ ds'\\
    &= \underset{\underset{s' \sim p_T(\cdot| s, a)}{r\sim p_R(\cdot|s, a)}}{\mathbb{E}} \left[\frac{1}{\gamma}\ p_{{Z^\pi}'} \left( \frac{z - r}{\gamma} \bigg| s', a' \right)\bigg |_{a'=\pi(s')} \right]\ \text{.}
\end{align}

\hfill $\square$

\corollary \label{prop:rec_cdf_def_z}  Let $Z^\pi \in \mathcal{Z}$ be the random return associated with the policy $\pi: \mathcal{S} \rightarrow \mathcal{A}$, which is a random variable mapping the state-action pair $(s, a)\in \mathcal{S} \times \mathcal{A}$ to the realisation of the return $z \in \mathbb{R}$. Additionally, let $p_R(r|s,a)$ be  the probability distribution from which the reward $r \in \mathbb{R}$ is drawn, and $p_T(s'|s,a)$ be the transition probability distribution. Finally, let $\mathcal{T}^\pi: \mathcal{Z} \rightarrow \mathcal{Z}$ be the distributional Bellman operator, and let ${Z^\pi}' \in \mathcal{Z}$ be a random variable such that $Z^\pi = \mathcal{T}^\pi {Z^\pi}'$. Then, the cumulative distribution functions $F_{Z^\pi}$ and $F_{{Z^\pi}'}$ associated with the random variables $Z^\pi$ and ${Z^\pi}'$ respect the following equality:
\begin{equation}
    F_{Z^\pi}(z| s, a) = \ \underset{\underset{s' \sim p_T(\cdot| s, a)}{r\sim p_R(\cdot|s, a)}}{\mathbb{E}} \left[ F_{{Z^\pi}'} \left( \frac{z - r}{\gamma} \bigg| s', a' \right)\bigg |_{a'=\pi(s')} \right] \quad \forall z \in \mathbb{R},\ s \in \mathcal{S},\ a \in \mathcal{A}\ \text{.}
\end{equation}

\paragraph{Proof} By considering the definition of the CDF together with Equation \eqref{EquationProposition1} given by Proposition \ref{prop:rec_pdf_def_z}, the following development can be obtained:
\begin{align}
    F_{Z^\pi}(z| s, a) 
    &= \int_{-\infty}^{z} p_{Z^\pi}(z^*|s, a) \; dz^* \\
    &= \int_{-\infty}^{z} \underset{\underset{s' \sim p_T(\cdot| s, a)}{r\sim p_R(\cdot|s, a)}}{\mathbb{E}} \left[ \frac{1}{\gamma}\ p_{{Z^\pi}'} \left( \frac{z^* - r}{\gamma} \bigg| s', a' \right)\bigg |_{a'=\pi(s')}\right] \ dz^* \\
    &= \underset{\underset{s' \sim p_T(\cdot| s, a)}{r\sim p_R(\cdot|s, a)}}{\mathbb{E}} \left[ \int_{-\infty}^{z} \frac{1}{\gamma}\ p_{{Z^\pi}'} \left( \frac{z^* - r}{\gamma} \bigg| s', a' \right)\bigg |_{a'=\pi(s')} \ dz^* \right] \\
    &= \ \underset{\underset{s' \sim p_T(\cdot| s, a)}{r\sim p_R(\cdot|s, a)}}{\mathbb{E}} \left[ \int_{-\infty}^{\frac{z - r}{\gamma}} p_{{Z^\pi}'} \left( z^{**} \bigg| s', a' \right)\bigg |_{a'=\pi(s')} \ dz^{**} \right] \\
    &= \ \underset{\underset{s' \sim p_T(\cdot| s, a)}{r\sim p_R(\cdot|s, a)}}{\mathbb{E}} \left[ F_{{Z^\pi}'} \left( \frac{z - r}{\gamma} \bigg| s', a' \right)\bigg |_{a'=\pi(s')} \right]\ \text{.}
\end{align}

\hfill $\square$

\vspace{0.3cm}

As previously mentioned, the case of the QF is more complex and involves an important approximation. Let $Z^\pi \in \mathcal{Z}$ be the random return associated with the policy $\pi: \mathcal{S} \rightarrow \mathcal{A}$, which is a random variable mapping the state-action pair $(s, a)\in \mathcal{S} \times \mathcal{A}$ to the realisation of the return $z \in \mathbb{R}$. Additionally, let $p_R(r|s,a)$ be  the probability distribution from which the reward $r \in \mathbb{R}$ is drawn, and $p_T(s'|s,a)$ be the transition probability distribution. Finally, let $\mathcal{T}^\pi: \mathcal{Z} \rightarrow \mathcal{Z}$ be the distributional Bellman operator, and let ${Z^\pi}' \in \mathcal{Z}$ be a random variable such that $Z^\pi = \mathcal{T}^\pi {Z^\pi}'$. Then, the quantile functions $F^{-1}_{Z^\pi}$ and $F^{-1}_{{Z^\pi}'}$ associated with the random variables $Z^\pi$ and ${Z^\pi}'$ can be linked based on an approximation as the following:
\begin{equation}
\label{EquationApproximation}
    F^{-1}_{Z^\pi}(\tau| s, a) \simeq \ \underset{\underset{s' \sim p_T(\cdot| s, a)}{r\sim p_R(\cdot|s, a)}}{\mathbb{E}} \left[ r + \gamma F^{-1}_{{Z^\pi}'}(\tau|s', a')\big |_{a'=\pi(s')} \right] \quad \forall \tau \in [0, 1],\ s \in \mathcal{S},\ a \in \mathcal{A}\ \text{.}
\end{equation}

Empirical research on this approximation suggests that it leads to a random variable modelling the quantity $Z^{\pi}$ with the correct expectation but potentially different higher-order moments. Moreover, the error resulting from this approximation is observed to increase with the stochasticity characterising the dynamics of the MDP (transition and reward distributions $p_T$ and $p_R$). On the contrary, Equation \eqref{EquationApproximation} no longer relies on an approximation in the deterministic case. As explained in Section \ref{SectionAbstraction}, this particular approximation may have two completely different implications depending on the objective pursued. If the distributional RL algorithm is used to learn ordinary decision-making policies maximising the expectation of the random return, the approach remains totally sound since the probability distribution learnt has the correct first-order moment. On the contrary, this approximation becomes really problematic if the intention is to learn the complete probability distribution of the random return for implementing risk-aware policies.\\

\section{Implementation details about UMNN}
\label{AppendixUMNN}

As explained in Section \ref{SectionUMNN}, the UMNN requires the solving of an integral, which may be a computationally expensive operation. For the sake of efficiency, this integral is numerically computed via the Clenshaw-Curtis quadrature. This technique presents the key advantage of converging exponentially fast for Lipshitz functions. In practice, only a few function evaluations are required for reaching satisfying accuracy, and these operations can be executed in parallel. This approach makes the complete forward computation of the UMNN quite efficient. Regarding the backward pass, the Leibniz rule can be used to make it more memory efficient. This technique enables to compute the derivative of an integral with respect to its inputs as the integral of the derivatives. For the interested reader, more details about the complete implementation of both forward and backward computations can be found in Appendix B of the research paper originally introducing the UMNN architecture \cite{Wehenkel2019}.\\

Another operation which has to be efficiently implemented is the expectation of the random return $Z^{\pi}$. Indeed, this important quantity is repeatedly evaluated in the UMDQN algorithm, since the decision-making policy $\pi$ selects the action maximising the expectation of the random return. The approach implemented for efficiently and accurately estimating the expectation of the random return $Z^{\pi}$ is described hereafter for the different versions of the UMDQN distributional RL algorithm.

\paragraph{UMDQN-KL algorithm} As hinted in Section \ref{SectionUMNN}, the PDF of the random return $Z^{\pi}$ is modelled with a UMNN as $p_{Z^{\pi}}(z) = g(z) \sigma' \left( \int_0^z g(t) dt + \beta \right)$, where the function $\sigma'(\cdot)$ denotes the PDF of a normal distribution (or equivalently the derivative of a sigmoid function). Consequently, the expectation of the random return $Z^{\pi}$ can be expressed as follows:
\begin{equation}
    \mathbb{E}\left[Z^{\pi} \right] = \int_{z_{\text{min}}}^{z_{\text{max}}} z g(z) \sigma' \left( \int_0^z g(t) dt + \beta \right) dz\ \text{.}
\end{equation}

A straightforward but inefficient solution would be to independently solve each inner integral for different values of the return $z$. Instead, for improved efficiency, these inner integrals are solved simultaneously by making use of the same neural network evaluation multiple times. The UMNN is first evaluated at evenly separated points between $z_\text{min}$ and $z_\text{max}$, and a composite Simpson's rule is then applied to approximate the inner integrals. Thereafter, the expectation of the random return $Z^{\pi}$ is finally computed by estimating the outer integral using the Monte Carlo approach.

\paragraph{UMDQN-C algorithm} Section \ref{SectionUMNN} explains that the CDF of the random return $Z^{\pi}$ is modelled with a UMNN as $F_{Z^{\pi}}(z) = \sigma \left( \int_0^z g(t) dt + \beta \right)$, where the function $\sigma(\cdot)$ is a sigmoid function (or equivalently the CDF of a normal distribution). Consequently, the PDF of the random return can be directly derived as $p_{Z^{\pi}}(z) = g(z) \sigma' \left( \int_0^z g(t) dt + \beta \right)$, and the expectation of the random return $Z^{\pi}$ can be evaluated by following the methodology described in the previous paragraph.

\paragraph{UMDQN-W algorithm} When the probability distribution of the random return $Z^{\pi}$ is represented through the QF, no particular improvement is implemented and the expectation is simply estimated using Monte Carlo, similarly to the state-of-the-art QR-DQN, IQN and FQF distributional RL algorithms.\\

\newpage

\section{Implementation details about the UMDQN algorithm}
\label{AppendixPseudocode}

For the sake of completeness and to ease the reader's understanding of the novel approach proposed, this section provides several implementation details and the detailed pseudocodes for the three versions of the UMDQN algorithm presented in this research work. Hence, the UMDQN-KL, UMDQN-C and UMDQN-W are thoroughly explained in Algorithms \ref{UMDQN-KL}, \ref{UMDQN-C} and \ref{UMDQN-W}, respectively.\\

\vspace*{\fill}
\begin{algorithm}[H]
\small
\caption{UMDQN-KL algorithm}
\begin{algorithmic} 
\STATE Initialise the experience replay memory $M$ of capacity $C$.
\STATE Initialise the main UMNN weights $\theta$ (Xavier initialisation).
\STATE Initialise the target UMNN weights $\theta^- = \theta$.
\FOR{episode = 0 \TO $N$}
    \FOR{$t = 0$ \TO $T$, or until episode termination}
        \STATE Acquire the state $s$ from the environment $\mathcal{E}$.
        \STATE With probability $\epsilon$, select a random action $a \in \mathcal{A}$.
        \STATE Otherwise, select $a = \argmax_{a' \in \mathcal{A}} \mathbb{E}\left[G_Z(s, a'; \theta)\right]$.
        \STATE Interact with the environment $\mathcal{E}$ with action $a$ to get the next state $s'$ and the reward $r$.
        \STATE Store the experience $e = (s, a, r, s')$ in $M$.
        \IF{$t \% T' = 0$}
            \STATE Randomly sample from $M$ a minibatch of $N_e$ experiences $e_i = (s_i, a_i, r_i, s_{i}^{'})$.
            \STATE Derive a discretisation of the domain $\mathcal{X}$ by sampling $N_z$ returns $z \sim \mathcal{U}([z_{\text{min}}, z_{\text{max}}])$.
            \FOR{$i = 0$ \TO $N_e$}
                \FORALL{$z \in \mathcal{X}$}
                    \IF{$s_{i}^{'}$ is terminal}
                        \STATE Set $y_i(z) = \frac{1}{\sigma \sqrt{2 \pi}} \exp{\left(- \frac{1}{2} \left(\frac{z - \mu}{\sigma} \right)^2\right)}$ with $\mu = r_i$ and $\sigma = \frac{z_{\text{max}} - z_{\text{min}}}{N_z}$.
                    \ELSE
                        \STATE Set $y_i(z) = \frac{1}{\gamma}\ G_Z\left(\frac{z - r_i}{\gamma} \bigg| s_{i}^{'}, \argmax_{a_{i}^{'} \in \mathcal{A}} \mathbb{E}\left[G_Z(s_{i}^{'}, a_{i}^{'}; \theta^-)\right]; \theta^-\right)$.
                    \ENDIF
                \ENDFOR
            \ENDFOR
            \STATE Compute the loss $\mathcal{L}_{KL}(\theta) = \sum_{i=0}^{N_e} \left( \sum_{z \in \mathcal{X}}\ y_i(z) \log \left(\frac{y_i(z)}{G_Z(z|s_i, a_i; \theta)}\right) \right)$.
            \STATE Clip the resulting gradient in the range $[0, 1]$.
            \STATE Update the main UMNN parameters $\theta$ using the ADAM optimiser.
        \ENDIF
        \STATE Update the target UMNN parameters $\theta^- = \theta$ every $N^-$ steps.\\
        \STATE Anneal the $\epsilon$-greedy exploration parameter $\epsilon$.
    \ENDFOR
\ENDFOR
\end{algorithmic} 
\label{UMDQN-KL}
\end{algorithm}
\vspace*{\fill}

\newpage

\vspace*{\fill}
\begin{algorithm}[H]
\small
\caption{UMDQN-C algorithm}
\begin{algorithmic} 
\STATE Initialise the experience replay memory $M$ of capacity $C$.
\STATE Initialise the main UMNN weights $\theta$ (Xavier initialisation).
\STATE Initialise the target UMNN weights $\theta^- = \theta$.
\FOR{episode = 0 \TO $N$}
    \FOR{$t = 0$ \TO $T$, or until episode termination}
        \STATE Acquire the state $s$ from the environment $\mathcal{E}$.
        \STATE With probability $\epsilon$, select a random action $a \in \mathcal{A}$.
        \STATE Otherwise, select $a = \argmax_{a' \in \mathcal{A}} \mathbb{E}\left[G_Z(s, a'; \theta)\right]$.
        \STATE Interact with the environment $\mathcal{E}$ with action $a$ to get the next state $s'$ and the reward $r$.
        \STATE Store the experience $e = (s, a, r, s')$ in $M$.
        \IF{$t \% T' = 0$}
            \STATE Randomly sample from $M$ a minibatch of $N_e$ experiences $e_i = (s_i, a_i, r_i, s_{i}^{'})$.
            \STATE Derive a discretisation of the domain $\mathcal{X}$ by sampling $N_z$ returns $z \sim \mathcal{U}([z_{\text{min}}, z_{\text{max}}])$.
            \FOR{$i = 0$ \TO $N_e$}
                \FORALL{$z \in \mathcal{X}$}
                    \IF{$s_{i}^{'}$ is terminal}
                        \STATE Set $y_i(z) = \begin{cases}
                                                0 & \text{if } z < r_i, \\
                                                1 & \text{otherwise.}
                                            \end{cases}$
                    \ELSE
                        \STATE Set $y_i(z) = G_Z\left(\frac{z - r_i}{\gamma} \bigg| s_{i}^{'}, \argmax_{a_{i}^{'} \in \mathcal{A}} \mathbb{E}\left[G_Z(s_{i}^{'}, a_{i}^{'}; \theta^-)\right]; \theta^-\right)$.
                    \ENDIF
                \ENDFOR
            \ENDFOR
            \STATE Compute the loss $\mathcal{L}_C(\theta) = \sum_{i=0}^{N_e} \left(\sum_{z \in \mathcal{X}} \left(y_i(z) - G_Z(z|s_i, a_i; \theta)\right)^2\right)^{1/2}$.
            \STATE Clip the resulting gradient in the range $[0, 1]$.
            \STATE Update the main UMNN parameters $\theta$ using the ADAM optimiser.
        \ENDIF
        \STATE Update the target UMNN parameters $\theta^- = \theta$ every $N^-$ steps.\\
        \STATE Anneal the $\epsilon$-greedy exploration parameter $\epsilon$.
    \ENDFOR
\ENDFOR
\end{algorithmic} 
\label{UMDQN-C}
\end{algorithm}
\vspace*{\fill}

\newpage

\vspace*{\fill}
\begin{algorithm}[H]
\small
\caption{UMDQN-W algorithm}
\begin{algorithmic} 
\STATE Initialise the experience replay memory $M$ of capacity $C$.
\STATE Initialise the main UMNN weights $\theta$ (Xavier initialisation).
\STATE Initialise the target UMNN weights $\theta^- = \theta$.
\FOR{episode = 0 \TO $N$}
    \FOR{$t = 0$ \TO $T$, or until episode termination}
        \STATE Acquire the state $s$ from the environment $\mathcal{E}$.
        \STATE With probability $\epsilon$, select a random action $a \in \mathcal{A}$.
        \STATE Otherwise, select $a = \argmax_{a' \in \mathcal{A}} \mathbb{E}\left[G_Z(s, a'; \theta)\right]$.
        \STATE Interact with the environment $\mathcal{E}$ with action $a$ to get the next state $s'$ and the reward $r$.
        \STATE Store the experience $e = (s, a, r, s')$ in $M$.
        \IF{$t \% T' = 0$}
            \STATE Randomly sample from $M$ a minibatch of $N_e$ experiences $e_i = (s_i, a_i, r_i, s_{i}^{'})$.
            \STATE Sample $N_{\tau}$ values for the first quantile fraction $\tau_i \sim \mathcal{U}([0, 1])$.
            \STATE Sample $N_{\tau}$ values for the second quantile fraction $\tau_j \sim \mathcal{U}([0, 1])$.
            \FOR{$k = 0$ \TO $N_e$}
                \FOR{$i = 0$ \TO $N_{\tau}$}
                    \FOR{$j = 0$ \TO $N_{\tau}$}
                        \STATE $y_k(\tau_j) = \begin{cases}
                                                    r_k & \text{if $s_{k}^{'}$ terminal,} \\
                                                    r_k + \gamma \ G_Z\left(\tau_j \big| s_{k}^{'}, \argmax_{a_{k}^{'} \in \mathcal{A}} \mathbb{E}\left[G_Z(s_{k}^{'}, a_{k}^{'}; \theta^-)\right]; \theta^-\right) & \text{otherwise.}
                                                \end{cases}$
                        \ENDFOR
                    \STATE $\delta_{ij}(k) = y_k(\tau_j) - G_Z(\tau_i|s_k, a_k; \theta)$.
                \ENDFOR
            \ENDFOR
            \STATE Compute the loss $\mathcal{L}_W(\theta) = \sum_{k=0}^{N_e} \left(\sum_{i=0}^{N_{\tau}} \mathbb{E}_j \left[\rho_{\tau_i}^{\kappa}(\delta_{ij}(k))\right]\right)$.
            \STATE Clip the resulting gradient in the range $[0, 1]$.
            \STATE Update the main UMNN parameters $\theta$ using the ADAM optimiser.
        \ENDIF
        \STATE Update the target UMNN parameters $\theta^- = \theta$ every $N^-$ steps.\\
        \STATE Anneal the $\epsilon$-greedy exploration parameter $\epsilon$.
    \ENDFOR
\ENDFOR
\end{algorithmic} 
\label{UMDQN-W}
\end{algorithm}
\vspace*{\fill}

\newpage

Modelling the probability distribution of the random return for a terminal state may be tricky and deserves a brief discussion. In this case, the RL agent shall not receive any future rewards, and the random return distribution degenerates into a Dirac distribution shifted by the value of the last reward collected. In practice, such a particular probability distribution may be quite difficult to approximate with a DNN, depending on the distribution representation. Moreover, it may potentially lead to numerical instabilities when computing the loss. For these reasons, this research work makes the choice to smooth out the Dirac distribution whenever appropriate. For the UMDQN-KL algorithm learning a PDF, a normal distribution with a tiny standard deviation is used as a replacement for the problematic Dirac distribution. For the UMDQN-C algorithm which is based on the random return CDF, the step function with infinite slope is supplanted by a smoother version with a large constant slope. Finally, the case of the UMDQN algorithm is left untouched since the QF of a Dirac distribution is trivial to model with a DNN (constant function).\\

As explained in Section \ref{SectionUMDQN}, the loss defined in Equation \eqref{EquationLoss} is approximated in the UMDQN algorithm, which may introduce a bias. This problem has already been demonstrated for the Wasserstein distance \cite{Bellemare2017C51} and a solution has been proposed \cite{Dabney2018QRDQN}: the \textit{(conditional) quantile regression} method \cite{Koenker2005}. Without going into too much detail, this alternative approach is based on the \textit{quantile regression loss}, which is an asymmetric convex loss function respectively penalising overestimation and underestimation errors with weights $\tau$ and $1 - \tau$, with $\tau \in [0, 1]$ being a quantile fraction. This technique is used in the UMDQN-W algorithm, similarly to the state-of-the-art QR-DQN, IQN and FQF distributional RL algorithms. In fact, to ensure smoothness at zero, a slightly modified quantile regression loss is used by these algorithms, the \textit{quantile Huber loss} which is defined for the error $x \in \mathbb{R}$ as follows:
\begin{equation}
    \rho_{\tau}^{\kappa}(x) = \left|\tau - 1_{\{x < 0\}}\right| \frac{\mathcal{H}_{\kappa}(x)}{\kappa}\ \text{,}
\end{equation}
\begin{equation}
    \mathcal{H}_{\kappa}(x) = \left\{
                                \begin{array}{ll}
                                    \frac{1}{2}x^2 & \mbox{if } |x| \leq \kappa, \\
                                    \kappa(|x| - \frac{1}{2} \kappa) & \mbox{otherwise,}
                                \end{array}
                               \right.
\end{equation}

where the threshold $\kappa$ is a parameter to be tuned. An illustration of the quantile Huber loss with $\kappa = 1$ is provided in Figure \ref{QuantileHuberLoss} below. This alternative loss function is evaluated on the pairwise temporal difference (TD) errors $\delta_{ij}$ expressed as follows:
\begin{equation}
    \delta_{ij} = r + \gamma  F_{Z^{\pi}}^{-1}\left(\tau_j|s', \pi(s') \right) -  F_{Z^{\pi}}^{-1}\left(\tau_i|s, a\right)\ \text{.}
\end{equation}

\begin{figure}[H]
    \centering
    \includegraphics[width=0.6\linewidth]{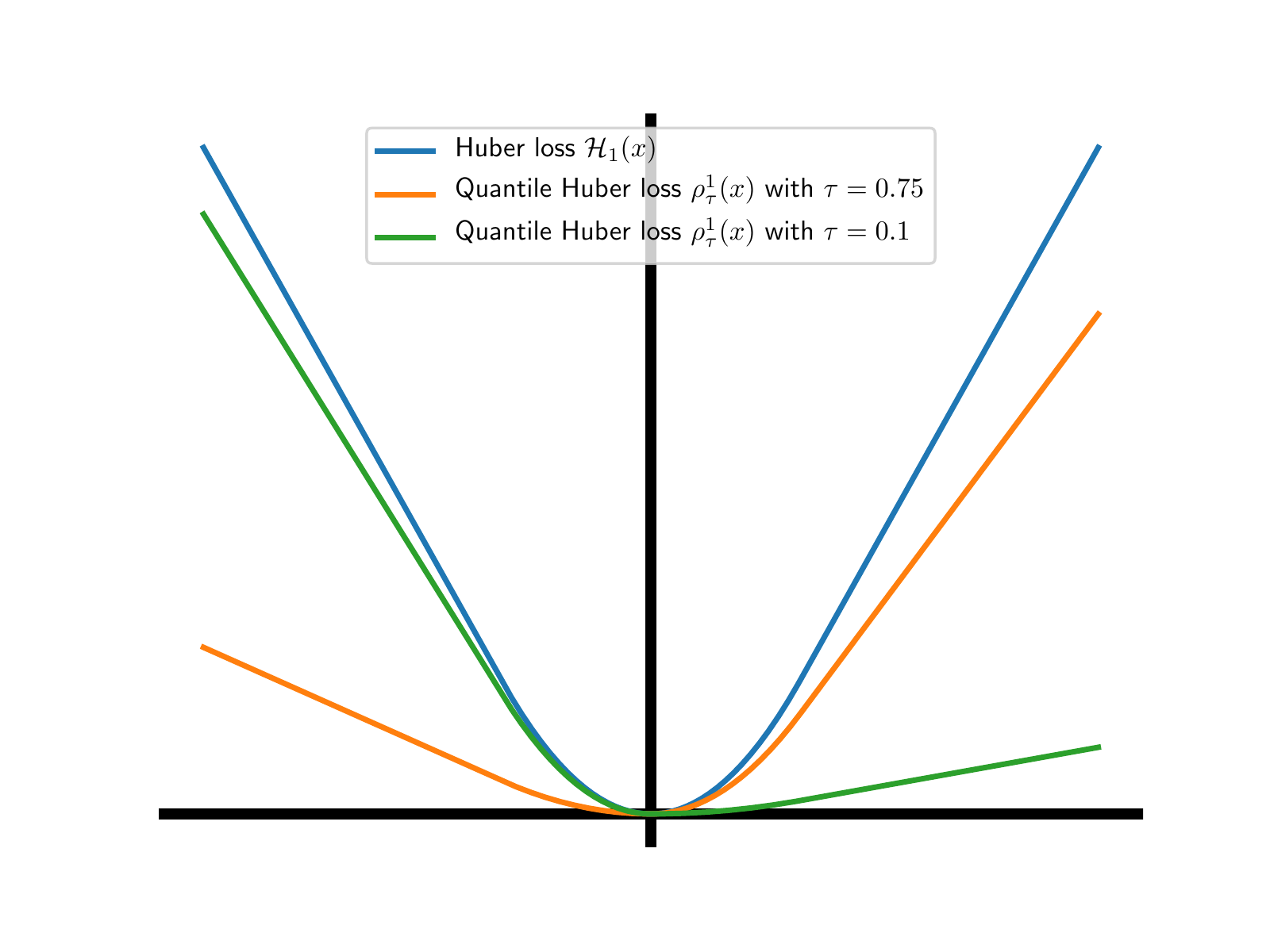}
    \caption{Illustration of the quantile Huber loss.}
    \label{QuantileHuberLoss}
\end{figure}

\section{Benchmark environments}
\label{AppendixEnvironments}

\begin{figure}[H]
    \centering
    \includegraphics[width=1\linewidth, trim={2.9cm 7cm 2.9cm 7cm}, clip]{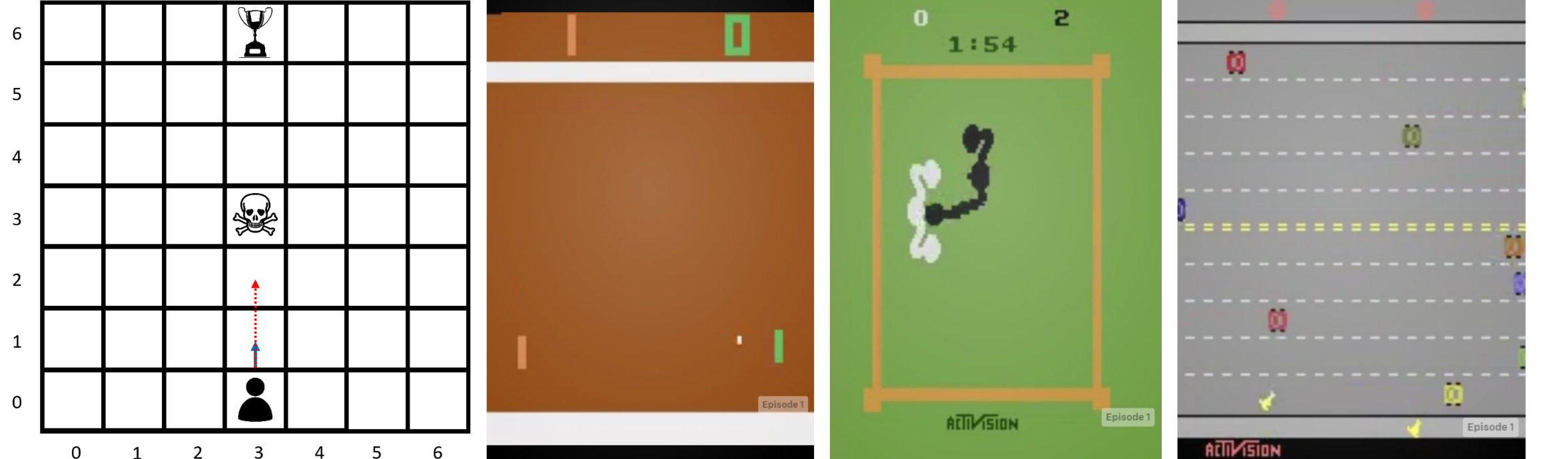}
    \caption{Illustration of some benchmark environments with, from left to right, the stochastic grid world and the Atari games Pong, Boxing and Freeway.}
    \label{BenchmarkEnvironments1}
\end{figure}

\begin{figure}[H]
    \centering
    \includegraphics[width=1\linewidth, trim={3.5cm 8cm 3.5cm 7.5cm}, clip]{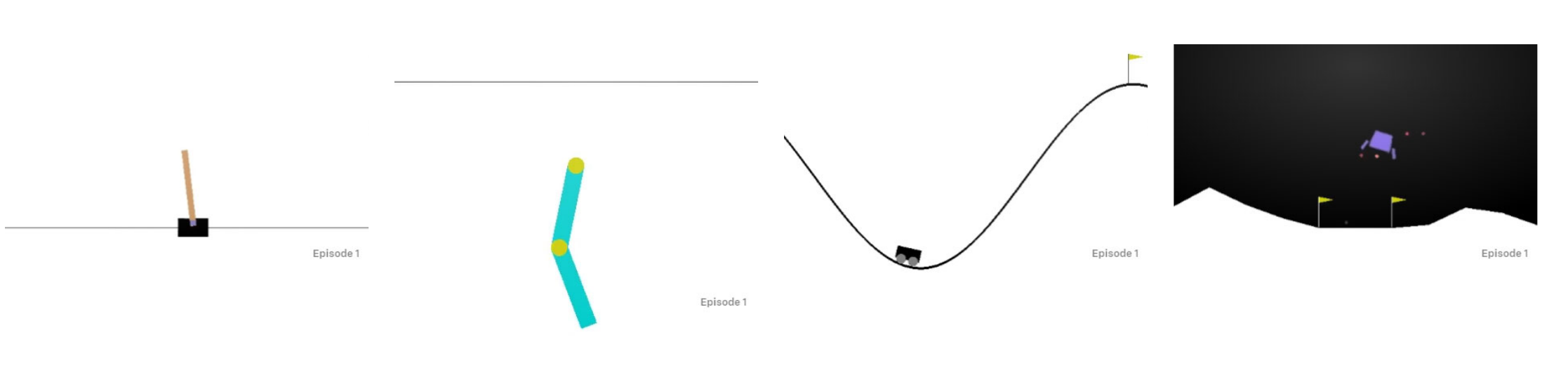}
    \caption{Illustration of some benchmark environments with, from left to right, the CartPole, Acrobot, MountainCar and LunarLander classic control problems.}
    \label{BenchmarkEnvironments2}
\end{figure}

\begin{figure}[H]
    \centering
    \includegraphics[width=0.75\linewidth, trim={4.5cm 3.5cm 4.5cm 3.5cm}, clip]{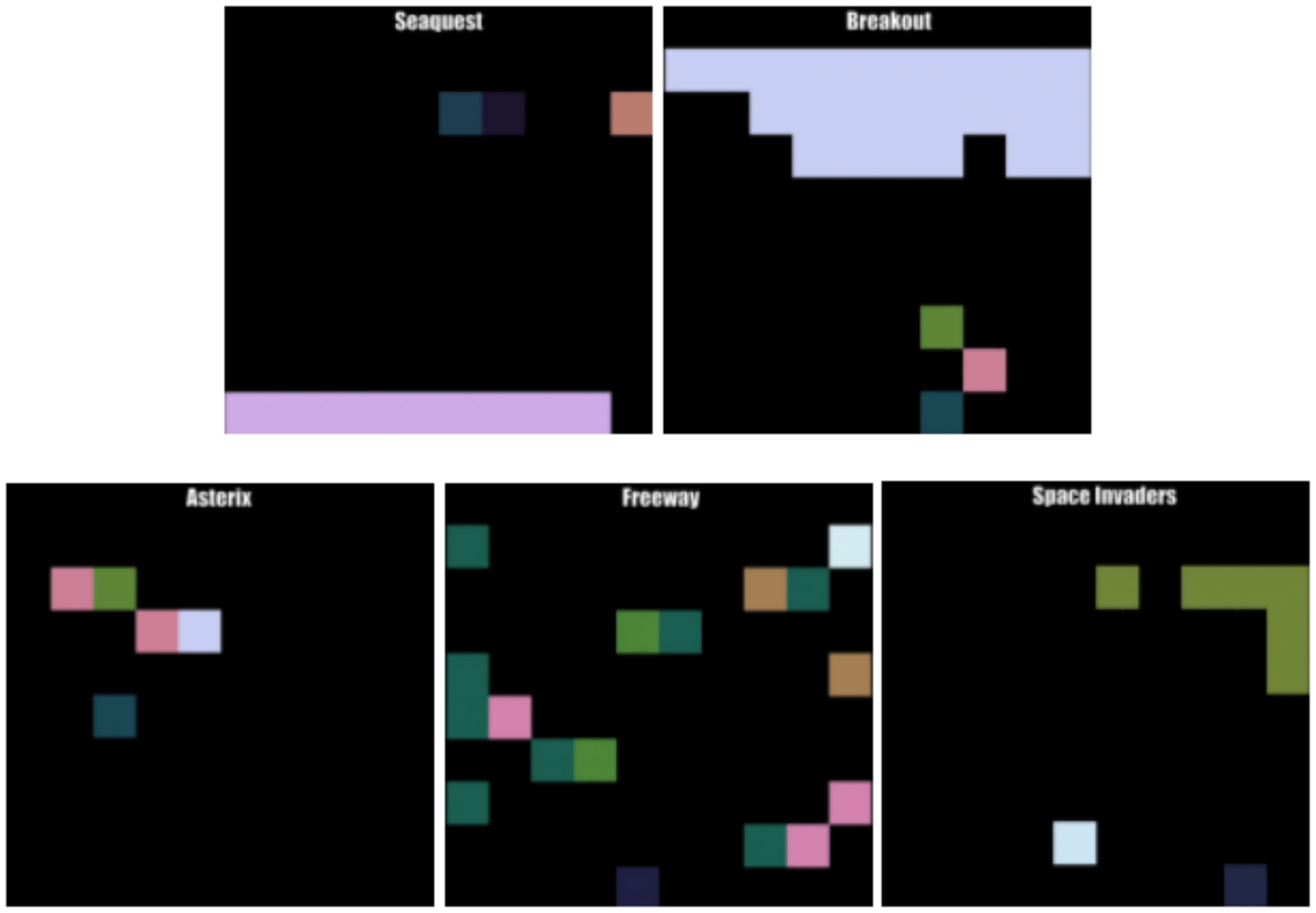}
    \caption{Illustration of some benchmark environments with all the MinAtar games.}
    \label{BenchmarkEnvironments3}
\end{figure}

This section provides an accurate description of the benchmark environments adopted in this research work, together with various implementation details. Additionally, these environments and their associated control problems are illustrated in Figures \ref{BenchmarkEnvironments1}, \ref{BenchmarkEnvironments2} and \ref{BenchmarkEnvironments3}.

\paragraph{Stochastic grid world} This benchmark environment is a $7 \times 7$ grid world, an environment which is commonly considered for analysing and evaluating the performance of RL algorithms. The objective of the agent is simply to reach a certain target location which is fixed, while avoiding a fixed trap. The particularity of this grid world is that both the transition and reward functions are stochastic ($p_T$ and $p_R$). The intent behind this additional complexity is to better highlight the impact of the distributional RL approach and analyse the probability distributions learnt. The underlying MDP can be defined as follows:

\begin{itemize}
    \item [$\bullet$] $\mathcal{S} \in \{0, ..., 6 \} \times \{0, ..., 6 \}$, a state $s$ being composed of the two coordinates of the agent within the grid,
    \item [$\bullet$] $\mathcal{A} = \{\texttt{RIGHT},\ \texttt{UP},\ \texttt{LEFT},\ \texttt{DOWN}\}$, with an action $a$ being a moving direction,
    \item [$\bullet$] $p_R(r|s, a) \sim \mathcal{N}(\mu,\sigma^{2})$ where:
        \begin{itemize}
            \item $\mu = 1$ if the agent reaches the target location (terminal state),
            \item $\mu = -1$ if the agent falls into the trap (terminal state),
            \item $\mu = 0$ otherwise,
            \item $\sigma = 0.1$ at anytime,
        \end{itemize}
    \item [$\bullet$] $p_T(s'|s, a)$ associates a 50\% chance to move twice in the chosen direction instead of once, while keeping the agent within the $7 \times 7$ grid world (no border crossing allowed),
    \item [$\bullet$] $p_0$ associates an equal probability to all states $s_0 \in \mathcal{S}$, except for the two states corresponding to the trap and target locations which have a null probability,
    \item [$\bullet$] $\gamma = 0.5$.
\end{itemize}

\paragraph{Selection of Atari games} This benchmark environment consists of a set of three representative Atari games from the Atari-57 benchmark \cite{Bellemare2013}: Pong, Boxing and Freeway. Similarly to the stochastic grid world, the control problems are made slightly more complex to highlight the impact of the distributional RL approach. Indeed, the deterministic Atari games are made stochastic by using the sticky action generalisation technique (stochastic transitions, but still deterministic rewards). The implementation adopted is the \texttt{\{\}NoFrameskip} from OpenAI gym \cite{Brockman2016}, together with the following wrappers:

\begin{itemize}
    \item [$\bullet$] Formatting of a frame to 84 $\times$ 84 pixels,
    \item [$\bullet$] Normalisation of the values of the pixels,
    \item [$\bullet$] Clipping of the reward to $\{+1,\ 0,\ -1\}$,
    \item [$\bullet$] Sending of the episode termination signal when all the agent's lives are lost,
    \item [$\bullet$] Execution of a random number of \texttt{NOOP} actions at the beginning of an episode (maximum 30),
    \item [$\bullet$] Execution of sticky actions with a $0.25$ probability,
    \item [$\bullet$] Frame skipping and maximisation operation with period 4,
    \item [$\bullet$] Stacking of the final 4 frames.
\end{itemize}

\paragraph{Selection of classic control environments} This benchmark environment consists of a set of four classic control problems from the popular OpenAI Gym toolkit \cite{Brockman2016}: CartPole, Acrobot, MountainCar and LunarLander. Without going into too much detail, these environments can be characterised as follows:

\begin{itemize}

    \item [$\bullet$] \texttt{CartPole-v0}: The objective is to balance a pole attached by a non-actuated joint to a cart moving along a frictionless track. The state is composed of four continuous values: the cart position, the cart velocity, the pole angle and the pole velocity at the tip. The agent's action is either to push the cart to the left or to the right. A reward of +1 is received for each time step with the pole remaining balanced. An episode terminates when the pole angle is more than $\pm12^{\circ}$ or when the cart reaches the edge of the display, but also if the episode length is greater than 200.
    
    \item [$\bullet$] \texttt{Acrobot-v1}: This system is composed of a double-jointed pendulum, with the joint between the two links being actuated. The objective is to swing the pendulum so that the end of the outer link reaches a given height. The state is a six-dimensional vector describing the system's angles and velocities. To achieve its goal, the agent has three actions at its disposal: either applying no torque, or applying a fixed torque to the left or to the right. The agent is given a reward of -1 for each time step before achieving the objective position. An episode either terminates when this objective is achieved or when the episode length exceeds 500.
    
    \item [$\bullet$] \texttt{MountainCar-v0}: The objective is to drive an underpowered car up a steep hill. To achieve that goal, the agent has to learn to leverage potential energy by driving back and forth for gaining momentum. The state consists of both the position and velocity of the car. The agent's action can either be to push the car to the left, do nothing or push the car to the right. A reward of -1 is received at each time step until the goal position is eventually reached. An episode terminates when this particular position is achieved, or if the episode length is greater than 200.
    
    \item [$\bullet$] \texttt{LunarLander-v2}: This environment consists of a simulated 2D world within which the objective is to safely land a lander with a limited amount of fuel on a target location. The RL state is composed of 8 values: the two coordinates of the lander, its linear velocities in the horizontal and vertical directions, its angle and angular velocity, as well as two booleans representing whether each leg is in contact with the ground or not. To achieve its objective, the agent has access to four actions: do nothing, fire the left orientation engine, fire the main engine and fire the right orientation engine. A reward between +100 and +140 is received for safely landing and coming to rest at the designated location. Additionally, a crash results in receiving a -100 reward, while coming to rest induces a reward of +100. There is also a +10 reward generated for each leg with ground contact. Finally, rewards of -0.3 and -0.03 are respectively obtained for each time step firing the main and side engines. The termination of an episode occurs when the lander crashes or gets outside of the viewport, or when the lander is no longer awake (meaning that it does not move nor collide with any other body).
    
\end{itemize}

\paragraph{Selection of MinAtar games} This benchmark environment consists of a set of five MinAtar games \cite{Young2019}: Seaquest, Breakout, Asterix, Freeway and SpaceInvaders. As explained in Section \ref{SectionBenchmarkEnvironments}, the core objective behind these environments is to make RL experimentation around Atari games more accessible and efficient. To do so, MinAtar reduces the representation complexity of five representative Atari games, while avoiding as much as possible altering the mechanics of the original games. The alternative state representation is of dimension $10 \times 10 \times n$ and binary, where $n$ is the number of channels representing a game-specific object. In addition to other useful features, MinAtar games also include stochasticity in the form of sticky actions and randomised spawn locations, which is particularly important for analysing distributional RL algorithms. \\

All the benchmark environments used in this research paper are illustrated in Figures \ref{BenchmarkEnvironments1}, \ref{BenchmarkEnvironments2}, \ref{BenchmarkEnvironments3}. To end this section, it has to be mentioned that this research work makes the choice to evaluate the performance of a decision-making policy by computing the cumulative reward achieved, similarly to previous works in both classical and distributional RL. This approach is sound for the benchmark environments studied thanks to terminal states or a maximum number of steps preventing the performance from indefinitely increasing.\\

\newpage

\section{Hyperparameters}
\label{AppendixParameters}

This section presents the main hyperparameters adopted for generating the results presented in both Section \ref{SectionResults} and \ref{AppendixResults}. The most important criterion taken into account for the selection/tuning of these hyperparameters is the fair comparison between the DRL algorithms studied, while considering at the same time the values reported by the state-of-the-art distributional RL algorithms. First of all, Table \ref{HyperparametersDescription} provides a brief description of the hyperparameters to be tuned in the scope of this research work. Then, Table \ref{HyperparametersDomain} presents the domain $\mathcal{X}$ selected for the different benchmark environments studied (lower and upper bounds). Finally, Tables \ref{HyperparametersGridWorld}, \ref{HyperparametersAtari}, \ref{HyperparametersClassicControl} and \ref{HyperparametersMinAtar} present the hyperparameters used for generating the results for all the benchmark environments presented in \ref{AppendixEnvironments}.\\

A complicated choice when it comes to hyperparameters tuning concerns the domain $\mathcal{X}$. Firstly, is it fairer to have the same lower bound $z_{\text{min}}$ and upper bound $z_{\text{max}}$ for the domain $\mathcal{X}$ for all the benchmark environments as it is generally done in scientific literature, or slightly tune these two hyperparameters for each environment? Secondly, if tuned, how to efficiently select relevant values for these two bounds without requiring complicated analyses? The present research work makes the choice to specialise the domain $\mathcal{X}$ for each benchmark environment. This decision is motivated by the diversity of the environments, with completely different ranges for the random return. For each control problem, a quick analysis is performed to estimate the minimum and maximum returns based on the shape of the reward probability distribution $p_R$.\\

Let's consider for instance the CartPole environment. Knowing that a +1 reward is obtained at each time step until episode termination and that the discount factor is equal to 0.99, it is possible to estimate the lower and upper bounds for the returns. In the worst case, the agent totally fails and the episode ends after a few time steps, meaning that the minimum return is close to 0. On the contrary, if the agent manages to continuously keep the pole balanced, the maximum return can be estimated as follows:
\begin{equation*}
    z_{\text{max}} = \sum_{i=0}^{\infty} 0.99^i \simeq 100 \ \text{.}
\end{equation*}

After consideration of a small margin, an appropriate domain $\mathcal{X}$ is obtained for this particular control problem. The same analysis can be repeated for the other benchmark environments, to get Table \ref{HyperparametersDomain}.\\

\begin{table}[H]
  \small
  \caption{Description of the main hyperparameters associated with the distributional RL algorithms studied.}
  \label{HyperparametersDescription}
  \centering
  \begin{tabular}{ll}
    \toprule
    \textbf{Hyperparameter} & \textbf{Description} \\
    \midrule
    Network structure & Structure of the DNN representing the random return $Z^{\pi}$ (neurons per layer). \\
    Discount factor & Discount factor $\gamma$ adopted for the Q-learning update. \\
    Learning rate & Learning rate of the DL optimiser (ADAM). \\
    Optimiser epsilon & Epsilon of the DL optimiser (ADAM) to improve numerical stability. \\
    Main update frequency & Frequency $T'$ (in number of steps) at which the main network is updated. \\
    Target update frequency & Frequency $N^-$ (in number of steps) at which the target network is updated. \\
    Replay memory capacity & Capacity $C$ (in number of experiences) of the experience replay memory $M$. \\
    Batch size & Size of the batch $N_e$ (in experiences) used for each gradient descent iteration. \\
    $\epsilon$-greedy start & Initial value of $\epsilon$, for the $\epsilon$-greedy exploration technique. \\
    $\epsilon$-greedy end & Final value of $\epsilon$, for the $\epsilon$-greedy exploration technique. \\
    $\epsilon$-greedy decay & Exponential decay (in steps) of $\epsilon$, for the $\epsilon$-greedy exploration technique. \\
    $\epsilon$-greedy test & Value of $\epsilon$ when testing the policy, for the $\epsilon$-greedy exploration technique. \\
    Number of $z$ values & Number of returns $N_z$ used for representing distributions (PDF and CDF). \\
    Number of $\tau$ values & Number of quantile fractions $N_{\tau}$ used for representing distributions (QF). \\
    \bottomrule
  \end{tabular}
\end{table}

\begin{table}[H]
  \small
  \caption{Domain $\mathcal{X}$ set for the different benchmark environments.}
  \label{HyperparametersDomain}
  \centering
  \begin{tabular}{lll}
    \toprule
    \textbf{Benchmark environment} & \textbf{Lower bound of $\mathcal{X}$} & \textbf{Upper bound of $\mathcal{X}$} \\
    \midrule
    Stochastic grid world & -2 & 2 \\
    Atari Pong & -5 & 5 \\
    Atari Boxing & -1 & 10 \\
    Atari Freeway & -1 & 10 \\
    CartPole & -10 & 110 \\
    Acrobot & -110 & 10 \\
    MountainCar & -110 & 10 \\
    LunarLander & -150 & 200 \\
    MinAtar Asterix & -1 & 10 \\
    MinAtar Breakout & -1 & 10 \\
    MinAtar Freeway & -1 & 10 \\
    MinAtar Seaquest & -1 & 10 \\
    MinAtar SpaceInvaders & -1 & 20 \\
    \bottomrule
  \end{tabular}
\end{table}

\begin{table}[H]
  \small
  \caption{Hyperparameters selected for the stochastic grid world benchmark environment.}
  \label{HyperparametersGridWorld}
  \centering
  \begin{tabular}{llll}
    \toprule
    \textbf{Hyperparameter} & \textbf{UMDQN-KL} & \textbf{UMDQN-C} & \textbf{UMDQN-W} \\
    \midrule
    Network structure & $[128]_{\text{DNN}} + [128]_{\text{UMNN}}$ & $[128]_{\text{DNN}} + [128]_{\text{UMNN}}$ & $[128]_{\text{DNN}} + [128]_{\text{UMNN}}$ \\
    Discount factor & 0.5 & 0.5 & 0.5 \\
    Learning rate & $10^{-4}$ & $10^{-4}$ & $10^{-4}$ \\
    Optimiser epsilon & $10^{-5}$ & $10^{-5}$ & $10^{-5}$ \\
    Main update frequency & 1 & 1 & 1 \\
    Target update frequency & 1000 & 1000 & 1000 \\
    Replay memory capacity & $10^{4}$ & $10^{4}$ & $10^{4}$ \\
    Batch size & 32 & 32 & 32 \\
    $\epsilon$-greedy start & 1.0 & 1.0 & 1.0 \\
    $\epsilon$-greedy end & 0.01 & 0.01 & 0.01 \\
    $\epsilon$-greedy decay & $10^{4}$ & $10^{4}$ & $10^{4}$ \\
    $\epsilon$-greedy test & 0.001 & 0.001 & 0.001 \\
    Number of $z$ values & 200 & 200 & - \\
    Number of $\tau$ values & - & - & 200 \\
    \bottomrule
  \end{tabular}
\end{table}

\begin{table}[H]
  \small
  \caption{Hyperparameters selected for the Atari games benchmark environments.}
  \label{HyperparametersAtari}
  \centering
  \begin{tabular}{llll}
    \toprule
    \textbf{Hyperparameter} & \textbf{UMDQN-KL} & \textbf{UMDQN-C} & \textbf{UMDQN-W} \\
    \midrule
    Network structure & DQN $+ \ [128]_{\text{UMNN}}$ & DQN $+ \ [128]_{\text{UMNN}}$ & DQN $+ \ [128]_{\text{UMNN}}$ \\
    Discount factor & 0.99 & 0.99 & 0.99 \\
    Learning rate & $5 \times 10^{-5}$ & $5 \times 10^{-5}$ & $5 \times 10^{-5}$ \\
    Optimiser epsilon & $10^{-5}$ & $10^{-5}$ & $10^{-5}$ \\
    Main update frequency & 4 & 4 & 4 \\
    Target update frequency & $10^{4}$ & $10^{4}$ & $10^{4}$ \\
    Replay memory capacity & $10^{5}$ & $10^{5}$ & $10^{5}$ \\
    Batch size & 32 & 32 & 32 \\
    $\epsilon$-greedy start & 1.0 & 1.0 & 1.0 \\
    $\epsilon$-greedy end & 0.01 & 0.01 & 0.01 \\
    $\epsilon$-greedy decay & $10^{6}$ & $10^{6}$ & $10^{6}$ \\
    $\epsilon$-greedy test & 0.001 & 0.001 & 0.001 \\
    Number of $z$ values & 200 & 200 & - \\
    Number of $\tau$ values & - & - & 200 \\
    \bottomrule
  \end{tabular}
\end{table}

\begin{table}[H]
  \small
  \caption{Hyperparameters selected for the classic control benchmark environments.}
  \label{HyperparametersClassicControl}
  \centering
  \begin{tabular}{llll}
    \toprule
    \textbf{Hyperparameter} & \textbf{UMDQN-KL} & \textbf{UMDQN-C} & \textbf{UMDQN-W} \\
    \midrule
    Network structure & $[128]_{\text{DNN}} + [128]_{\text{UMNN}}$ & $[128]_{\text{DNN}} + [128]_{\text{UMNN}}$ & $[128]_{\text{DNN}} + [128]_{\text{UMNN}}$ \\
    Discount factor & 0.99 & 0.99 & 0.99 \\
    Learning rate & $10^{-4}$ & $10^{-4}$ & $10^{-4}$ \\
    Optimiser epsilon & $10^{-5}$ & $10^{-5}$ & $10^{-5}$ \\
    Main update frequency & 1 & 1 & 1 \\
    Target update frequency & 1000 & 1000 & 1000 \\
    Replay memory capacity & $10^{4}$ & $10^{4}$ & $10^{4}$ \\
    Batch size & 32 & 32 & 32 \\
    $\epsilon$-greedy start & 1.0 & 1.0 & 1.0 \\
    $\epsilon$-greedy end & 0.01 & 0.01 & 0.01 \\
    $\epsilon$-greedy decay & $10^{4}$ & $10^{4}$ & $10^{4}$ \\
    $\epsilon$-greedy test & 0.001 & 0.001 & 0.001 \\
    Number of $z$ values & 200 & 200 & - \\
    Number of $\tau$ values & - & - & 200 \\
    \bottomrule
  \end{tabular}
\end{table}

\begin{table}[H]
  \small
  \caption{Hyperparameters selected for the MinAtar games benchmark environments.}
  \label{HyperparametersMinAtar}
  \centering
  \begin{tabular}{llll}
    \toprule
    \textbf{Hyperparameter} & \textbf{UMDQN-KL} & \textbf{UMDQN-C} & \textbf{UMDQN-W} \\
    \midrule
    Network structure & DQN $+ \ [128]_{\text{UMNN}}$ & DQN $+ \ [128]_{\text{UMNN}}$ & DQN $+ \ [128]_{\text{UMNN}}$ \\
    Discount factor & 0.99 & 0.99 & 0.99 \\
    Learning rate & $5 \times 10^{-5}$ & $5 \times 10^{-5}$ & $5 \times 10^{-5}$ \\
    Optimiser epsilon & $10^{-5}$ & $10^{-5}$ & $10^{-5}$ \\
    Main update frequency & 4 & 4 & 4 \\
    Target update frequency & $10^{4}$ & $10^{4}$ & $10^{4}$ \\
    Replay memory capacity & $10^{5}$ & $10^{5}$ & $10^{5}$ \\
    Batch size & 32 & 32 & 32 \\
    $\epsilon$-greedy start & 1.0 & 1.0 & 1.0 \\
    $\epsilon$-greedy end & 0.01 & 0.01 & 0.01 \\
    $\epsilon$-greedy decay & $10^{6}$ & $10^{6}$ & $10^{6}$ \\
    $\epsilon$-greedy test & 0.001 & 0.001 & 0.001 \\
    Number of $z$ values & 200 & 200 & - \\
    Number of $\tau$ values & - & - & 200 \\
    \bottomrule
  \end{tabular}
\end{table}

\newpage

\section{Comparison with state-of-the-art distributional RL algorithms}
\label{AppendixResults}

\begin{figure}[H]
    \centering
    \begin{subfigure}[b]{0.328\textwidth}
        \centering
        \includegraphics[width=1\linewidth, trim={0.25cm 0cm 2.1cm 1cm}, clip]{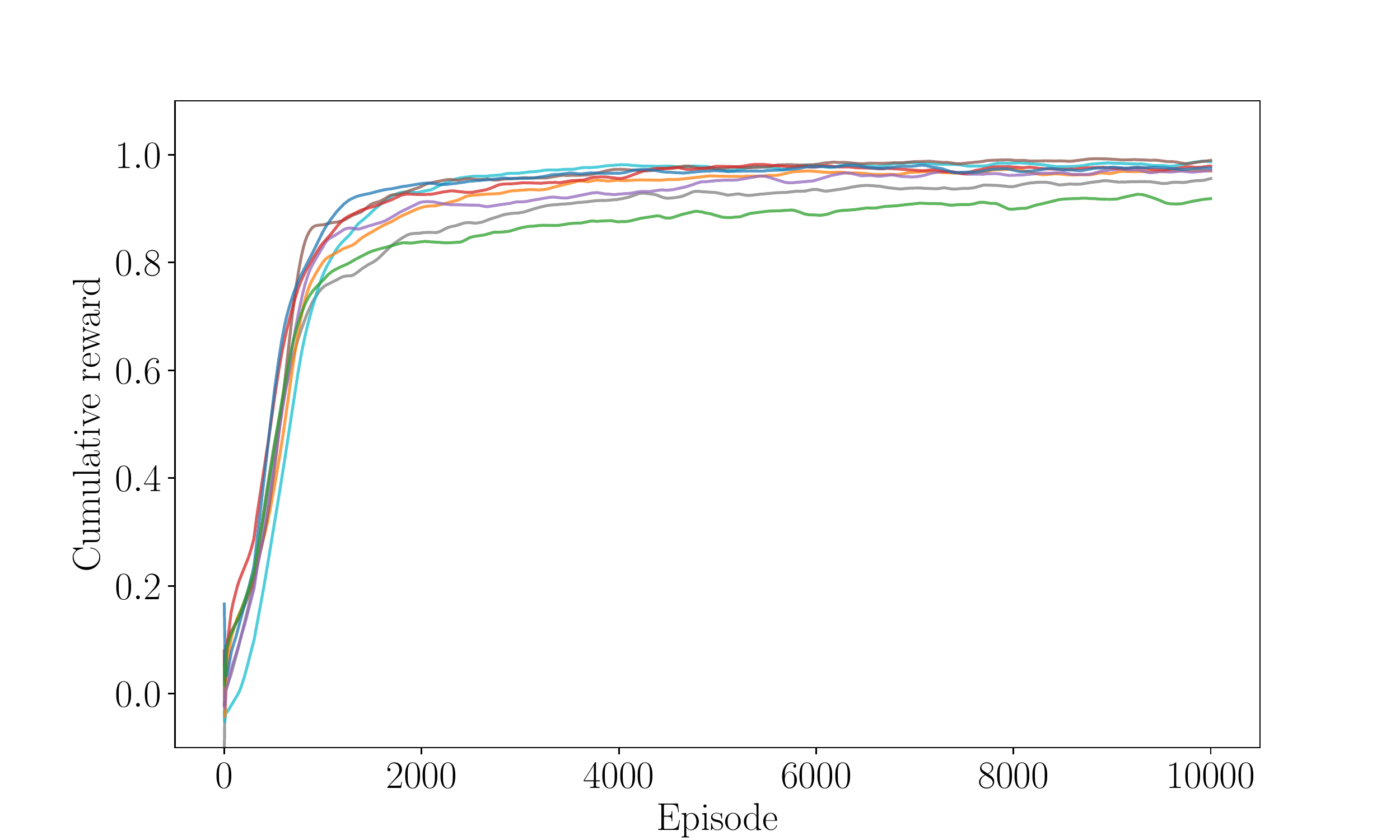}
        \caption{Stochastic grid world}
        \label{AdditionalResultsStochasticGridWorld}
    \end{subfigure}
    \hfill
    \begin{subfigure}[b]{0.328\textwidth}
        \centering
        \includegraphics[width=1\linewidth, trim={0.25cm 0cm 2.1cm 1cm}, clip]{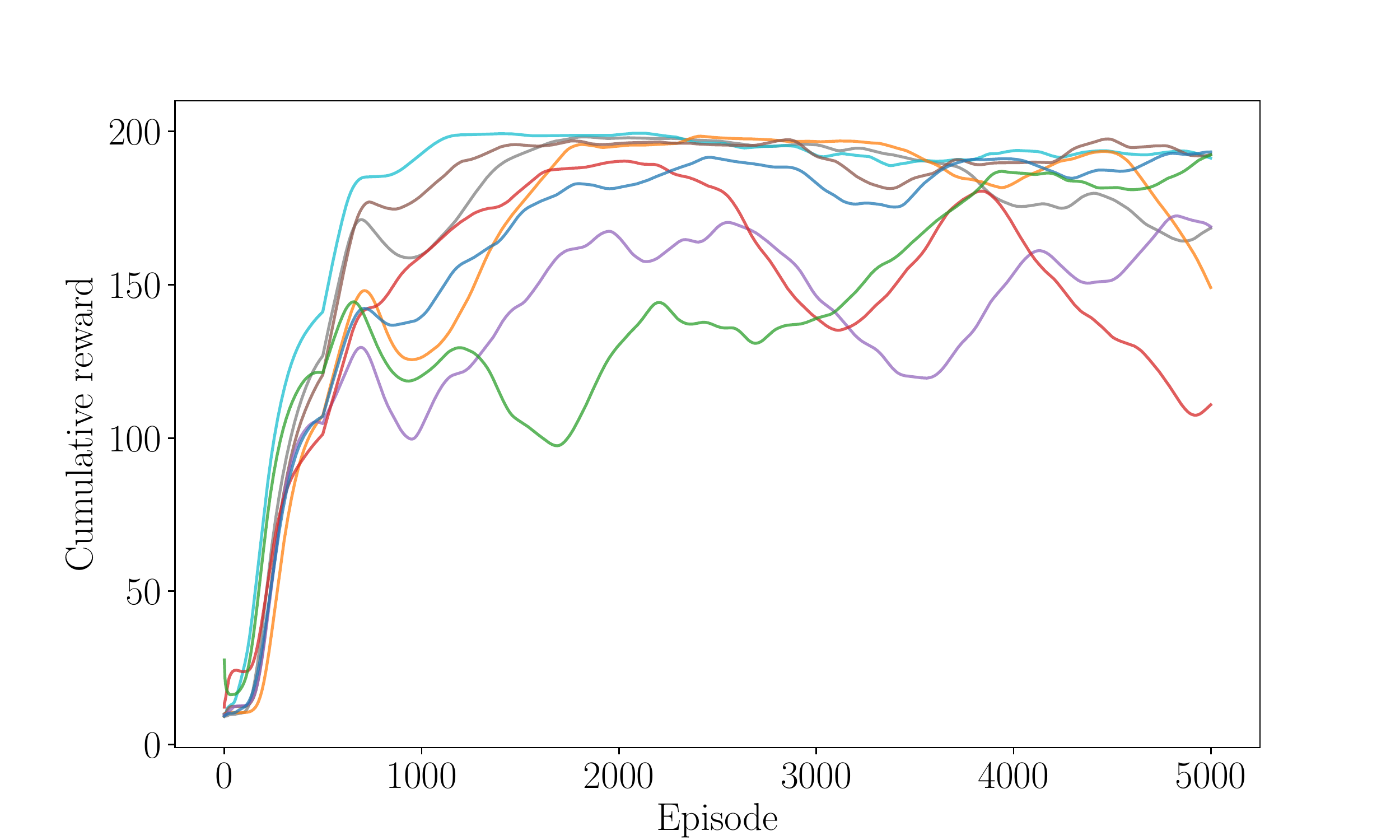}
        \caption{CartPole}
        \label{AdditionalResultsCartPole}
    \end{subfigure}
    \hfill
    \begin{subfigure}[b]{0.328\textwidth}
        \centering
        \includegraphics[width=1\linewidth, trim={0.25cm 0cm 2.1cm 1cm}, clip]{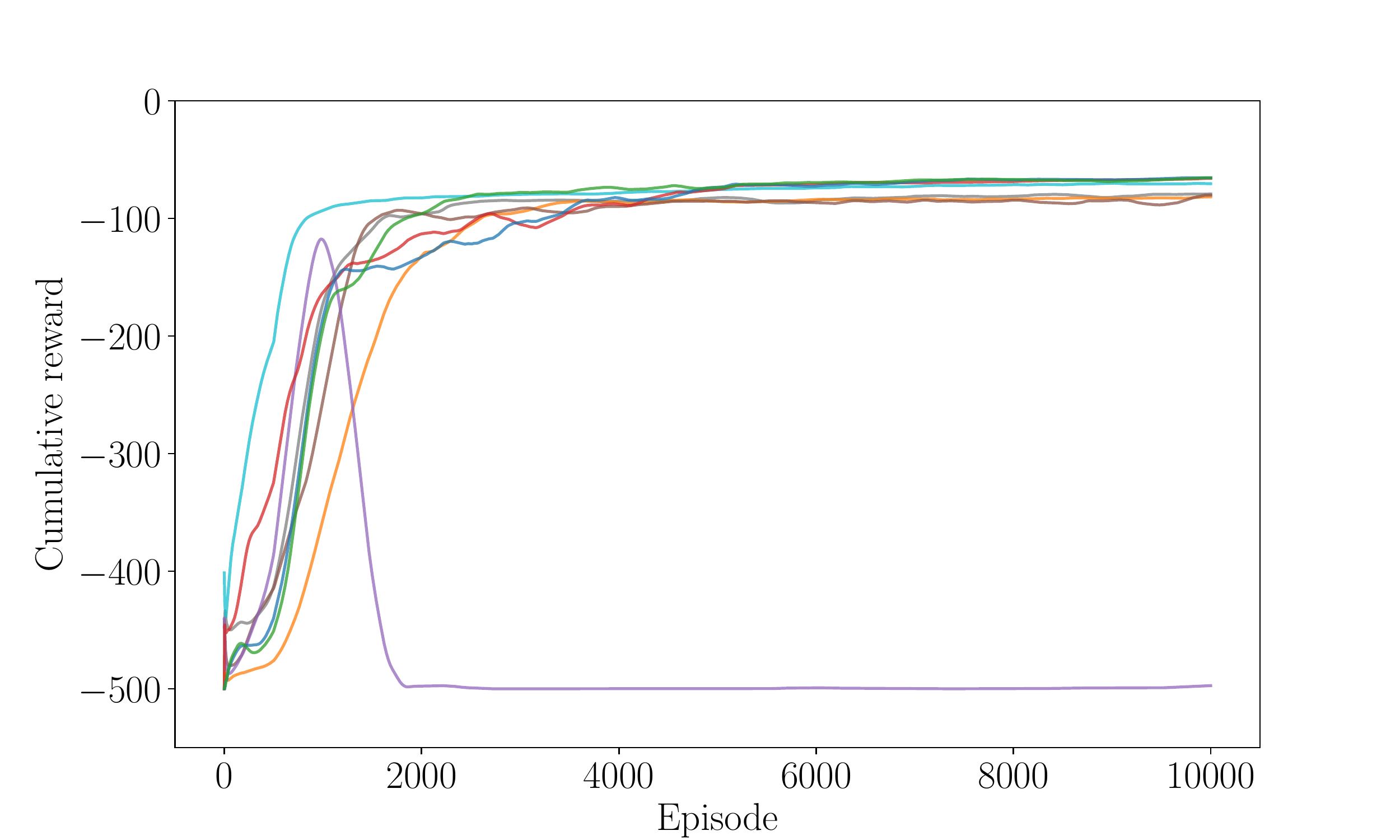}
        \caption{Acrobot}
        \label{AdditionalResultsAcrobot}
    \end{subfigure}
    \hfill
    \begin{subfigure}[b]{0.328\textwidth}
        \centering
        \includegraphics[width=1\linewidth, trim={0.25cm 0cm 2.1cm 1cm}, clip]{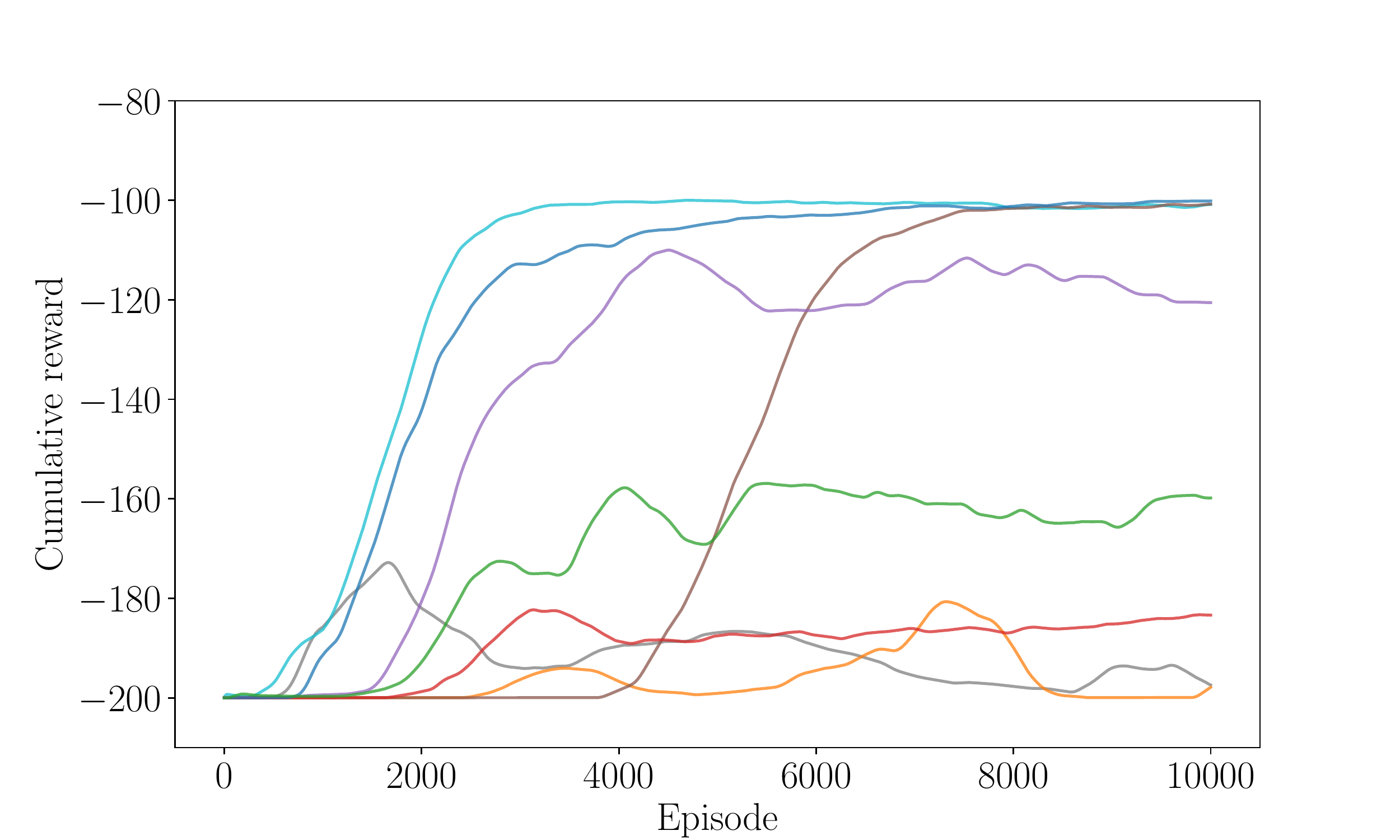}
        \caption{MountainCar}
        \label{AdditionalResultsMountainCar}
    \end{subfigure}
    \hfill
    \begin{subfigure}[b]{0.328\textwidth}
        \centering
        \includegraphics[width=1\linewidth, trim={0.25cm 0cm 2.1cm 1cm}, clip]{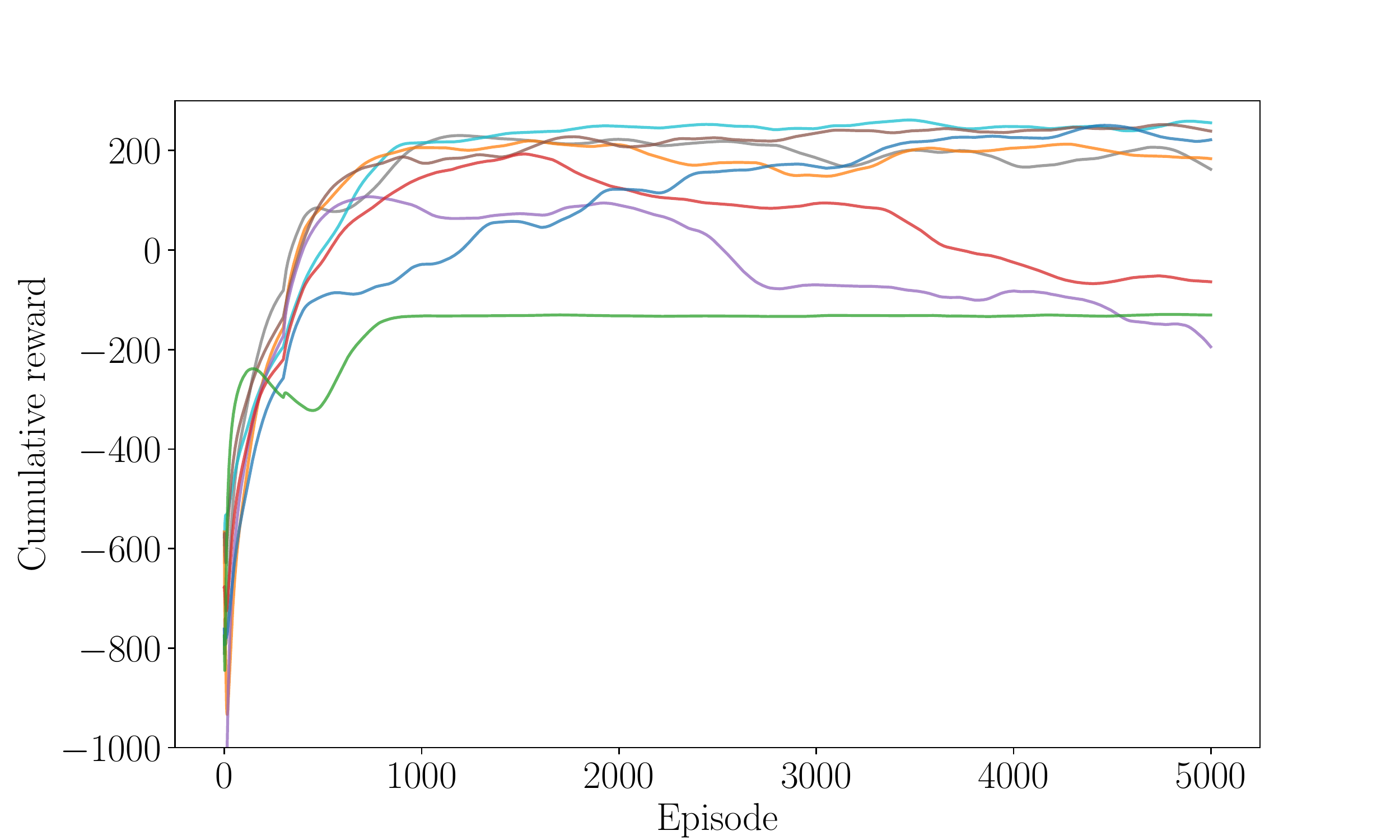}
        \caption{LunarLander}
        \label{AdditionalResultsLunarLander}
    \end{subfigure}
    \hfill
    \begin{subfigure}[b]{0.328\textwidth}
        \centering
        \includegraphics[width=1\linewidth, trim={0.25cm 0cm 2.1cm 1cm}, clip]{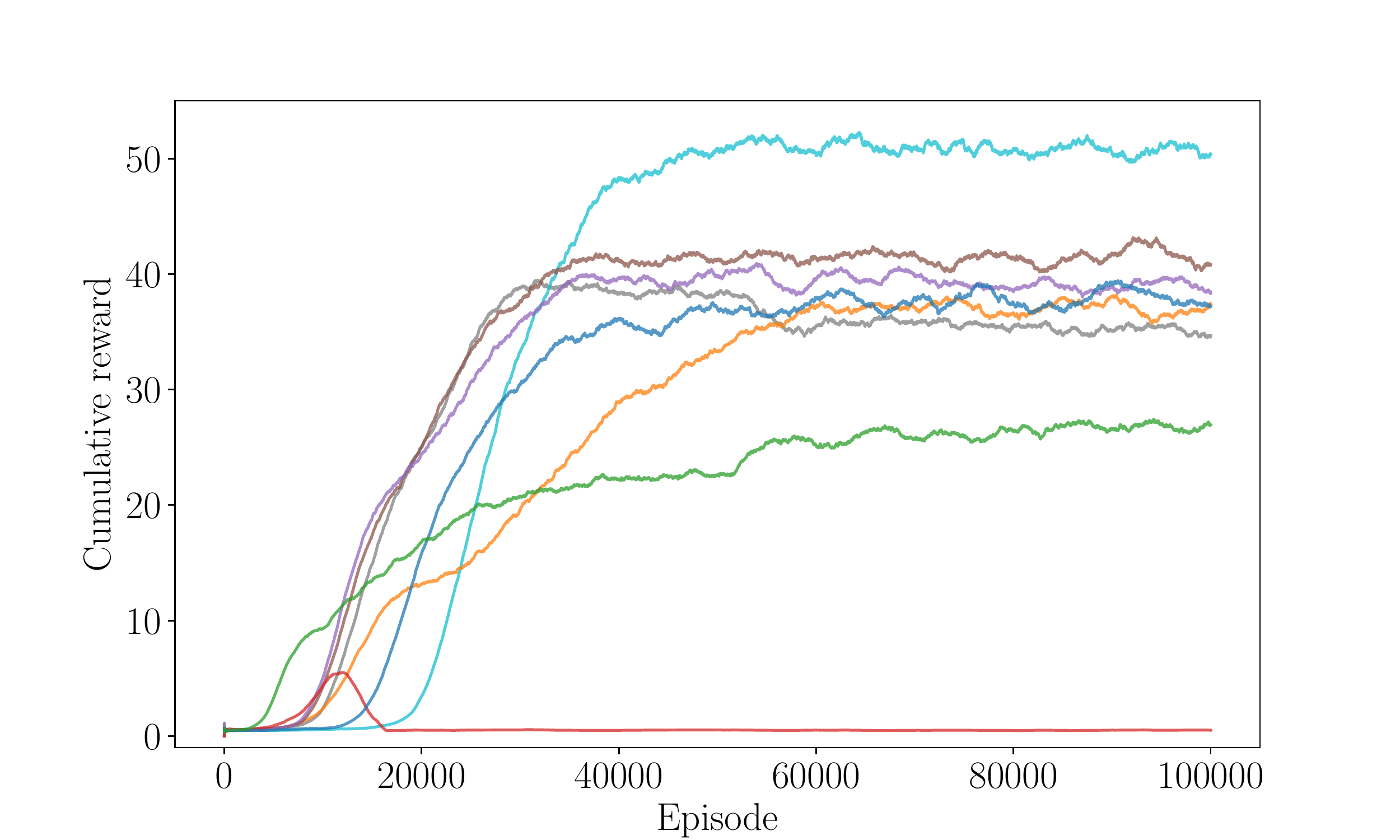}
        \caption{MinAtar Asterix}
        \label{AdditionalResultsMinAtarAsterix}
    \end{subfigure}
    \hfill
    \begin{subfigure}[b]{0.328\textwidth}
        \centering
        \includegraphics[width=1\linewidth, trim={0.25cm 0cm 2.1cm 1cm}, clip]{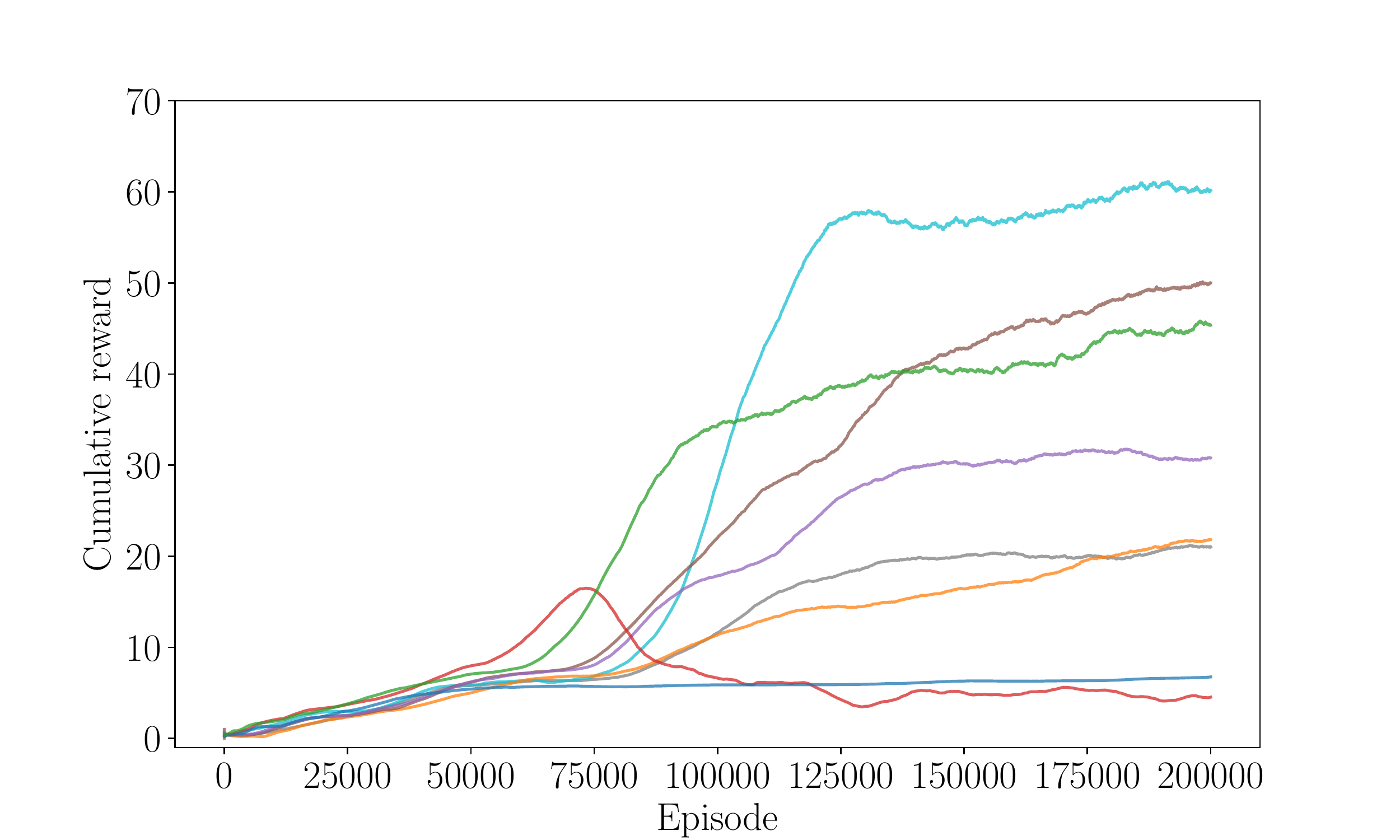}
        \caption{MinAtar Breakout}
        \label{AdditionalResultsMinAtarBreakout}
    \end{subfigure}
    \hfill
    \begin{subfigure}[b]{0.328\textwidth}
        \centering
        \includegraphics[width=1\linewidth, trim={0.25cm 0cm 2.1cm 1cm}, clip]{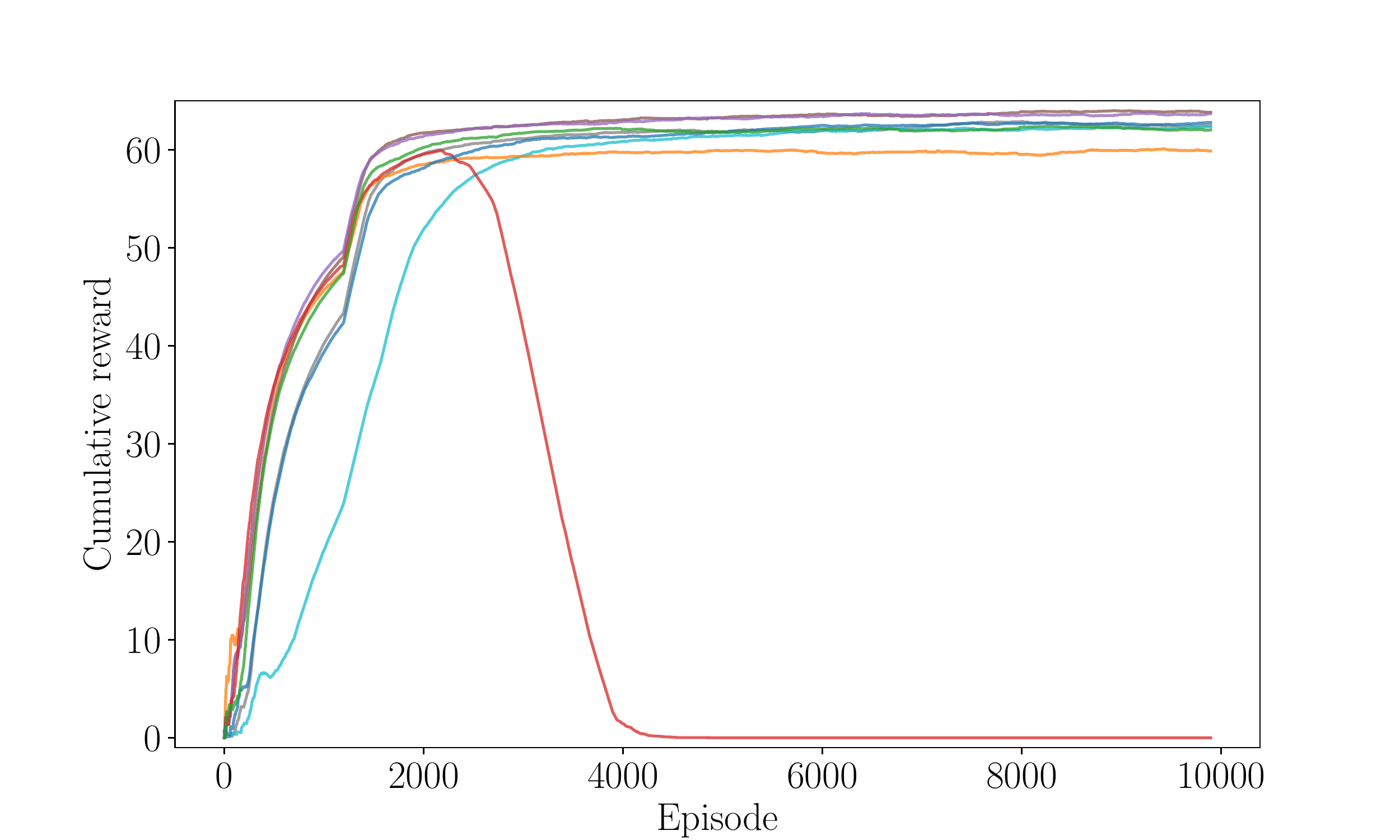}
        \caption{MinAtar Freeway}
        \label{AdditionalResultsMinAtarFreeway}
    \end{subfigure}
    \hfill
    \begin{subfigure}[b]{0.328\textwidth}
        \centering
        \includegraphics[width=1\linewidth, trim={0.25cm 0cm 2.1cm 1cm}, clip]{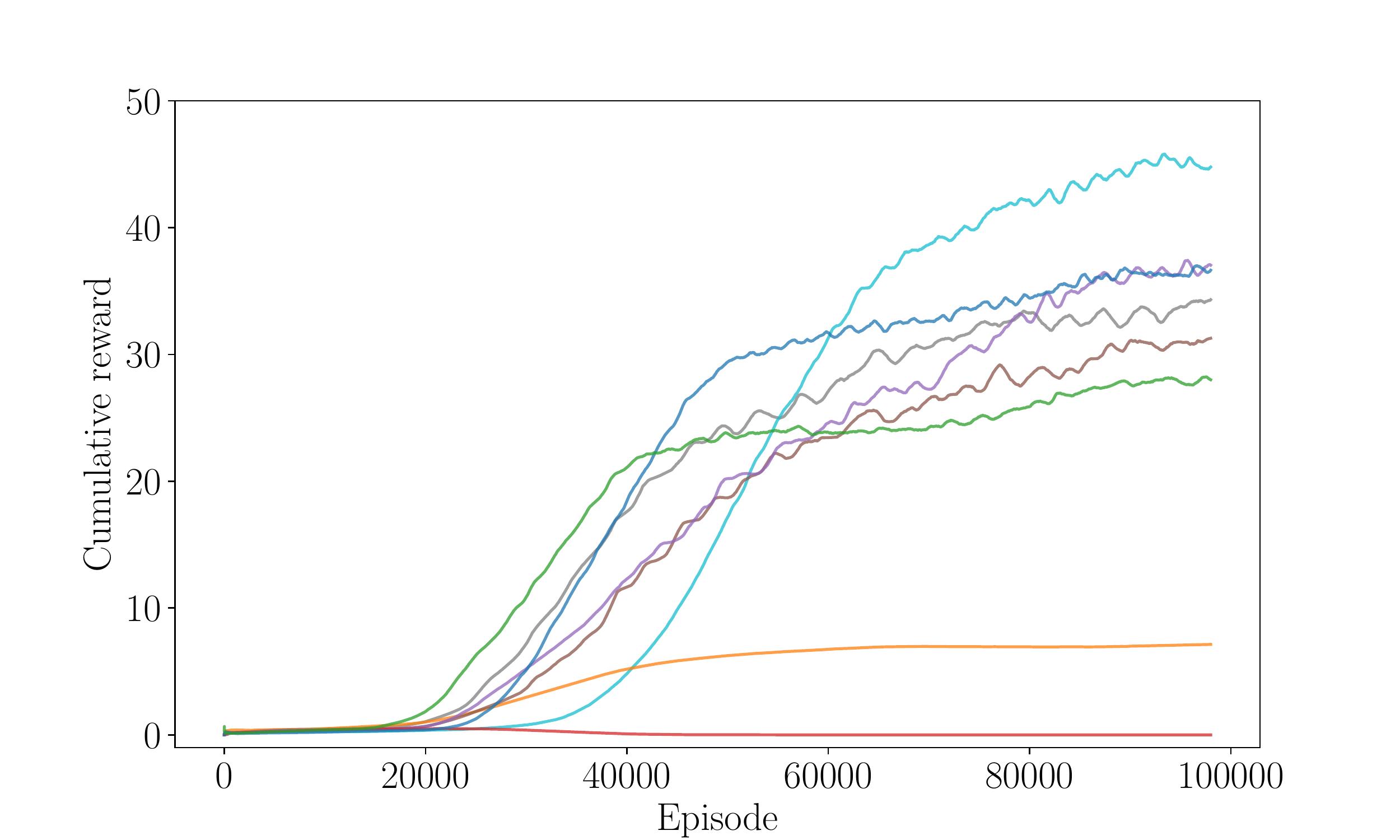}
        \caption{MinAtar Seaquest}
        \label{AdditionalResultsMinAtarSeaquest}
    \end{subfigure}
    \hfill
    \begin{subfigure}[b]{0.328\textwidth}
        \centering
        \includegraphics[width=1\linewidth, trim={0.25cm 0cm 2.1cm 1cm}, clip]{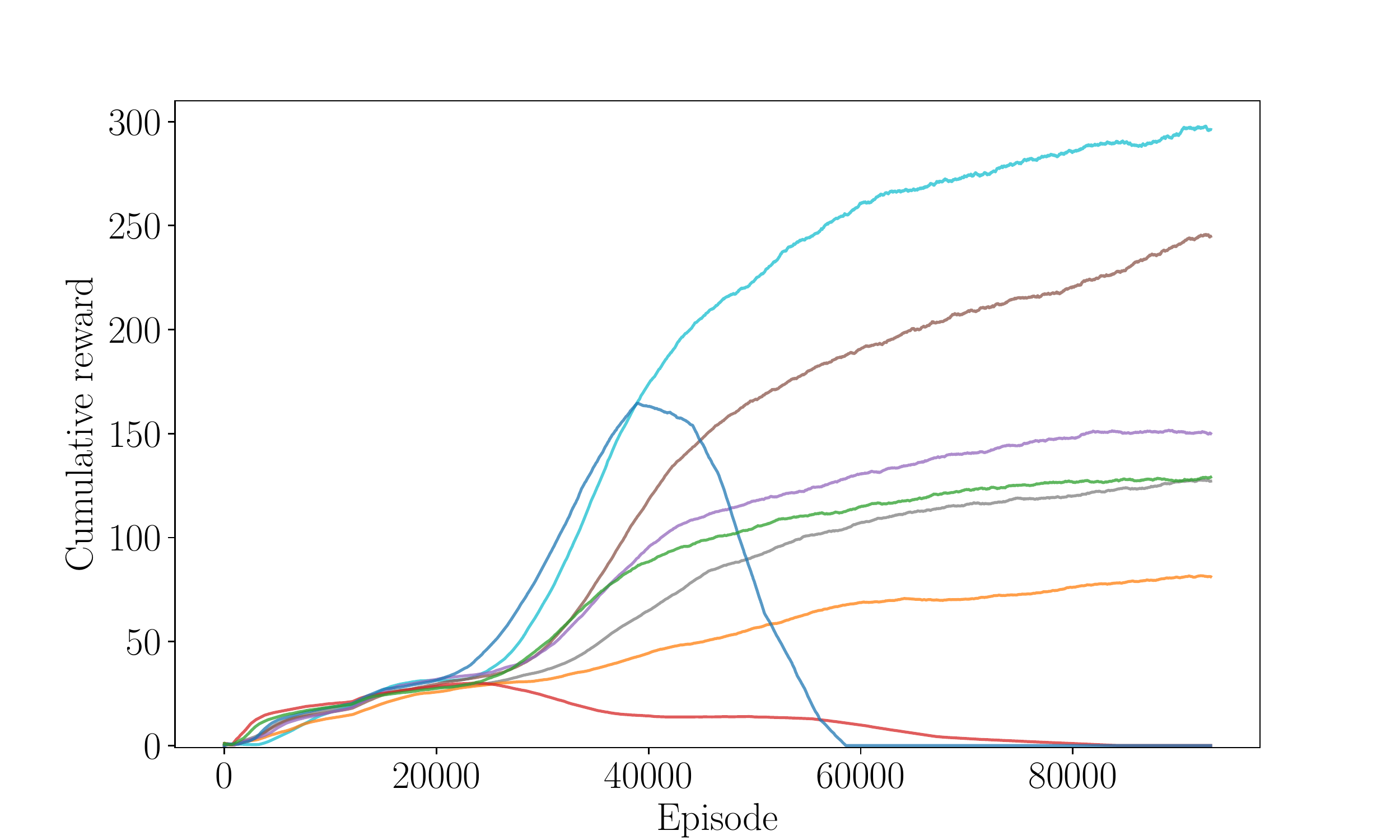}
        \caption{MinAtar SpaceInvaders}
        \label{AdditionalResultsMinAtarSpaceInvaders}
    \end{subfigure}
    \hfill
    \begin{subfigure}[b]{0.328\textwidth}
        \centering
        \includegraphics[width=1\linewidth, trim={0.25cm 0cm 2.1cm 1cm}, clip]{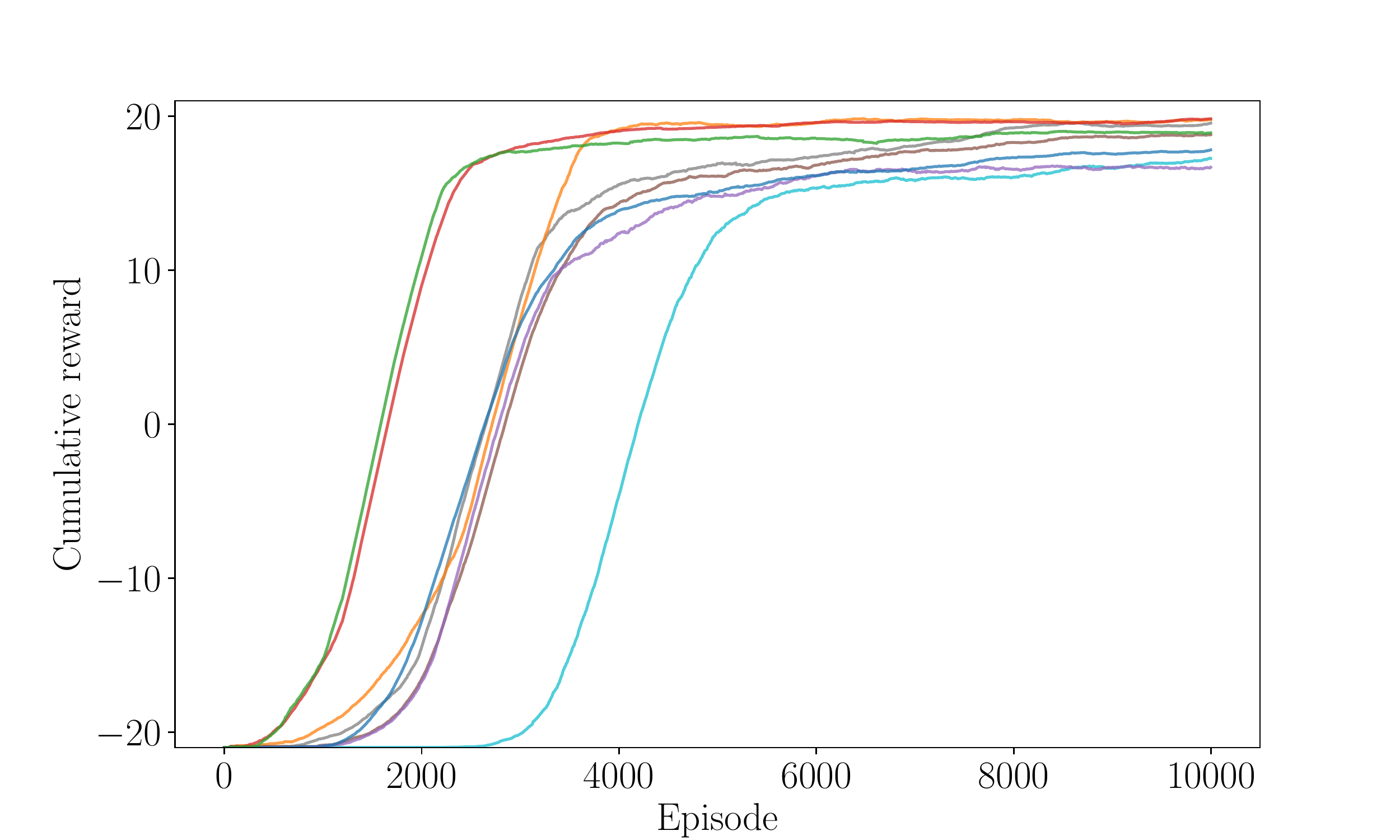}
        \caption{Atari Pong}
        \label{AdditionalResultsAtariPong}
    \end{subfigure}
    \hfill
    \begin{subfigure}[b]{0.328\textwidth}
        \centering
        \includegraphics[width=1\linewidth, trim={0.25cm 0cm 2.1cm 1cm}, clip]{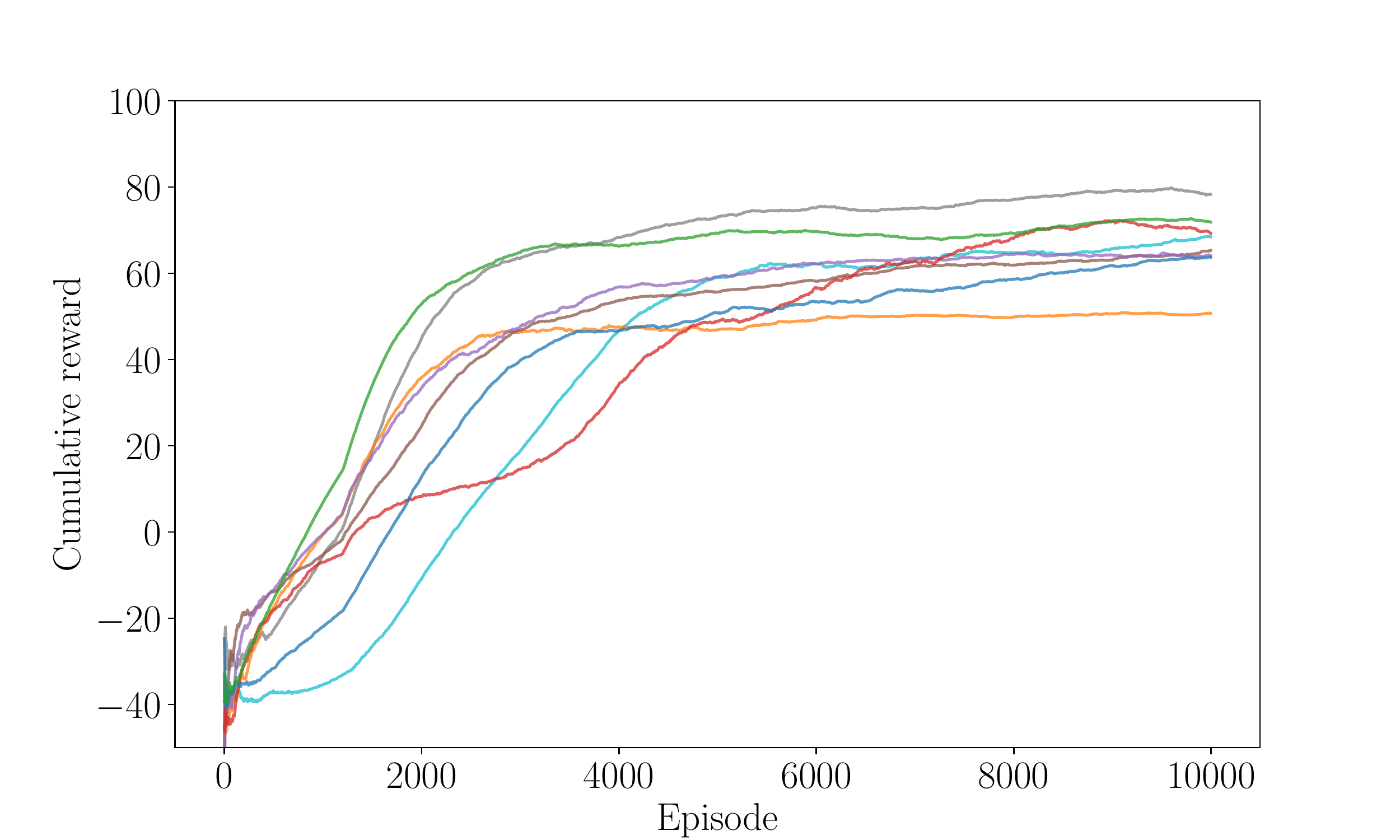}
        \caption{Atari Boxing}
        \label{AdditionalResultsAtariBoxing}
    \end{subfigure}
    \hfill
    \begin{subfigure}[b]{0.328\textwidth}
        \centering
        \includegraphics[width=1\linewidth, trim={0.25cm 0cm 2.1cm 1cm}, clip]{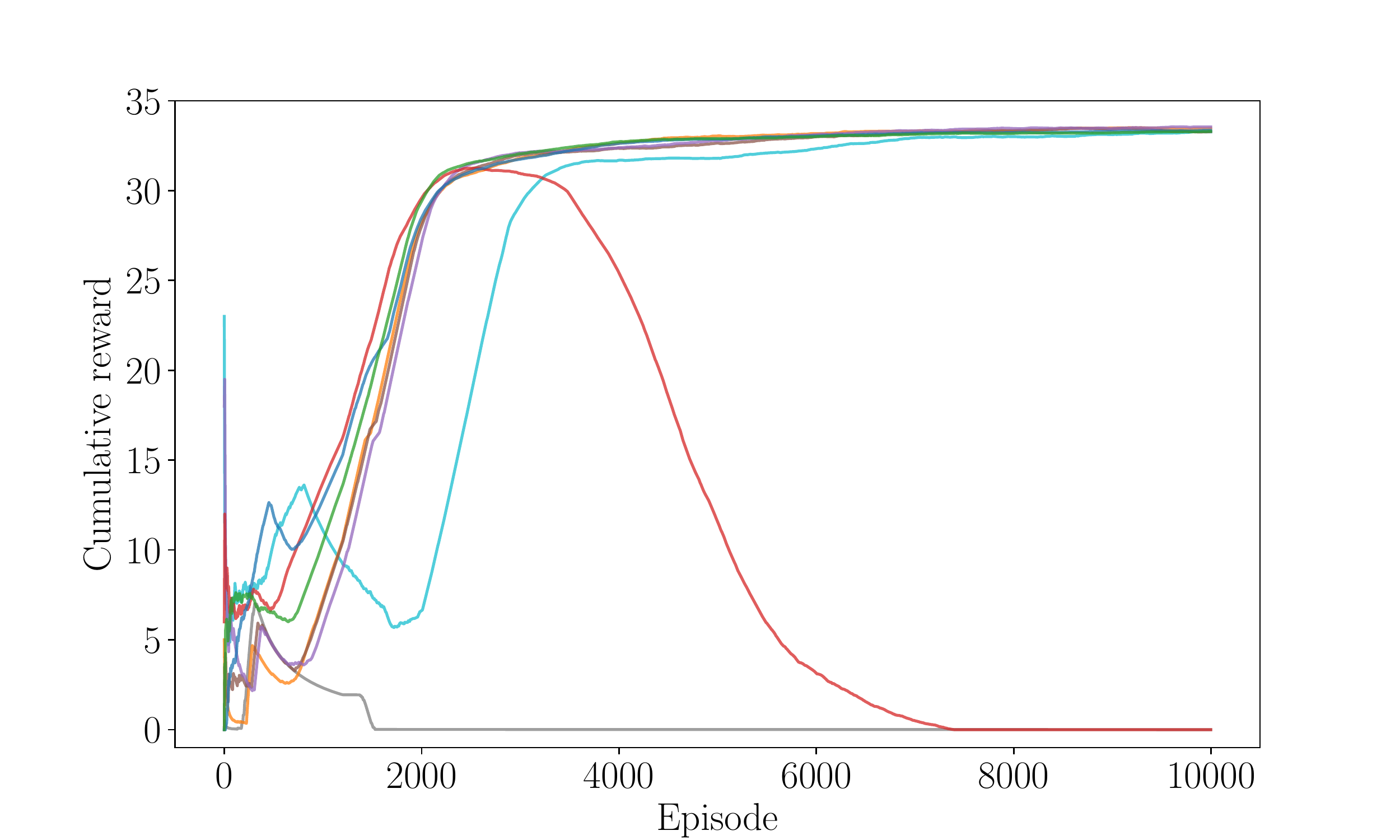}
        \caption{Atari Freeway}
        \label{AdditionalResultsAtariFreeway}
    \end{subfigure}
    \hfill
    \begin{subfigure}[b]{0.7\textwidth}
        \centering
        \includegraphics[width=1\linewidth, trim={1.9cm 13.1cm 1.9cm 12cm}, clip]{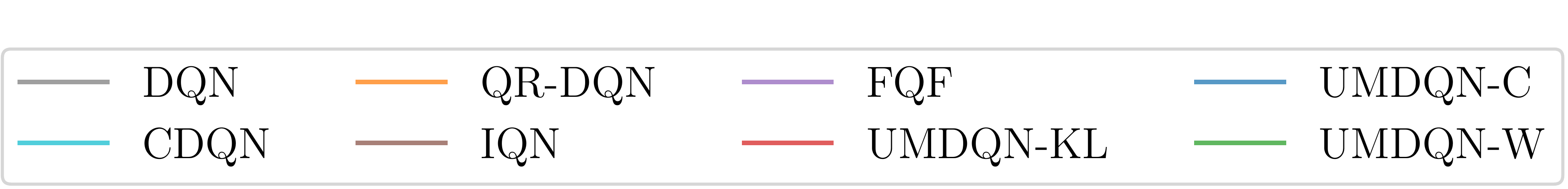}
        \label{AdditionalResultsLegend}
    \end{subfigure}
    \caption{Comparison of the UMDQN algorithm performance with the state-of-the-art distributional RL algorithms.}
    \label{AdditionalResults}
\end{figure}

Even though it is not an objective nor a contribution of the present research work, this section presents a brief comparison of the novel UMDQN algorithm with some state-of-the-art distributional RL algorithms, for the sake of completeness. In particular, the DQN \cite{Mnih2015}, CDQN \cite{Bellemare2017C51}, QR-DQN \cite{Dabney2018QRDQN}, IQN \cite{Dabney2018IQN} and FQF \cite{Yang2019FQF} are evaluated on the benchmark environments from Section \ref{SectionBenchmarkEnvironments} and \ref{AppendixEnvironments} alongside the three versions of the UMDQN algorithm. This empirical comparison is presented in Figure \ref{AdditionalResults}. As explained in Section \ref{SectionResults}, the results are averaged over five different random seeds for better reliability, and post-processed using the moving average technique to further smooth the curves. However, for the sake of readability, the variances of the distributional RL algorithms are no longer depicted on the plots.\\

Since the complete Atari-57 benchmark \cite{Bellemare2013} has not been taken into account for evaluating the novel UMDQN distributional RL algorithm proposed, this research work does not make any claim regarding the top-performing approach for this particular benchmark. Still, it is interesting to observe that the UMDQN algorithm achieves an impressive performance which is on par with the state-of-the-art distributional RL algorithms on the benchmark environments adopted in this research paper. Indeed, the UMDQN algorithm consistently ranks in the top three in terms of decision-making policy performance for all these benchmark environments. In the authors' opinion, this result consolidates the soundness of the proposed approach together with the relevance of the conclusions drawn by this research work.\\

For reproducibility purposes, the complete code executed to generate the results presented in this appendix, including the implementation of the state-of-the-art distributional RL algorithms, is publicly available at the following link: \url{https://github.com/ThibautTheate/Unconstrained-Monotonic-Deep-Q-Network-algorithm}. As far as hyperparameters tuning is concerned, values similar to those presented in \ref{AppendixParameters} are adopted to ensure a fair comparison between the different distributional RL algorithms.\\

\end{document}